\documentclass[11pt]{article}
\pdfoutput=1
\usepackage[final]{acl}   % camera-ready: non-anonymous, no line/page numbers

\usepackage{times}
\usepackage{latexsym}
\usepackage{amsmath,amssymb}
\usepackage{booktabs}
\usepackage{graphicx}
\graphicspath{{./}{../}}
\usepackage{hyperref}
\usepackage{xcolor}
\usepackage{enumitem}
\usepackage{multirow}
\usepackage{tabularx}
\newcolumntype{Y}{>{\centering\arraybackslash}X}
\usepackage{subcaption}
\usepackage{microtype}
\usepackage{placeins}
\usepackage{float}
\usepackage{afterpage}
\usepackage{stfloats}

\usepackage[T1]{fontenc}
\usepackage[utf8]{inputenc}

\usepackage[normalem]{ulem}
\newif\ifshowchanges
\showchangesfalse
\ifdefined\cleanbuild\showchangesfalse\fi   % set by: pdflatex -jobname=camera_ready_clean "\def\cleanbuild{1}\input{camera_ready.tex}"
\ifshowchanges
  \newcommand{\old}[1]{{\color{red}\sout{#1}}}
  \newcommand{\new}[1]{{\color{blue}#1}}
  \newcommand{\replace}[2]{\old{#1}\new{#2}}
\else
  \newcommand{\old}[1]{}
  \newcommand{\new}[1]{#1}
  \newcommand{\replace}[2]{#2}
\fi

\newcommand{\Rtext}{R_{\text{textbook}}}
\newcommand{\Rpat}{R_{\text{patient}}}
\newcommand{\Rclin}{R_{\text{note}}}
\newcommand{\Rcol}{R_{\text{colloquial}}}

\newcommand{\hlayer}{h_t^{\ell}}
\newcommand{\Dtest}{\mathcal{D}_{\text{test}}}

\title{Investigating Linear Probe Robustness to Linguistic Register, Medical Specialty, and Corpus Shifts in Medical QA}

\author{Nishant Mishra$^{1,2}$, Ameen Abu-Hanna$^{1,2}$, \and Iacer Calixto$^{1,2}$ \\
  $^{1}$Department of Medical Informatics, Amsterdam UMC, University of Amsterdam \\
  $^{2}$Amsterdam Public Health, Methodology, Amsterdam, The Netherlands \\
  \texttt{\{n.mishra, a.abu-hanna, i.coimbra\}@amsterdamumc.nl}}

\begin{document}
\maketitle

\begin{abstract}
%Linear classifiers trained on an LLM's hidden states (\textit{linear probes}) can flag factual errors from a single forward pass, which allows LLM assistants to be monitored cheaply for factuality. 
%The geometric reason behind the phenomenon is that true and false statements separate along a stable direction in hidden-state space (a \textit{truth direction}) that survives input shifts.
%Prior work disagrees on whether this holds, but the disagreement is hard to interpret because cross-dataset probe transfer experiments confound several kinds of input change at once.
%We isolate three such variables in medical QA: writing style (register), domain (medical specialty), and corpus (dataset).
%We build a controlled-rewriting benchmark of 4,000 medical-fact variants from 500 MedQA questions in four styles (textbook, patient, clinical note, colloquial), annotated with clinical specialties, and paired with a second medical exam corpus (MedMCQA) for cross-dataset evaluation. Probing four open-weight LLMs (2–8B), we find the truth direction is largely robust to writing style (mean $\Delta_r \approx 0.10$ AUROC on held-out facts) and to medical specialty ($\Delta_\text{specialty} \approx 0.03$), but loses most of its discrimination across datasets ($\Delta_\text{dataset} \approx 0.21$), a drop roughly two times the within-dataset register gap. The truth direction is real within a dataset and breaks across datasets: a linear probe is a robust within-dataset triage tool, not a standalone safety check.
Linear classifiers trained on hidden states of a large language model (LLM)---\textit{linear probes}---can flag factual errors from a single forward pass. %, which allows LLM assistants to be monitored cheaply for factuality. 
Geometrically, that implies that true and false statements separate along a stable direction in hidden state space, i.e., the \textit{truth direction}.
Prior work disagrees on whether this \replace{generalizes}{generalises} across input shifts, but the disagreement is hard to interpret because cross-dataset probe transfer experiments confound several kinds of input change at once.
We isolate three such variables in medical question-answering (QA): writing style (register), domain (medical specialty), and corpus (dataset).
\replace{We build a benchmark using 500 MedQA entries rewritten in four styles each (textbook, patient, clinical note, colloquial), annotated with clinical specialties, and paired with a second medical exam corpus (MedMCQA) for cross-dataset evaluation.}{We build a benchmark using 500 MedQA entries, each rewritten into four styles (textbook, patient, clinical note, colloquial), annotated with clinical specialty, and grouped with two other exam corpora, MedMCQA and MMLU-medical, for cross-dataset evaluation.}
\replace{Probing four open-weight LLMs (2–8B), we find that the truth direction is largely robust to writing style (mean $\Delta_\text{register} \approx 0.10$ AUROC on held-out facts) and to medical specialty ($\Delta_\text{specialty} \approx 0.03$), but loses most of its discrimination across datasets ($\Delta_\text{dataset} \approx 0.21$), a drop roughly two times the within-dataset register gap.}{Probing four open-weight LLMs (2--8B), we find that the truth direction is largely robust to writing style (mean $\Delta_\text{register} \approx 0.10$ AUROC on held-out facts) and to medical specialty ($\Delta_\text{specialty} \approx 0.03$), but degrades unevenly across corpora: by $0.12$ AUROC on MMLU-medical and by $0.21$ on MedMCQA, roughly twice the register gap.}
\new{The register result replicates with a second generator and carries over to human-written patient questions.}
\replace{Finally, the truth direction holds within a dataset, but breaks across datasets, suggesting the signal a linear probe recovers is partly bound to dataset structure rather than just medical knowledge.}{The truth direction is therefore largely stable within the medical domain but breaks under some corpus shifts, and question format does not explain the break, which suggests that the signal a linear probe recovers is partly bound to dataset structure rather than to medical knowledge alone.}
%a robust within-dataset triage tool, not a standalone safety check.
\end{abstract}

\section{Introduction}
\label{sec:intro}

A medical LLM that confidently generates a medically sound-looking but incorrect dosage, contraindication, or differential diagnosis can cause serious harm. Detecting such errors before they reach a clinician or a patient is the central safety problem of medical NLP, and a growing line of work proposes that linear classifiers reading a frozen model's hidden states (\emph{linear probes}) can flag factual incorrectness from a single forward pass \citep{azaria2023internal,marks2024geometry,burger2024truth}. The geometric intuition is that true and false statements live in different regions of the LLM's hidden state space, and the probe learns a hyperplane that separates them; the normal to that hyperplane is what prior work calls the \emph{truth direction} \citep{burns2023discovering, marks2024geometry}. Recent work has extended this to medical applications such as clinical knowledge recovery and adverse drug reaction detection \citep{berkowitz2025ping,berkowitz2025adr}.

For these probes to work in practice, their truth signal should be robust to the changes in input language a deployed system actually sees. A clinical assistant might be queried in formal examination-style prose by a medical student, in \replace{more }{}lay language on a consumer-health app, in telegraphic shorthand from an electronic health record, or in informal language by general users. These four ways of phrasing the same question are what we call linguistic registers, defined as stylistic configurations of vocabulary and syntax that hold meaning constant while varying surface form \citep{biber1988variation}. LLMs are pretrained on large quantities of formal, textbook-like data \citep{penedo2023refinedweb}, and medical QA benchmarks \replace{they are evaluated on }{}are also mostly written in \replace{the same}{that} register \citep{jin2021medqa,pal2022medmcqa,singhal2023medpalm}\replace{. That makes}{, which makes} textbook the natural register to train a probe on. Whether the signal it learns survives shifts to other registers 
%a deployed system sees 
has not yet been measured in medical LLMs.

Moreover, prior literature disagrees on whether such a signal genuinely exists as a \textit{stable} geometric property of LLM representations or not. \citet{marks2024geometry} and \citet{burger2024truth} report probes that recover a truth direction surviving causal interventions. \citet{Levinstein2024lie} and \citet{haller2025brittle} show that probes rely on superficial training-data resemblance and break under small surface perturbations, including paraphrasing and negation. \citet{orgad2025llms} report that truth probes only \replace{generalize}{generalise} across tasks that require similar underlying skills but break across task families.
The difficulty 
%in interpreting this behaviour is that any
here is that these
cross-dataset or cross-task 
%evaluation conflates
evaluations conflate
many factors at once: the writing style of the questions, the topics they cover, the way wrong answers were authored, and the dataset's overall difficulty. A negative cross-dataset result could reflect any (or a combination) of these. We argue that until probe transfer is measured along each axis in isolation, this disagreement cannot be resolved.

In this work, we measure how the truth direction \replace{generalizes}{generalises} under each of three input changes in isolation: writing style (\emph{register}), domain (\emph{medical specialty}), and \emph{corpus} (a different medical QA dataset). To isolate register, we build a controlled-rewriting benchmark with 4{,}000 variants of 500 MedQA (textbook) questions \citep{jin2021medqa}, each paraphrased into three registers (patient-facing, clinical-note, colloquial) with both the correct answer and a wrong-answer distractor. To isolate specialty, we use the S-MedQA \citep{yan-etal-2026-infect} specialty annotations, which map a subset of MedQA questions to 1 of 15 clinical specialties. To isolate corpus, we transfer the probe to MedMCQA \citep{pal2022medmcqa}, another medical exam dataset\new{, and to MMLU-medical \citep{hendrycks2021mmlu} as a third, format-matched corpus}. We find that the linear truth direction is robust to medical specialty ($\Delta_\text{specialty} \approx 0.03$ AUROC) and largely robust to register ($\Delta_\text{register} \approx 0.10$ AUROC), but \replace{breaks across corpora ($\Delta_\text{dataset} \approx 0.21$).}{degrades unevenly across corpora. Transfer to MMLU-medical decreases AUROC by $0.12$, while transfer to MedMCQA by $0.21$ ($\Delta_\text{dataset}$). We replicate our main finding regarding register with a second generator. We also show that the performance is unchanged when low-fidelity wrong-answer rewrites are filtered out, and carries over to human-written patient questions.}

We confirm the signal is dominantly linear since an unregularized difference-of-means probe \citep{marks2024geometry} matches 
ours,
%logistic regression, 
while a nonlinear MLP probe adds no meaningful gain. The probe outperforms basic output-based uncertainty baselines (entropy and self-consistency) by 6–11 AUROC points, and matches a self-evaluation baseline that reads the LLM's own \texttt{yes} token probability when asked whether the answer is correct \citep{kadavath2022language}. Finally, raw probe scores are severely miscalibrated. Together, our findings indicate that linear probes are a viable within-dataset monitoring tool for medical QA, but not yet a standalone safety check that transfers across corpora. \new{All corpora we use are exam-style multiple-choice QA. Our benchmark, code, prompts, and judging rubric are publicly released.\footnote{\new{\url{https://github.com/mnishant2/MedProbe_release}}}}

\section{Methodology}
\label{sec:methodology}

\subsection{The three shifts}
\label{sec:methodology:shifts}
We study probe transfer along three input changes: \textbf{register} (the same MedQA medical statement rewritten into four surface forms), \textbf{specialty} (held-out medical specialties from S-MedQA), and \textbf{corpus} (MedQA to MedMCQA, a second medical exam dataset). Holding two of these fixed while varying the third lets us attribute an observed change in probe behaviour to that single shift.
% that any cross-dataset evaluation simultaneously varies
% in contrast to prior cross-dataset evaluations which conflate all three \citep{orgad2025llms}.

\subsection{Linguistic registers}
\label{sec:methodology:registers}
Each of the four registers we evaluate has a distinct linguistic profile. \emph{Textbook} ($\Rtext$) is the formal prose register that dominates medical LLM evaluation benchmarks, e.g., MMLU \citep{hendrycks2021mmlu}
%(USMLE, MMLU-medical, PubMedQA) 
and is closest to the medical content seen during pretraining. \emph{Patient-facing} ($\Rpat$) is grammatical but informal, with full sentences and lay vocabulary. \emph{Clinical note} ($\Rclin$) uses telegraphic shorthand, sentence fragments, and heavy abbreviation. \emph{Colloquial} ($\Rcol$) uses loose syntax, lowercase, and internet filler (``tbh'', ``lol''). \old{Holding the underlying medical fact fixed while varying the register lets us attribute a change in probe behaviour to surface form rather than content.}

\subsection{Probing protocol}
\label{sec:methodology:probing}
Each input is presented to the LLM as a chat-templated \texttt{yes/no} prompt, \emph{``Is this answer medically correct? Respond with Yes or No.''} The yes/no framing concentrates the truth signal at the last question token, consistent with prior probing work \citep{azaria2023internal,marks2024geometry}, and keeps the sentence-final completion fixed across registers so register effects are not confounded by prompt format. We capture the hidden state representation at the last question token at every second layer of the LLM and use it to train our linear probe, which is an $L_2$-regularised logistic regression model in line with probe configurations used in previous work \citep{azaria2023internal,li2023inferencetime}.
\section{Experimental Setup}
\label{sec:setup}

%\section{Methodology}
%\label{sec:setup}
\subsection{Data}
\label{sec:setup:data}

We primarily use two medical QA datasets. MedQA \citep{jin2021medqa} provides USMLE-style clinical vignettes in four-option multiple-choice form, and is the source corpus for our rewriting benchmark and within-corpus experiments. MedMCQA \citep{pal2022medmcqa} provides Indian medical entrance exam questions in the same multiple-choice format. The two corpora differ in distractor style, specialty mix, and surface conventions and not only register, making MedMCQA a useful test for cross-corpus stability.

We randomly sample 500 facts from the MedQA's test split. For each fact $i$ we pair the question with the gold correct answer $a_i^+$ and one of the three MCQ distractors $a_i^-$, sampled uniformly at random when several are available. Each (question, answer) pair is then rewritten by Claude Sonnet 4.5 into the four registers of \S\ref{sec:methodology:registers}, with both the correct and the wrong-answer polarity preserved, yielding $500 \times 4 \times 2 = 4{,}000$ controlled variants. Full details on dataset generation, including prompts, evaluation, model selection, and label noise in incorrect variants, are provided in Appendix~\ref{app:dataset}. From MedMCQA, we draw 500 facts from the validation split for cross-corpus probe transfer, and a separate 100-fact subset, which we additionally rewrite into the same four registers to verify that the within-corpus register pattern holds within MedMCQA itself.

\new{We add three evaluation-only sets, never used for training. MMLU-medical \citep{hendrycks2021mmlu} contributes 500 items from five subjects (anatomy, clinical knowledge, college and professional medicine, medical genetics) in the same four-option format as MedQA and MedMCQA. The reformatted MedMCQA set contains its 100 items rewritten by Sonnet into MedQA-style vignettes with the correct answer and distractor held fixed verbatim, so only question format moves. MedRedQA \citep{nguyen-etal-2023-medredqa} provides 84 unedited questions posted by members of the public and answered by verified clinicians. The clinician reply is used only to source the correct claim and is never shown to the probe (Appendix~\ref{app:medredqa}).}

% Sonnet was selected after a 50-fact pilot against GPT-4o-mini and Gemini 3 Flash, as the only generator clearing a factual-fidelity-on-wrong threshold of 0.85 across all four registers (pooled fidelity 0.878 on the full 500-fact run). The full pilot, the cross-family LLM-as-judge rubric, inter-rater agreement, and the residual $\sim$12\% label noise on wrong variants are reported in

We train probes on a fact-level 80/20 split of the textbook variants ($\mathcal{D}_\text{train}$ and $\mathcal{D}_\text{test}$), keeping both polarities of any given fact on the same side of the split. The textbook held-out set ($n{=}200$) serves as the in-distribution baseline. For the three non-textbook registers we evaluate on the corresponding test variants per register. \new{Table~\ref{tab:splits} lists the training, selection, and evaluation sets of every experiment.}

% NEW (camera-ready): splits table, metareview item 2 / kJP9 W1
\begin{table}[t]
\centering
%\scriptsize
%\setlength{\tabcolsep}{2.5pt}
%\renewcommand{\arraystretch}{1.0}
%\begin{tabularx}{\columnwidth}{@{}>{\raggedright\arraybackslash}p{1.45cm}>{\raggedright\arraybackslash}X>{\raggedright\arraybackslash}X@{}}
\resizebox{\columnwidth}{!}{%
\begin{tabular}{p{2.5cm}p{6.2cm}p{5cm}}
\toprule
\textbf{Experiment} & \textbf{Train set / layer selection} & \textbf{Evaluation} \\
\midrule
Register ($\mathcal{M}_\text{textbook}$) & 400 textbook facts (80\%) / best layer per (LLM, register) chosen on held-out set & same 100 held-out facts in all 4 registers ($n{=}200$ variants each) \\
Mixed register ($\mathcal{M}_\text{all}$) & 400 facts with equal share of the 4 registers / as above & as above \\
Specialty ($\mathcal{M}_\text{specialty}$) & textbook variants of 7 of 15 specialties (5 random partitions) / 80/20 inner split & textbook variants of the 8 held-out specialties \\
Corpus & $\mathcal{M}_\text{textbook}$, frozen at its MedQA-best layer & 500 MedMCQA val.\ facts; 500 MMLU-medical; 100 reformatted MedMCQA; 84 MedRedQA \\
Within-MedMCQA & MedMCQA-textbook 80\% of 100 facts & 20 held-out facts, 4 registers \\
\bottomrule
%\end{tabularx}
\end{tabular}}
\caption{\new{Splits per experiment, all at the fact level: no fact appears on both sides of a split, and both polarities of a fact move together.}}
\label{tab:splits}
\end{table}

\paragraph{Specialty annotations.} S-MedQA \citep{yan-etal-2026-infect} categorises a subset of MedQA questions by clinical specialty. An exact-question-match join with our 500-fact sample maps 351 of them to one of 15 specialties; the remaining 149 are excluded from the specialty experiment of \S\ref{sec:setup:specialty}.

\subsection{Models}
\label{sec:setup:models}
We probe four open-weight instruction-tuned LLMs spanning three architecture families: \texttt{Gemma-2-2B-it} \citep{gemma2024}, \texttt{Gemma-3-4B-it} \citep{gemma3}, \texttt{Qwen2.5-7B-Instruct} \citep{qwen25}, and \texttt{Llama-3-8B-Instruct} \citep{llama3} to ensure our findings \replace{generalize}{generalise} across scale and architecture. \replace{Our selection targets the 2B-8B parameter regime, consistent with prior interpretability work \mbox{\citep{marks2024geometry,burger2024truth,Levinstein2024lie,orgad2025llms}} and tractable for exhaustive per-layer probe training across our experimental grid.}{The 2B--8B regime matches prior interpretability work \citep{marks2024geometry,burger2024truth,Levinstein2024lie,orgad2025llms} and keeps exhaustive per-layer probing tractable.} Per-LLM specifications, inference settings, and the layer indices are detailed in Appendix~\ref{app:reproducibility}.

% Linear probing requires hidden-state activations, which proprietary APIs do not expose.
% \subsection{Models}
% \label{sec:setup:models}

% We probe four open-weight instruction-tuned LLMs spanning three architecture families: \textbf{Gemma-2-2B-it} \citep{gemma2024}, \textbf{Gemma-3-4B-it} \citep{gemma3}, \textbf{Qwen2.5-7B-Instruct} \citep{qwen25}, and \textbf{Llama-3-8B-Instruct} \citep{llama3}. Hidden-state dimensions range from 2304 to 4096 and depths from 26 to 34 layers. Per-LLM specifications, inference settings, and the layer indices we extract activations from are in Appendix~\ref{app:reproducibility}.
% \subsection{Experiments}
% \label{sec:setup:experiments}

% We run six experiments. The first three measure probe transfer along the three input shifts of \S\ref{sec:methodology:shifts}: register (\S\ref{sec:setup:register}), specialty (\S\ref{sec:setup:specialty}), and corpus (\S\ref{sec:setup:corpus}). The remaining three characterize the probe along dimensions orthogonal to transfer, namely whether the truth signal is linear (\S\ref{sec:setup:linearity}), whether it adds information beyond the LLM's own output-level uncertainty (\S\ref{sec:setup:baselines}), and whether its outputs can be read as probabilities (\S\ref{sec:setup:calibration}). All experiments use the probing protocol of \S\ref{sec:methodology:probing} unless stated otherwise.
\subsection{Evaluation}
\label{sec:setup:eval}

\paragraph{Evaluation metrics.} Our primary metric is the Area Under the Receiver Operating Characteristic curve (AUROC), the probability that a randomly drawn correct example receives a higher probe score than a randomly drawn incorrect one, where 0.5 corresponds to chance and 1.0 to perfect ranking. For the calibration analysis we additionally report Expected Calibration Error (ECE) with 10 equal-width probability bins \citep{guo2017calibration}, with values below 0.05 generally considered well-calibrated. We report ECE on raw probe outputs and after two post-hoc calibrators: Platt scaling \citep{platt1999probabilistic}, which fits a one-parameter logistic transform of the raw scores, and isotonic regression \citep{zadrozny2002transforming}, which fits a piecewise-constant monotonic transform with more parameters but a stronger fit when sample size permits. \new{Alongside AUROC we report accuracy and F1 at a fixed $0.5$ cutoff and accuracy at the threshold maximising Youden's $J$. Because every test set is balanced by construction (one correct and one wrong variant per fact), balanced accuracy equals plain accuracy.}

\paragraph{Confidence intervals.} All AUROC numbers carry a 95\% bootstrap CI from 1{,}000 iterations of \emph{fact-level} resampling. Both polarities of a given fact enter or leave the bootstrap sample together, since treating them as independent would understate CI width. Two CIs are \emph{disjoint} when they do not overlap.\new{ Disjoint CIs are a conservative sign of a real difference; overlapping CIs do not show that two conditions are equal.}

\subsection{Experiments}
\label{sec:setup:experiments}

\subsubsection{Register transfer}
\label{sec:setup:register}

To measure robustness to surface form, we train an $L_2$-regularised logistic regression probe ($\mathcal{M}_\text{textbook}$, $C=1.0$) on every second layer of the LLM using the textbook 80\% training split. The probe takes the hidden state at the last question token as input and predicts whether the medical statement is correct via binary cross-entropy loss. We evaluate every register on the \emph{same held-out 20\% fact split} ($n{=}200$) used for textbook, so the non-textbook variants come from facts the probe has never seen at training time and the four register conditions are matched on the underlying fact set. For each (LLM, register) condition we evaluate every layer's probe on that condition's test set and report the best-layer AUROC. The choice is stable, with a median of 8 layers falling within bootstrap-CI overlap of the best layer (Appendix~\ref{app:probe-ablations}); we also report two alternative single-layer rules (Appendix~\ref{app:layer-rules}). The register transfer gap is $\Delta_\text{register} = \text{AUROC}(\mathcal{D}_\text{test}^{r}) - \text{AUROC}(\mathcal{D}_\text{test}^{R_\text{textbook}})$, where a negative $\Delta_\text{register}$ indicates a performance drop on register $r$.

We also train two other probe variants in the same register transfer context. A label-permutation probe ($\mathcal{M}_\text{perm}$) trained on shuffled correctness labels serves as a sanity check that the probe reads a true correctness signal rather than spurious activation structure. A mixed-register probe ($\mathcal{M}_\text{all}$) trained on data drawn equally from all four registers tests whether any observed register gap is structural or merely a training-coverage deficit; we report the per-register recovery $\rho_r = \text{AUROC}_{\mathcal{M}_\text{all}}(\mathcal{D}_\text{test}^{r}) - \text{AUROC}_{\mathcal{M}_\text{textbook}}(\mathcal{D}_\text{test}^{r})$, where positive $\rho_r$ indicates that mixed-register training improves on register $r$. \new{To check that the register effect is not a property of one rewriter, we regenerate the benchmark with a second generator from a different model family (Gemini 3 Flash Preview) and repeat the full $\mathcal{M}_\text{textbook}$ protocol on it (see Appendix~\ref{app:generator-ablation}). To validate that results are not driven by label noise in the wrong-answer rewrites, we repeat the analysis after removing the wrong variants that the judges score as low-fidelity (see Appendix~\ref{app:fidelity-filter}).}

% \paragraph{Main probe $\mathcal{M}_\text{textbook}$.} For each of the four LLMs we train one $L_2$-regularised logistic regression probe per layer ($C=1.0$, every second layer of the LLM) on the textbook 80\% training split with binary cross-entropy loss. The probe takes as input the hidden state at the last question token and predicts whether the medical statement is correct. For each (LLM, register) condition we evaluate every layer's probe on that register's test set (textbook held-out 20\%; all 1{,}000 variants per register for the three non-textbook registers) and report the best-layer AUROC. The layer choice is stable since the median number of layers within bootstrap-CI overlap of the best layer is 8, so the choice is not brittle (layer-wise sweeps in Appendix~\ref{app:probe-ablations}, Figure~\ref{fig:layerwise}).

% \paragraph{Mixed-register probe $\mathcal{M}_\text{all}$.} We repeat the main-probe protocol unchanged except that the training data is drawn equally from all four registers instead of from textbook alone. The fact-level 80/20 split, the per-layer training, the per-(LLM, register) best-layer selection, and the input token position are all identical to $\mathcal{M}_\text{textbook}$. Evaluation uses the same three held-out registers, so the comparison against $\mathcal{M}_\text{textbook}$ isolates the effect of training-data composition.

\subsubsection{Specialty transfer}
\label{sec:setup:specialty}

To measure robustness to medical sub-domain, we use the 351 specialty-tagged facts from \S\ref{sec:setup:data} and partition the 15 specialties into a 7-specialty training half and an 8-specialty held-out half. For each LLM we draw five independent random partitions, and for each partition we train a probe ($\mathcal{M}_\text{specialty}$) using the same logistic protocol as in \S\ref{sec:setup:register} on the textbook variants in the training half, then evaluate on the textbook variants in the held-out half. Because register is fixed at textbook in both halves, any AUROC change is isolated to the specialty axis rather than surface form. The in-distribution reference for each split is a within-train, fact-level 80/20 held-out set, and the specialty transfer gap is $\Delta_\text{specialty} = \text{AUROC}(\mathcal{D}_\text{held-out specialties}) - \text{AUROC}(\mathcal{D}_\text{within-train held-out})$, averaged across the five splits.

% \paragraph{Specialty probe $\mathcal{M}_\text{specialty}$.} Of the 500 MedQA facts in our benchmark, 351 were matched to one of 15 clinical specialties through the S-MedQA join (\S\ref{sec:setup:data}). For each LLM we draw five independent random partitions of these 15 specialties into a 7-specialty training half and an 8-specialty held-out half, train the main-probe protocol on the textbook variants in the training half, and evaluate on the textbook variants in the held-out half. Register is held fixed at textbook in both halves, so any drop is attributable to the specialty axis rather than to surface form. The in-distribution baseline for each split is a within-train fact-level 80/20 held-out set, and the reported $\Delta_\text{specialty}$ is the mean drop across the five splits.
\subsubsection{Corpus transfer}
\label{sec:setup:corpus}

To measure transfer across dataset distributions, we apply the MedQA-trained main probe ($\mathcal{M}_\text{textbook}$) to the 500 MedMCQA validation facts described in \S\ref{sec:setup:data}, without retraining. We evaluate it at the exact same best-textbook layer chosen on MedQA. The cross-corpus gap is $\Delta_\text{dataset} = \text{AUROC}(\mathcal{D}_\text{MedMCQA}) - \text{AUROC}(\mathcal{D}_\text{test}^{R_\text{textbook}})$. \new{The same frozen probe is also applied to MMLU-medical, the reformatted MedMCQA items, and the MedRedQA questions (\S\ref{sec:setup:data}), which separates a change of corpus from a change of question format and tests the probe on human-written text.}

To interpret this drop, we conduct a within-MedMCQA replication on the 100-fact MedMCQA subset (\S\ref{sec:setup:data}), rewritten into the four registers using Sonnet. We train a fresh probe on the MedMCQA-textbook 80\% split using the same protocol as $\mathcal{M}_\text{textbook}$ and evaluate on the three non-textbook registers. This replication has two purposes. It verifies that our register-invariance finding \replace{generalizes}{generalises} to a second independent dataset, and it tests whether the zero-shot cross-corpus failure reflects mismatched linguistic styles or a structural failure tied to dataset-specific item format (Appendix~\ref{app:medmcqa-register}).

\subsubsection{Linearity}
\label{sec:setup:linearity}

To test whether the truth signal is strictly linear, we compare the logistic probe ($\mathcal{M}_\text{textbook}$) against two structural alternatives evaluated at the exact same layer. First, the MLP probe ($\mathcal{M}_\text{MLP}$) replaces the linear classifier with a one-hidden-layer network (128 ReLU units, $L_2$ regularisation), trained on the textbook split with early stopping on a 10\% within-train validation set. If a deeper classifier substantially improves AUROC, the signal is not fully captured by a linear direction.

Second, following \citet{marks2024geometry}, we construct a difference-of-means probe ($\mathcal{M}_\text{diff}$) whose direction is the difference between the per-class hidden-state means on the textbook training split, with the decision boundary at the midpoint of the two projected class means. Because it has no learned weights and no regularisation, this comparison tests whether the truth signal exists as a pure geometric property of the representation space, and it isolates whether the $L_2$ penalty in $\mathcal{M}_\text{textbook}$ helps or hinders the signal.

% \paragraph{MLP probe $\mathcal{M}_\text{MLP}$.} To test whether the truth signal is linearly recoverable or whether nonlinearity buys substantial AUROC, we train a one-hidden-layer MLP (128 ReLU units, $L_2$ regularisation matched to the logistic baseline) at the same best-textbook layer as $\mathcal{M}_\text{textbook}$, on the same training split, with early stopping on a 10\% within-train validation split carved off the textbook training data. Comparison to $\mathcal{M}_\text{textbook}$ at the same layer keeps everything but the probe architecture constant.

% \paragraph{Difference-of-means probe $\mathcal{M}_\text{diff}$.} Following \citet{marks2024geometry}, we compute the per-class hidden-state means on the textbook training split at the best-textbook layer and define the probe direction as the difference of those means. The decision boundary is the midpoint of the two projected class means. The probe has no learned weights and no hyperparameters to tune, and the comparison to $\mathcal{M}_\text{textbook}$ at the same layer tells us whether the $L_2$ regularisation of the logistic probe is itself contributing to or detracting from the signal.

% \paragraph{Permutation probe $\mathcal{M}_\text{perm}$.} As a sanity check that the probe is reading correctness signal rather than spurious structure in the activation space, we shuffle the correctness labels uniformly at random within the textbook training set and rerun the main-probe protocol. Test labels are unchanged.
\subsubsection{Output-only baselines}
\label{sec:setup:baselines}

To test whether the hidden state provides information beyond what the LLM's natural outputs already convey, we compare $\mathcal{M}_\text{textbook}$ against three baselines that score correctness from output-side signals alone. Token entropy ($\mathcal{B}_\text{ent}$) generates up to 20 tokens at temperature 0 and reports the mean per-token entropy of the softmax distribution, treating high generation uncertainty as a proxy for factual incorrectness. Self-consistency ($\mathcal{B}_\text{sc}$) draws three samples at $\tau{=}0.7$ and reports the fraction that agree with the deterministic ($\tau{=}0$) answer \citep{wang2023selfconsistency}. $P(\text{True})$ ($\mathcal{B}_\text{ptrue}$) follows \citet{kadavath2022language}, reading the first-generated-token logits $z$ from the same yes/no prompt used for $\mathcal{M}_\text{textbook}$ and computing the model's self-evaluation confidence as $\mathcal{B}_\text{ptrue} = \text{softmax}(z)_{\texttt{Yes}} \big/ \bigl(\text{softmax}(z)_{\texttt{Yes}} + \text{softmax}(z)_{\texttt{No}}\bigr)$.

\subsubsection{Calibration}
\label{sec:setup:calibration}

To test whether raw probe scores can be read as probabilities of correctness, we split the textbook 20\% held-out set into a 10\% calibration half and a 10\% final-test half. The two post-hoc calibrators of \S\ref{sec:setup:eval}, Platt scaling and isotonic regression, are fit on the calibration half. We then report raw, Platt-scaled, and isotonic-scaled ECE on the final-test half. The same protocol runs for each of the four LLMs across all four registers, yielding 16 (LLM, register) combinations in total.
\section{Results and Discussion}
\label{sec:results}

\old{We present our findings in two parts. First, we isolate the transferability of the truth direction across our three axes: stylistic register (\S\ref{sec:transfer_register}), medical specialty (\S\ref{sec:transfer_specialty}), and dataset corpus (\S\ref{sec:transfer_corpus}). Second, we evaluate the probe's architectural and deployment properties by testing the linearity of the learned direction (\S\ref{sec:linearity}), comparing it against standard output-only baselines (\S\ref{sec:baselines}), and assessing its calibration (\S\ref{sec:calibration}); a qualitative analysis of recurring high-confidence error patterns is reported in Appendix~\ref{app:errors}.}\new{We first isolate transfer along the three proposed axes (\S\ref{sec:transfer_register}--\S\ref{sec:transfer_corpus}), then test for linearity (\S\ref{sec:linearity}), compare to output-only baselines (\S\ref{sec:baselines}), and evaluate model calibration (\S\ref{sec:calibration}). High-confidence error patterns are analysed in Appendix~\ref{app:errors}.} Across all tables and figures, a negative $\Delta$ indicates a drop in AUROC relative to the textbook baseline.
% Table~\ref{tab:headline} reports the per-LLM transfer gaps for $\mathcal{M}_\text{textbook}$ on the three shifts of \S\ref{sec:methodology:shifts}, alongside the diff-means probe $\mathcal{M}_\text{diff}$ at the same layer. The rest of this section unpacks each shift in turn.
% We chose these three axes because they are what simultaneously shift when one moves between medical-QA datasets, as in \citet{orgad2025llms}'s cross-dataset fragility finding. Other axes such as model scale, prompt template, language are out of scope and discussed in Limitations. Results are organised by axis: surface form (\S\ref{sec:transfer_register}), sub-domain (\S\ref{sec:transfer_specialty}), corpus (\S\ref{sec:transfer_corpus}), probe-design ablations (\S\ref{sec:linearity}), and three subsections on the divergence between internal and external confidence (\S\ref{sec:baselines}, \S\ref{sec:calibration}, \S\ref{sec:failure_modes}). Table~\ref{tab:headline} reports the per-model AUROC drop along each axis.
% Results are organized by axis: surface form (\S\ref{sec:transfer_register}), sub-domain (\S\ref{sec:transfer_specialty}), corpus (\S\ref{sec:transfer_corpus}), linear readability (\S\ref{sec:linearity}), and the divergence between the probe's internal estimate and the model's external confidence (\S\ref{sec:internal-external}). Table~\ref{tab:headline} reports the per-model AUROC drop along each axis.

\begin{table*}[t]
\centering
\footnotesize
\setlength{\tabcolsep}{4pt}
% Threshold-metric columns added (Acc@0.5, Acc@Youden on the textbook held-out split; F1 in the appendix);
% the diff-means "R_clin gain" column moved to Appendix~\ref{app:diff-means} (Table~\ref{tab:diff-means-full}).
\begin{tabular}{@{}lc cccc ccc c@{}}
\toprule
& & \multicolumn{4}{c}{\textbf{In-distribution} ($\Rtext$)} & \multicolumn{3}{c}{\textbf{Transfer gaps} ($\mathcal{M}_\text{textbook}$)} & \textbf{Diff-means} \\
\cmidrule(lr){3-6} \cmidrule(lr){7-9} \cmidrule(l){10-10}
& \textbf{Best} & & \new{\textbf{Acc}} & \new{\textbf{F1}} & \new{\textbf{Acc}} & \textbf{Register} & \textbf{Specialty} & \textbf{Corpus} & \textbf{Register} \\
\textbf{LLM} & \textbf{layer} & \textbf{AUROC} & \new{\textbf{@0.5}} & \new{\textbf{@0.5}} & \new{\textbf{@Youden}} & $\boldsymbol{\Delta_\text{register}}$ & $\boldsymbol{\Delta_\text{specialty}}$ & $\boldsymbol{\Delta_\text{dataset}}$ & $\boldsymbol{\Delta_\text{register}^\text{diff}}$ \\
\midrule
Gemma-2-2B-it       & 15 & .733 [.67,.79] & \new{.640} & \new{.625} & \new{.685} & $-$.064 & $-$.053 & $-$.22 & $-$.045 \\
Gemma-3-4B-it       & 25 & .768 [.71,.83] & \new{.695} & \new{.684} & \new{.715} & $-$.085 & $-$.010 & $-$.21 & $-$.079 \\
Qwen2.5-7B-Instruct & 23 & .795 [.74,.85] & \new{.735} & \new{.720} & \new{.765} & $-$\replace{.124}{.126} & $-$.016 & $-$.23 & $-$.073 \\
Llama-3-8B-Instruct & 29 & .802 [.74,.86] & \new{.740} & \new{.740} & \new{.765} & $-$.104 & $-$.045 & $-$.20 & $-$.101 \\
\midrule
\textbf{Mean} & --- & \textbf{.775} & \new{\textbf{.703}} & \new{\textbf{.692}} & \new{\textbf{.733}} & \textbf{$-$.095} & \textbf{$-$.031} & \textbf{$-$.21} & \textbf{$-$.075} \\
\bottomrule
\end{tabular}
\caption{In-distribution performance and transfer gaps for $\mathcal{M}_\text{textbook}$ on the three shifts of \S\ref{sec:methodology:shifts}, with $\mathcal{M}_\text{diff}$ \citep{marks2024geometry} at the same layer. All $\Delta$ are AUROC drops on the held-out side; AUROC carries 95\% bootstrap CIs. All values are on the same held-out 20\% fact split ($n{=}200$); layer-selection sensitivity in Appendix~\ref{app:layer-rules}, per-register threshold metrics in Appendix~\ref{app:thresholds}, diff-means detail in Table~\ref{tab:diff-means-full}, Appendix~\ref{app:diff-means}.}
\label{tab:headline}
\end{table*}

\subsection{Register transfer}
\label{sec:transfer_register}

\begin{figure}[t]
\centering
\includegraphics[width=\columnwidth]{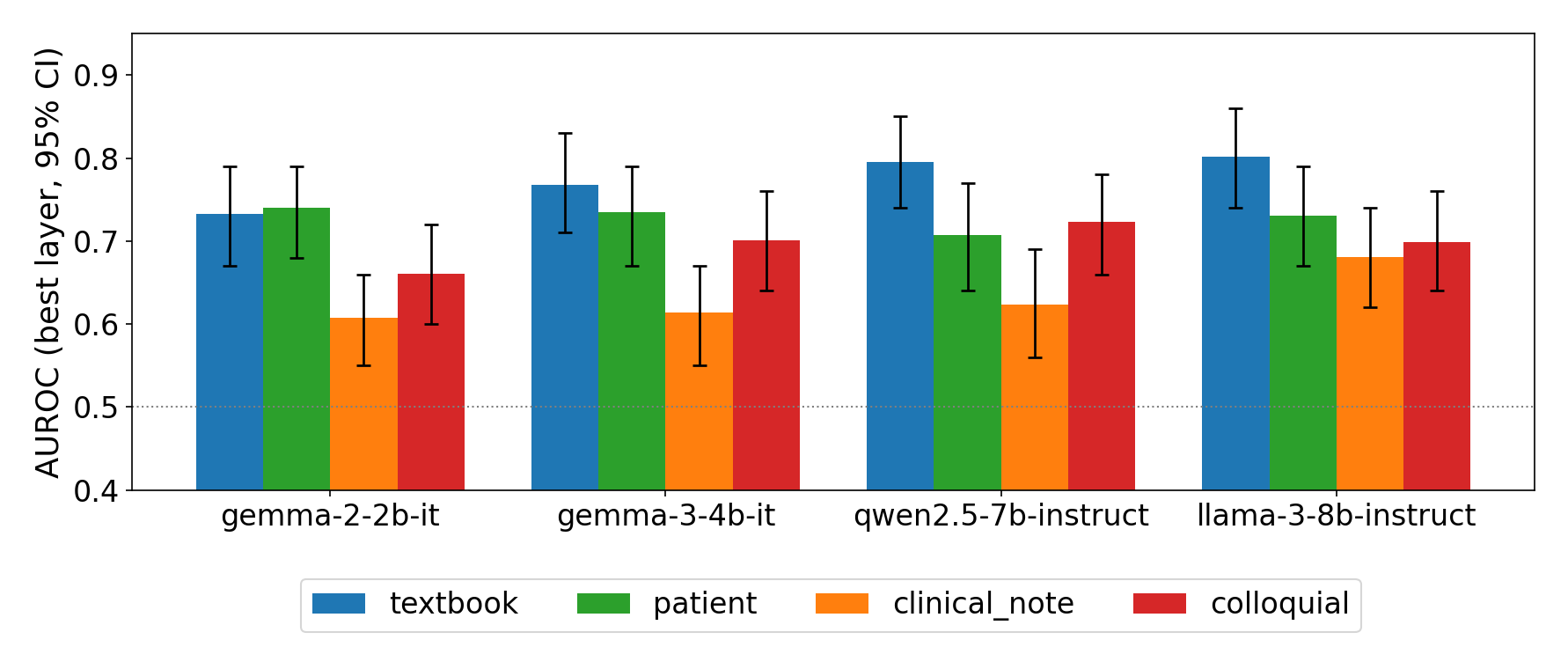}
\caption{Per-register best-layer AUROC with 95\% bootstrap CIs. $\Rtext$ is the training register; mean drop across the other three registers is 0.095 AUROC, with the largest on $\Rclin$. Per-register values in Table~\ref{tab:main}, Appendix~\ref{app:main-register-table}.}
\label{fig:register-bars}
\end{figure}

The truth direction learned on textbook activations transfers to the other three registers with a mean drop of $-0.095$ AUROC on the same held-out facts as textbook (Figure~\ref{fig:register-bars}; full per-register numbers in Table~\ref{tab:main}, Appendix~\ref{app:main-register-table}). Per-model register gaps range from $-0.064$ to $-0.126$ across the four LLMs and do not follow a clean scaling pattern\old{, with the smallest gap on Gemma-2-2B and the largest on Qwen2.5-7B}. $R_\text{pat}$ is the easiest to transfer to in three of four LLMs and $R_\text{clin}$ the hardest in all four. A label-permutation probe ($\mathcal{M}_\text{perm}$) collapses to mean AUROC of $0.488$\new{ (averaged over all layers)}, confirming the signal we measure is correctness-driven rather than spurious activation structure (Appendix~\ref{app:permutation}). \old{The gap is robust to how layers are selected: under a single best-mean layer across registers it is $0.080$, and under a strictly textbook-blind layer (the zero-shot upper bound) it is $0.160$; both remain below the cross-corpus drop reported in \S\ref{sec:transfer_corpus} (Appendix~\ref{app:layer-rules}).}\new{The gap holds under three layer-selection rules ($0.080$ to $0.160$; Appendix~\ref{app:layer-rules}), always below the cross-corpus drop of \S\ref{sec:transfer_corpus}.}

\new{Regenerating the full benchmark with a second generator from a different family (Gemini 3 Flash) gives a mean register gap of $0.079$ AUROC against $0.095$ for Sonnet, producing the same per-register ordering (details in App.~\ref{app:generator-ablation}). Dropping every fact whose wrong-answer variant falls below a judged factual-fidelity threshold moves the mean gap by at most $0.0023$ AUROC (see App.~\ref{app:fidelity-filter}).}

This gap is modest given how much the surface form actually shifts: $R_\text{clin}$ and $R_\text{col}$ variants sit 6--7$\times$ above the native MedQA perplexity under a reference LM (Gemma-2-2B-it; Appendix~\ref{app:perplexity}), far beyond a typo or an inserted distractor sentence\replace{. Register ordering by mean perplexity matches register ordering by mean drop, locating any brittleness at the high-perplexity end of the axis. The surface-perturbation brittleness reported by \mbox{\citet{haller2025brittle}} and \mbox{\citet{schouten2025truthvalue}} therefore holds for some perturbation classes but not for broad stylistic variation captured by register.}{, and register ordering by perplexity matches ordering by drop in probe AUROC. The surface-perturbation brittleness of \citet{haller2025brittle} and \citet{schouten2025truthvalue} therefore holds for some perturbation classes but not for broad stylistic variation.}

The residual gap is partly coverage-driven and partly structural. Training the probe on a balanced mix of all four registers ($\mathcal{M}_\text{all}$) under a matched protocol (same layer, same held-out facts) recovers a mean of $+0.061$ AUROC on the three non-textbook registers, concentrated on the hardest register $R_\text{clin}$ for the two larger LLMs ($+0.122$ on Gemma-3-4B, $+0.127$ on Llama-3-8B), at a small textbook cost of $\le 0.03$ AUROC (Table~\ref{tab:mixed}, Appendix~\ref{app:register-matrix}). \old{Mixing therefore addresses the coverage component of the register gap, especially for the most stylistically distant register.}\new{Mixing helps the register that transfers worst without producing a better probe overall: on the four-register panel, a probe trained on all four registers ($0.729$) is within $0.004$ of one trained on the patient register alone ($0.728$), and textbook is a representative rather than a favourable training register (panel means $0.725$ to $0.755$ across training registers; Appendix~\ref{app:register-matrix}).} We additionally observe systematic rejection of correct sentences containing explicit negation, which replicates the negation-generalisation failure of \citet{Levinstein2024lie} and \citet{schouten2025truthvalue} and is detailed in Appendix~\ref{app:errors}.

\new{\paragraph{Human-written patient questions.} Applied without retraining to 84 patient-authored MedRedQA questions, the probe reaches AUROC $0.695$ [$0.646$, $0.750$] on Qwen2.5-7B and $0.680$ [$0.622$, $0.734$] on Llama-3-8B, comparable to the Sonnet-written patient and colloquial registers, but falls to $0.606$ on Gemma-3-4B and $0.542$ on Gemma-2-2B. The signal is therefore recoverable on genuine patient writing for the more capable models, so register robustness does not depend only on how Sonnet writes. MedRedQA is a forum corpus that differs from MedQA in more than register, so part of the drop on the smaller models is a corpus shift; we report construction details and other caveats are in Appendix~\ref{app:medredqa} and \S\ref{sec:limits}.}

\subsection{Specialty transfer}
\label{sec:transfer_specialty}

% CAMERA-READY: two-panel figure; panel (b) promoted from the appendix (kJP9 W2)
\begin{figure*}[t]
\centering
\begin{subfigure}[b]{0.31\textwidth}
\centering
\includegraphics[width=\linewidth]{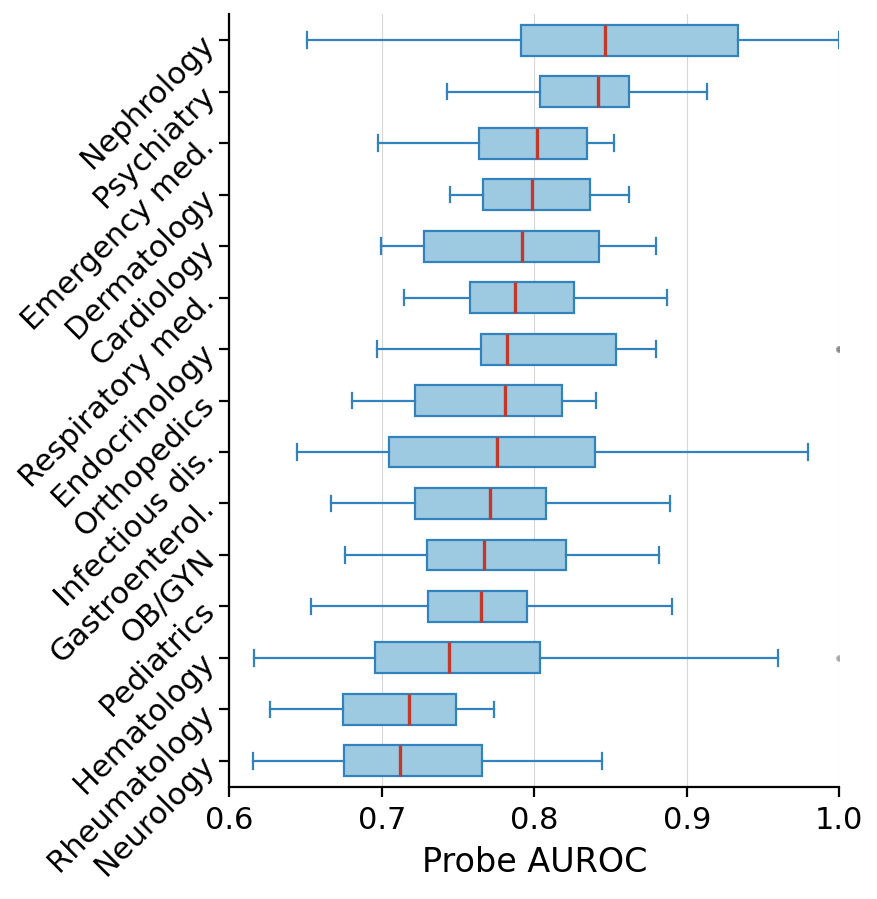}
\caption{Per-specialty AUROC.}
\label{fig:specialty-a}
\end{subfigure}\hfill
\begin{subfigure}[b]{0.67\textwidth}
\centering
% OLD: \includegraphics[width=\linewidth]{fig3_rarity_split.png}  (compact redraw: scripts/fig_rarity_compact.py)
\includegraphics[width=\linewidth]{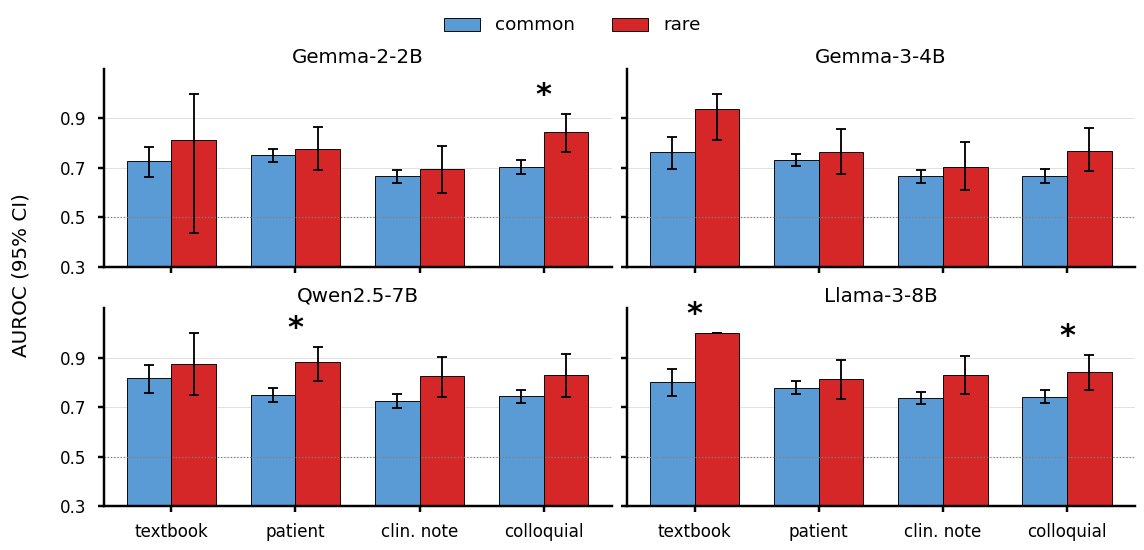}
\caption{\new{Common vs.\ rare disease AUROC per register.}}
\label{fig:specialty-b}
\end{subfigure}
\caption{\textbf{(a)} Probe AUROC across 15 medical specialties (351/500 facts, S-MedQA). Each box has one point per (LLM, register) condition. \replace{Median}{Mean} AUROC ranges from 0.71 (Rheumatology) to \replace{0.85}{0.86} (Nephrology). \new{\textbf{(b)} Common vs.\ rare AUROC per register with 95\% bootstrap confidence intervals (CIs); asterisks mark disjoint CIs. AUROC for rare diseases matches or exceeds common diseases in every condition.}}
\label{fig:specialty}
\end{figure*}

The truth direction generalises across medical sub-domains. Training the probe on textbook variants from a 7-specialty subset and testing on the 8 held-out specialties produces a mean drop of $\Delta_\text{specialty} = -0.031$ AUROC across five random partitions. This is smaller than the within-MedQA register gap and roughly seven times smaller than the cross-corpus drop detailed in \S\ref{sec:transfer_corpus}. This extends the topic-invariance \citet{marks2024geometry} reported for general factual statements to the medical domain: the linear direction the probe recovers is not specialty-specific, even though the medical content of each sub-domain differs substantially. \new{Probe AUROC on the 351 matched facts and on the 149 unmatched facts differs by at most $0.027$ in all four LLMs, with overlapping 95\% bootstrap CIs and no consistent direction (Table~\ref{tab:specialty-selection}, with a held-out-only replication in Appendix~\ref{app:specialty-selection}), so the matched subset is not an easier slice.}

% NEW (camera-ready): HZcj W4, placed in \S4.2 as promised in the response
\begin{table}[t]
\centering
\footnotesize
\setlength{\tabcolsep}{3pt}
\begin{tabular*}{\linewidth}{@{\extracolsep{\fill}}lcc@{}}
\toprule
\textbf{Model} & \textbf{Matched ($n{=}351$)} & \textbf{Unmatched ($n{=}149$)} \\
\midrule
Gemma-2B  & .709 [.686,.729] & .682 [.652,.711] \\
Gemma-3B  & .677 [.656,.698] & .671 [.636,.704] \\
Qwen-7B   & .738 [.716,.759] & .739 [.704,.770] \\
Llama-8B  & .741 [.718,.762] & .735 [.705,.764] \\
\bottomrule
\end{tabular*}
\caption{Probe AUROC on S-MedQA-matched vs.\ unmatched facts, pooled over non-textbook registers, per-cell best layer. CIs overlap in every row.}
\label{tab:specialty-selection}
\end{table}

\paragraph{Per-specialty variation.} This small transfer gap should not be mistaken for a uniform difficulty profile. Pooling all (LLM, register) probes by specialty, the mean AUROC ranges from \replace{$0.713$}{$0.710$} (Rheumatology) to \replace{$0.851$}{$0.859$} (Nephrology), a \replace{14}{15}-point spread (Figure~\ref{fig:specialty}). High-performing specialties have highly distinctive diagnostic vocabulary (e.g., Nephrology with BUN, creatinine, GFR), while low-performing specialties share overlapping symptom vocabulary (Neurology and Rheumatology both present with fatigue, joint pain, and weakness). The \replace{14}{15}-point spread is descriptive (absolute difficulty \replace{changes based on where a fact lives}{varies by specialty}) while the $\Delta_\text{specialty}$ of $-0.031$ is controlled (the same \replace{geometric }{}direction transfers across specialty sets). 
% Deployments should therefore report AUROC per specialty rather than a single dataset-level number, even if the underlying probe does not require per-specialty retraining.

\paragraph{Rare diseases.} Contrary to expectation, rare-disease AUROC meets or exceeds common-disease AUROC in all 16 (LLM, register) conditions, with \replace{three}{four} cells showing strictly disjoint CIs favouring rare diseases. \old{Full per-cell numbers and a hypothesised mechanism (distinctive, low-frequency vocabulary activating isolated regions of representation space) are provided in Appendix~\ref{app:rarity-full}.}\new{The largest gains are on the larger LLMs, e.g.\ Qwen2.5-7B on $\Rpat$ ($0.880$ rare vs.\ $0.748$ common). A plausible mechanism is that distinctive, low-frequency vocabulary activates isolated regions of representation space where the correct/wrong contrast separates more easily; the tagger's limits are discussed in \S\ref{sec:limits} and per-cell numbers are in Appendix~\ref{app:rarity-full}.}

\subsection{Corpus transfer}
\label{sec:transfer_corpus}

\begin{figure}[t]
\centering
% CAMERA-READY: figure regenerated (scripts/fig14_ladder.py) with the MMLU-medical and reformatted-MedMCQA bars.
% OLD: \includegraphics[width=\columnwidth]{fig14_register_vs_dataset.png}
\includegraphics[width=\columnwidth]{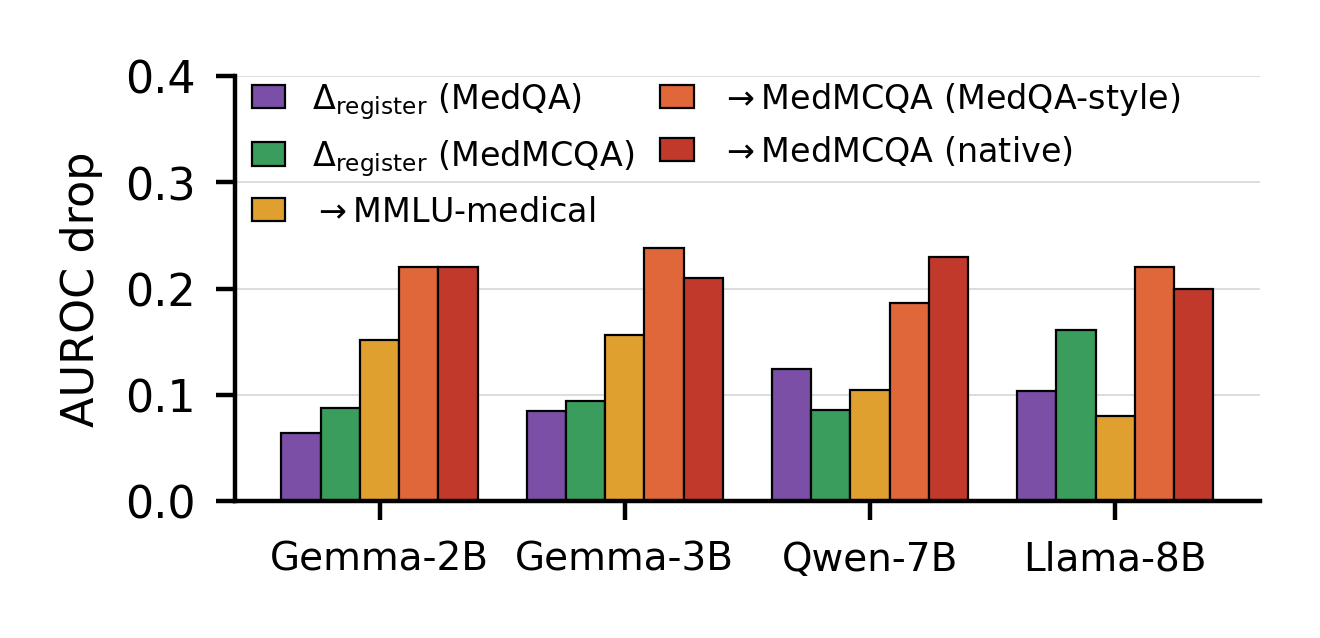}
\caption{The three shifts side by side, per LLM. Purple: within-MedQA register $\Delta_\text{register}$ ($n{=}500$). Green: within-MedMCQA register $\Delta_\text{register}$ ($n{=}100$). Red: MedQA-to-MedMCQA $\Delta_\text{dataset}$\new{; yellow and orange: the same probe on MMLU-medical and on MedMCQA reformatted into MedQA style}.}
\label{fig:register_vs_dataset}
\end{figure}

Applying the MedQA-trained probe to MedMCQA without retraining drops AUROC by a mean of $\Delta_\text{dataset} = -0.21$, roughly twice the within-MedQA register gap and consistent across LLMs (Table~\ref{tab:headline}). MedMCQA absolute AUROC sits between $0.52$ and $0.60$. \old{The drop is symmetric: a probe retrained on MedMCQA-textbook suffers a similar drop when transferred to MedQA-textbook, indicating the failure is mutual rather than a limitation specific to MedQA (Appendix~\ref{app:probe-results}).}\new{The drop is symmetric. A probe trained on native MedMCQA transfers back to MedQA at only $0.607$, so the failure is mutual and not specific to MedQA (Appendix~\ref{app:corpus-ladder}).} \old{This extends \citet{orgad2025llms}'s cross-dataset fragility from general factual QA into the medical-exam setting, demonstrating that the same breakage holds even when both source and target datasets are within the same task family.}

 % inside MedMCQA on matched held-out facts.(SME-authored MCQ distractors in MedMCQA versus rewritten prose in our $R_\text{textbook}$)Specialty-disjoint training inside MedQA costs only $-0.031$ AUROC (\S\ref{sec:transfer_specialty}), and a within-MedMCQA register replication on the 100-fact subset rewritten into the same four registers (\S\ref{sec:setup:corpus}) yields a register gap of $-0.107$ AUROC.

\paragraph{Register invariance generalises.} The within-MedMCQA register replication confirms that register robustness is not an artefact of MedQA: the mean within-MedMCQA register drop of $-0.107$ AUROC is similar to the within-MedQA gap of $-0.095$ and far smaller than the cross-corpus drop\old{. The truth direction is therefore register-stable inside each corpus and breaks only when the corpus itself changes}.

% CAMERA-READY: the corpus-ladder table moved to Appendix~\ref{app:corpus-ladder} (Table~\ref{tab:corpus-ladder-full}); the numbers are in the text below and in Figure~\ref{fig:register_vs_dataset}.

\paragraph{\replace{What explains the drop.}{Sources of the drop.}} \old{The cross-corpus drop is not attributable to differences in specialty distribution or register conventions between the two corpora. Even a worst-case additive combination of register and specialty effects across both corpora ($-0.095$ to $-0.149$ AUROC) falls below the $-0.21$ cross-corpus drop (Figure~\ref{fig:register_vs_dataset}). What remains is structural: the two corpora differ in question format, in distractor authorship, and in the adversarial pressure those distractors exert on the probe. We do not attempt a precise quantitative decomposition of these factors since register, specialty, and item construction interact non-linearly in representation space.}\new{Cross-corpus failure is uneven (Figure~\ref{fig:register_vs_dataset}, per-model values in Appendix~\ref{app:corpus-ladder}). The same probe loses $0.21$ AUROC on MedMCQA and $0.12$ on MMLU-medical, a third corpus with the same four-option format and expert-authored distractors. A change of corpus on its own therefore does not break the probe, and the larger drop has a cause specific to MedMCQA. Question format is excluded as that cause, because reformatting 100 MedMCQA items into MedQA-style vignettes with the answers held fixed leaves AUROC at $0.559$ against $0.561$ for native MedMCQA, with overlapping CIs. Topic and register are excluded as well, because even a worst-case additive combination of the specialty and register effects ($0.095$ to $0.149$ AUROC) falls below the MedMCQA drop (Figure~\ref{fig:register_vs_dataset}). By elimination, the residual sits with how MedMCQA's items are constructed, most plausibly distractor authorship and the adversarial pressure those distractors exert.}\old{We do not isolate which property is responsible and do not measure distractor adversarialness directly.}\new{ Distance from the training distribution does not predict failure either, since MedRedQA (\S\ref{sec:transfer_register}) sits further from MedQA than either MMLU or MedMCQA on any surface measure, yet the signal survives on the larger LLMs.}

\subsection{Linearity}
\label{sec:linearity}

The truth signal is dominantly linear. A one-hidden-layer MLP probe ($\mathcal{M}_\text{MLP}$) trained on the same textbook split and at the same best-textbook layer as $\mathcal{M}_\text{textbook}$ gains only $+0.037$ AUROC on average. In 15 of 16 (LLM, register) conditions, the MLP probe falls inside the bootstrap-CI overlap of the baseline logistic probe (Appendix~\ref{app:probe-ablations}). A nonlinear classifier, therefore, extracts no meaningful additional signal beyond what a linear hyperplane captures.

The unregularised difference-of-means probe ($\mathcal{M}_\text{diff}$, \citealp{marks2024geometry}) matches the logistic probe on mean register drop. By recovering the signal without learned weights, this confirms that the geometric-truth hypothesis of \citet{marks2024geometry} extends to the medical domain\old{: the truth direction exists as a structural property of the representation space}. $\mathcal{M}_\text{diff}$ is uniformly better than $\mathcal{M}_\text{textbook}$ on the hardest register, $R_\text{clin}$ (mean gain $+0.070$ AUROC, \replace{Table~\ref{tab:headline}}{Table~\ref{tab:diff-means-full}, Appendix~\ref{app:diff-means}}). We interpret this $R_\text{clin}$ advantage as evidence that $L_2$ regularisation in the logistic probe overweights the textbook training distribution, actively hurting generalisation to the most stylistically distant register.

\subsection{Output-only baselines}
\label{sec:baselines}

\begin{figure}[!htbp]
\centering
\includegraphics[width=\columnwidth]{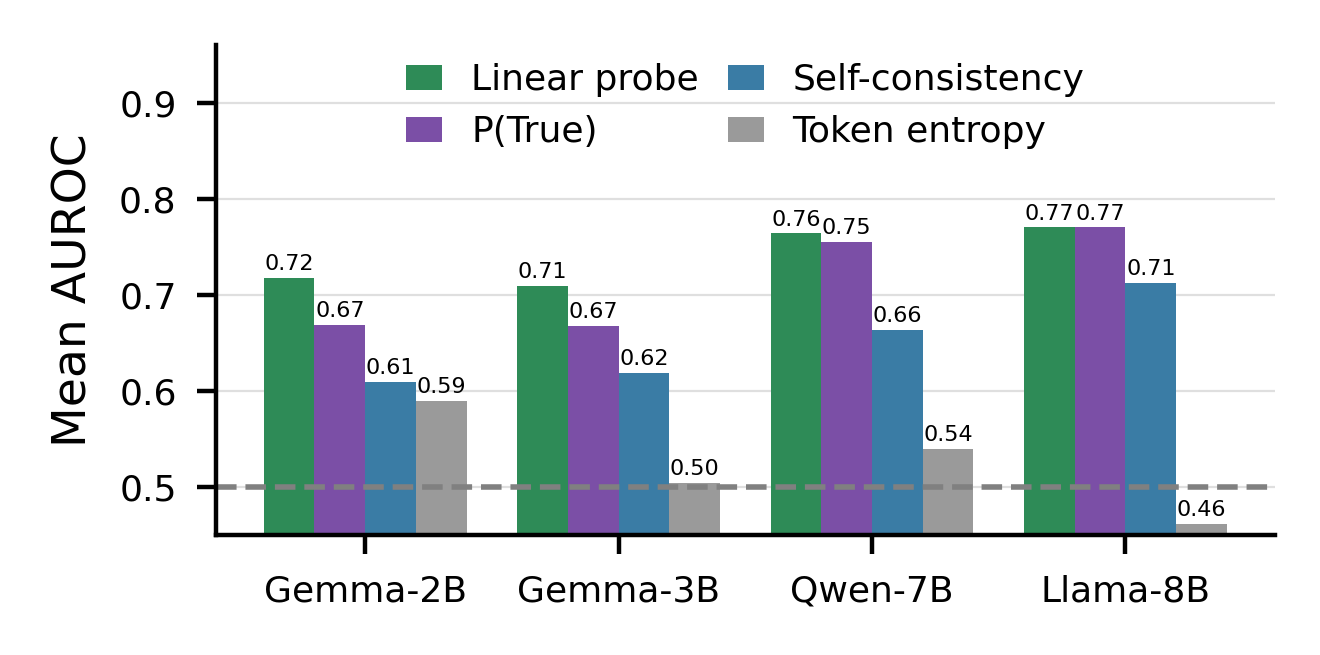}
\caption{Mean AUROC across registers, per LLM, for $\mathcal{M}_\text{textbook}$ (green) vs.\ output-only baselines: P(True) (purple), self-consistency at $\tau{=}0.7$ (blue), and token entropy (grey). Per-register breakdowns in Appendix~\ref{app:baselines}.}
\label{fig:output-baselines}
\end{figure}

The probe carries discriminative signal well beyond standard output-side confidence measures, though a strong self-evaluation baseline closes the gap on the larger LLMs. Across registers and LLMs, $\mathcal{M}_\text{textbook}$ outperforms $\mathcal{B}_\text{sc}$ by 6--11 AUROC points and $\mathcal{B}_\text{ent}$ by 13--31 AUROC points (Figure~\ref{fig:output-baselines}). $\mathcal{B}_\text{ent}$ hovers near or below chance, falling to 0.46 for Llama-3-8B.

$\mathcal{B}_\text{ptrue}$ is the strongest output-only baseline (mean across registers $0.67$--$0.77$). The probe exceeds $\mathcal{B}_\text{ptrue}$ by $+0.05$ AUROC on Gemma-2-2B and $+0.04$ on Gemma-3-4B, but \replace{ties on Qwen2.5-7B and Llama-3-8B (within $\pm 0.002$)}{ties on Llama-3-8B and is within $0.01$ on Qwen2.5-7B}. On these larger LLMs $\mathcal{B}_\text{ptrue}$ also has the tighter register profile ($\Delta_\text{register}$ of $0.063$ vs.\ $0.095$ for the probe; Appendix~\ref{app:baselines}), so the probe's remaining advantage is inference-side\replace{ as it reads off the hidden state at the question position in a single forward pass, before any answer token is generated, bypassing the latency and cost of generative sampling}{: it reads the hidden state at the question position in a single forward pass, before any answer token is generated}.

% \subsection{Probes vs.\ output-only baselines}
% \label{sec:baselines}

% $\mathcal{M}_\text{textbook}$ carries a discriminative signal that the LLM's weaker externally observable confidence signals do not, but a strong self-evaluation baseline matches it. The probe beats self-consistency $\mathcal{B}_\text{sc}$ by 6--11 AUROC points and token entropy $\mathcal{B}_\text{ent}$ by 13--31 points in mean across registers for every LLM (Figure~\ref{fig:output-baselines}); $\mathcal{B}_\text{ent}$ hovers near or below chance, falling to 0.46 for Llama-3-8B. The strongest output-only baseline is $\mathcal{B}_\text{ptrue}$ (mean 0.67--0.77), not self-consistency: the probe exceeds it by only $+0.05$ and $+0.04$ AUROC on Gemma-2-2B and Gemma-3-4B and \emph{ties} it on Qwen2.5-7B and Llama-3-8B (within $\pm 0.002$), with $\mathcal{B}_\text{ptrue}$ slightly exceeding the probe on the textbook register alone for the two larger LLMs. The probe's value over $\mathcal{B}_\text{ptrue}$ on these LLMs is therefore not raw discriminative power but pre-generation availability (a single forward pass at the question position, before any answer token is generated) and a marginally tighter register profile ($\Delta_\text{register} = 0.052$ vs.\ $0.060$ for $\mathcal{B}_\text{ptrue}$; Appendix~\ref{app:baselines}).
\subsection{Calibration}
\label{sec:calibration}

Raw probe outputs are not well calibrated. The mean \new{raw} ECE across all 16 (LLM, register) conditions is \replace{$0.358$}{$0.341$}\new{ on the calibration split (Table~\ref{tab:platt}, Appendix~\ref{app:calibration})}, with \replace{every condition sitting above $0.20$ (Figure~\ref{fig:ece}, Appendix~\ref{app:calibration})}{no condition below $0.19$ on the full held-out set (Figure~\ref{fig:ece})}. Platt scaling cuts the mean ECE \replace{roughly in half to $0.157$}{by more than half to $0.135$}, while isotonic regression yields smaller gains (\replace{$0.252$}{$0.234$})\replace{, likely due to its tendency to overfit at this calibration sample size}{, likely from overfitting at this sample size}. Even after Platt scaling, \replace{no condition reaches}{all but one of the 16 conditions stay above} the $0.05$ threshold conventionally considered well-calibrated. The probe ranks correct statements above incorrect ones (high AUROC), but its raw outputs cannot be read as probabilities of correctness. \old{This pattern aligns with \mbox{\citet{berkowitz2025ping}}, who report up to a 96\% ECE reduction on MedMCQA by training the probe \emph{jointly} for prediction and calibration. Our 56\% Platt reduction confirms their finding that robust calibration is best treated as a fundamental probe-design decision rather than a post-hoc patch.}\new{Our 60\% Platt reduction agrees with \citet{berkowitz2025ping} (\S\ref{sec:related}) that calibration should be built into the probe design. The Youden-optimal cutoff likewise swings between $0.04$ and $1.00$ across registers for one LLM (Appendix~\ref{app:thresholds}).} A qualitative inspection of the highest-confidence probe errors surfaces three recurring patterns (calibration collapse, plausible-but-wrong acceptance, and negation rejection), reported in Appendix~\ref{app:errors}.

\section{Related Work}
\label{sec:related}

\paragraph{Linear truth: evidence and skepticism.} \citet[SAPLMA;][]{azaria2023internal} and \citet[CCS;][]{burns2023discovering} first showed that internal LLM states encode truthfulness. \citet{marks2024geometry} demonstrated linear separability and causal intervenability of a truth direction,
%at the model sizes available in 2024,
including the result that a difference-of-means probe matches a learned classifier; \citet{burger2024truth} extended this to a 2-D subspace robust to negation.
Contrary to these findings, \citet{Levinstein2024lie} argue that truth probes face fundamental roadblocks (negation, paraphrasing, generalisation), and \citet{haller2025brittle} report systematic brittleness under small surface perturbations.
Our main findings sit between the two camps: we confirm negation and corpus shift as real failure modes, but show that register leaves the linear truth direction largely intact\old{, with a modest gap that mixing-based training partly addresses}.
%Our findings sit between the two camps. We confirm negation as a real failure mode and confirm large cross-corpus drops, but show that register, a substantially larger surface change than a typo, leaves the linear truth direction nearly intact. The reconciliation runs through perplexity (Section~\ref{sec:transfer_register}, Appendix~\ref{app:perplexity}): the register variants on which the probe survives sit largely inside the native-MedQA perplexity distribution, whereas MedMCQA text sits more than an order of magnitude higher in mean perplexity, consistent with the brittleness picture on the genuinely Out-Of-Distribution (OOD) end of the axis. Section~\ref{sec:linearity} replicates the Marks \& Tegmark linearity and diff-means claims in the medical domain.

% --- Context and dynamics paragraph removed (supervisor: tangential to the three-axis story).
% \paragraph{Context and dynamics.}
% \citet{schouten2025truthvalue}
% %\citet{schouten2025truthvalue}
% report that truth directions are context-sensitive: probes shift when irrelevant context is inserted. \citet{kossen2024semantic} with semantic entropy probes, \citet{bhatnagar2026halt} with DRIFT, and \citet{zhang2025icr} with the ICR Probe, read uncertainty signals from intermediate layers or hidden state dynamics rather than static snapshots, and \citet{farquhar2024detecting} apply semantic entropy at deployment. We sharpen the context-sensitivity finding by showing that register, a structured kind of context shift larger than an inserted distractor sentence, induces only modest drift.

\paragraph{Task specificity.} \citet{orgad2025llms} report 
%skill-specific truthfulness encoding: 
that
probes transfer within a task family but fail across task families. \citet{binkowski2025spectral} propose a complementary attention-map spectral-features probe, and \citet{obeso2025realtime} extend linear and LoRA probes to long-form generation\old{ with a long-form HealthBench evaluation}, finding that long-form-trained probes transfer to short-form QA but \replace{short-form training does not recover long-form performance}{the reverse is not true}.
We sharpen the task-specificity finding by separating it inside a single task (medical QA) into a register shift, a specialty shift, and a corpus shift, and showing that the corpus shift dominates the other two\new{, with the breakage appearing even between two exam corpora of the same task family}.

\paragraph{Medical probing and benchmarks.} \citet[PING;][]{berkowitz2025ping} and \citet{berkowitz2025adr} probe medical LLM hidden states for clinical knowledge recovery and adverse drug reaction detection\replace{ with strong calibration numbers (96\% ECE reduction).
We treat PING as the stronger calibration baseline rather than a competing approach. }{; PING trains the probe jointly for prediction and calibration and reports up to a 96\% ECE reduction on MedMCQA, and is a strong calibration reference.}
% \citet{ahsan2025bias} document mechanisms of demographic bias in healthcare LLMs. \citet{pal2023medhalt} (Med-HALT) and \citet{pandit2025medhallu} (MedHallu) provide medical hallucination benchmarks; both are graded for detection accuracy and do not support the controlled-rewriting intervention we need.

%=============================================================
\section{Conclusions and Future Work}
\label{sec:conclusion}

\replace{In this work, we investigated whether the linear truth direction in medical LLMs is a stable geometric property or a brittle artifact. By measuring probe transfer along three input shifts in isolation on matched held-out facts, we found}{We asked whether the linear truth direction in medical LLMs is a stable geometric property or a brittle artefact. Measuring probe transfer along three input shifts in isolation on matched held-out facts, we found} that the truth signal is largely robust to changes in linguistic style and medical sub-domain, but \old{breaks under a shift to a different corpus.}\new{degrades under corpus shift by an amount that depends on the target corpus, from $0.12$ AUROC on MMLU-medical to $0.21$ on MedMCQA, and the MedMCQA drop is not explained by question format.} This indicates the signal depends on how each corpus is built, not on medical content alone. \new{Within medicine, then, a truth direction does transfer across writing style, including human-written patient questions, across specialty, and to a variable extent across corpus.}
%The register gap is partly coverage-driven i.e
Moreover, we find that mixing all four registers at training time partially closes the register transfer gap.

% isolating dataset construction rather than surface form or sub-domain as the dominant cross-corpus barrier
% while no analogous augmentation closes the cross-corpus drop.
% Even the worst-case additive combination of register and specialty (0.095--0.149) sits well below the cross-corpus drop.
The signal is predominantly linear since a parameter-free difference-of-means probe \citep{marks2024geometry} matches L2-regularised logistic regression, extending their ``geometry-of-truth'' finding\old{ that the true and false statements separate linearly in hidden state space} to medical statements.
\replace{In addition, model representations carry information the model's outputs do not, with the probe outperforming entropy and self-consistency baselines, and matching a self-evaluation baseline.}{The probe outperforms entropy and self-consistency baselines and matches a self-evaluation baseline.}
Finally, 
%the one dimension where it falls short is calibration. Raw
raw probe scores are poorly calibrated, though post-hoc methods \replace{substantially reduce their expected calibration error}{halve their ECE}.

Our findings reconcile a major gap in the probing literature. The within-corpus robustness validates geometric-truth proponents \citep{marks2024geometry,burger2024truth}\old{ by confirming that the signal is dominantly linear and survives surface-form perturbations}. The across-corpus breakdown is consistent with the brittleness skeptics \citep{Levinstein2024lie,haller2025brittle} and refines the cross-dataset fragility from \citet{orgad2025llms}.
%, identifying what drives it in the medical QA context:
%In medical QA, the truth direction is bound to the training corpus's item format and distractor plausibility, rather than to medical knowledge as such.
%We decompose source of degradation medical QA.

 % extended here to medical statements under controlled stylistic variation
For clinical NLP practitioners, these findings establish strict boundaries for deployment of linear probes as \replace{mechanisms for factuality detection}{factuality detectors}. Because the cross-corpus drop is several times larger than stylistic and specialty shifts, probes must be treated as within-distribution safety tools and not deployed zero-shot to a new \replace{QA format}{corpus} without retraining. \replace{Second, because raw probe outputs suffer from severe miscalibration, post-hoc calibration is a mandatory safety layer. Additionally, per-specialty}{Because raw probe outputs are severely miscalibrated, post-hoc calibration is a mandatory safety layer, and per-specialty} AUROC should be reported alongside dataset-level numbers, since absolute probe difficulty varies substantially across clinical specialties\old{, even where the transfer direction is stable}.

The three-axis decomposition introduced here applies beyond multiple-choice QA, and extending it to long-form generation, real clinical text, \replace{other task families, and larger model scales}{and larger models} is a clear next step\old{ for understanding where hidden-state probes carry reliable factuality signal}. \old{A complementary direction is whether the partial recovery from mixed-register training extends to mixed-corpus training.} \new{Two other directions worth exploring are isolating which property of item construction drives the cross-corpus drop, and testing whether a two-dimensional truth subspace with a separate polarity direction \citep{burger2024truth} repairs the negation failure we document.}

\section{Limitations}
\label{sec:limits}

\new{Every corpus we use is exam-style multiple choice, and every wrong answer is either an expert-authored distractor or a controlled rewrite of one. Our claims are therefore about exam-format medical QA. Whether a linear truth direction behaves the same way on free-form clinical generation, where errors can be partially correct, context-dependent, or long, is not tested here and is the main direction of our follow-up work. The MedRedQA slice is small ($n{=}84$) and differs from MedQA on more than register, so it establishes that the signal survives on human-written patient text without quantifying the register effect in isolation on that text.}
The register-rewriting pipeline relies on a single primary generator (Claude Sonnet 4.5), itself an instruction-tuned LLM with stylistic preferences in what counts as a given register; the Gemini \replace{sub-replication}{replication} in Appendix~\ref{app:dataset} \replace{reduces but does not eliminate this concern}{and the fidelity filtering in Appendix~\ref{app:fidelity-filter} reduce but do not eliminate this concern}. \old{Both source corpora are English exam corpora, so the cross-corpus comparison confounds corpus with exam region, and whether the same picture holds on non-exam clinical text or in other languages is open.}\new{All corpora are English exam corpora, so the cross-corpus comparison confounds corpus with exam region, and whether the same picture holds in other languages is an open question.} The specialty-transfer measurement depends on the S-MedQA join, which covered 351 of 500 facts (70\%); the unmatched 149 facts are excluded from $\Delta_\text{specialty}$ but participate in every other measurement. The rarity tagger is a curated keyword classifier and misses conditions described by syndrome rather than by name, so the rare-disease finding should be read directionally rather than as a precise effect-size estimate.\new{ For the cross-corpus drop, we identify item construction as the residual source only by elimination: we do not isolate which construction property (distractor authorship, adversarial pressure, or another) is responsible, and we do not measure distractor adversarialness directly.}

Our probes use the last-question-token position; \citet{orgad2025llms} reports that exact-answer tokens carry more truth signal in generation-mode probes, which our yes/no protocol largely sidesteps but does not eliminate as a probe-design choice. Our four LLMs span 2--8B; \citet{obeso2025realtime} report linear probes on Llama-3.3-70B in long-form generation, and both the linearity and brittleness pictures may shift at that scale. Relatedly, the P(True) self-evaluation baseline matches $\mathcal{M}_\text{textbook}$ on raw AUROC\new{ to within $0.01$} for our two larger LLMs (Qwen2.5-7B and Llama-3-8B), and whether the AUROC gap reopens at larger scales is open. Our wrong answers are also SME-authored MCQ distractors, plausible-but-wrong by design; how these relate to the spontaneous fabrications a deployed LLM produces is an open empirical question, and probe performance in that distributional setting is not directly measured here.

%=============================================================
\section*{Ethics Statement}
The dataset derives from MedQA (USMLE exam items), a publicly released benchmark. No patient data is involved\new{ beyond the 84 publicly posted, already de-identified MedRedQA forum questions, which we use for evaluation only and do not redistribute}. The colloquial and patient-facing registers are model-generated stylistic rewrites, not representations of any real patient's speech. A probe with AUROC in the 0.70--0.80 range is not sufficient for patient-facing safety monitoring, and misapplication could produce false assurance about LLM reliability in clinical contexts.
\paragraph{License} MedQA, MedMCQA, S-MedQA, \new{MMLU, MedRedQA,} and the four LLMs we probe are released under licenses permitting research use\replace{;}{.} \old{we will release our 4,000-variant benchmark and code under a permissive research license (CC BY-NC 4.0 or equivalent)}\new{Our 4{,}000-variant benchmark, the evaluation-only variant sets, prompts, judge rubric, and the full pipeline code are released at \url{https://github.com/mnishant2/MedProbe_release} under a permissive research license (code under MIT, data under CC BY-NC 4.0).}
\paragraph{Use of AI assistants.} \old{Claude Sonnet 4.5 (with Gemini 3 Flash Preview as a robustness check) served as the register-rewriting generator that produced our benchmark, and Grok 4.1 Fast and GPT-5 Nano served as LLM judges; full prompts and the rubric are released (Appendix~\ref{app:prompt-figures}). AI assistants were also used during paper writing for editing, proofreading, and LaTeX formatting. All research design, experimental decisions, implementation, analyses, and conclusions are the authors' own.}\new{AI systems appear in this work in three distinct roles, all described in the methods. The four probed LLMs are the objects of study. Claude Sonnet 4.5, with Gemini 3 Flash Preview as a cross-family replication, generated the register rewrites. Grok 4.1 Fast (Grok 4.3 for the fidelity filtering of Appendix~\ref{app:fidelity-filter}, after 4.1 Fast was deprecated) and GPT-5 Nano acted as judges, and all prompts and the rubric are released (Appendix~\ref{app:prompt-figures}). Separately, AI assistants were used for writing and editing support: proofreading, LaTeX formatting, and copy-editing. They were not used for research design, experiment design, analysis, or interpretation, which are the authors' own.}

% TODO ACKNOWLEDGMENTS: funding, compute grants (e.g. Snellius / SURF), colleagues. Delete the section if not needed.
\section*{Acknowledgments}
\new{This publication is part of the project CaRe-NLP with file number NGF.1607.22.014 of the research programme AiNed Fellowship Grants, which is (partly) financed by the Dutch Research Council (NWO).}

\bibliography{custom}

%=============================================================
\appendix

\section{Extended preliminaries}
\label{app:preliminaries}

\replace{This appendix expands the definitions in Section~\ref{sec:methodology} for readers less familiar with linear-probing methodology.}{Some terms and notation used throughout the paper are defined here in more detail than Section~\ref{sec:methodology} allows.}

\paragraph{Linear probes.} A linear probe is a small supervised classifier (here, $L_2$-regularised logistic regression with $C{=}1.0$) trained to predict a label from a frozen LLM's hidden state $\hlayer$ at a chosen layer $\ell$ and token position $t$ \citep{alain2016understanding,azaria2023internal}. The LLM itself is not modified\replace{:}{, and} the probe reads activations off the side of a forward pass and outputs a scalar between 0 and 1 that we interpret as the model's internal estimate that a given (question, candidate-answer) pair is correct. We train probes on textbook-register activations only and evaluate them on held-out facts in every register, with no fine-tuning of the LLM. The probe's role is diagnostic\replace{, not generative: it asks}{. It asks} whether the LLM's hidden states linearly separate correct from incorrect answers, not whether the LLM itself produces correct answers.

\paragraph{Linguistic register.} A linguistic register is a stylistic configuration of vocabulary, syntax, and abbreviation density that the same content takes when produced for different audiences and channels \citep{biber1988variation,biber1995dimensions,stubbs-1997-book}. The register of a clinical textbook (formal, full sentences, Latinate vocabulary) differs sharply from a patient post on a health forum (first-person, colloquial), from a clinical note in an EHR (telegraphic, abbreviation-rich), and from an internet comment about a medical question (loose syntax, slang). All four can express the same fact. Section~\ref{sec:setup:data} explains why these specific four registers were chosen as the controlled-rewriting axis.

\old{\textbf{Metrics.} We report AUROC for ranking and ECE for calibration. AUROC (area under the receiver operating characteristic curve) is the probability that a randomly drawn correct-answer example receives a higher probe score than a randomly drawn incorrect-answer example; 0.5 is chance, 1.0 is perfect ranking. ECE (expected calibration error) is the mean gap between predicted probability and observed accuracy across probability bins; values below 0.05 are usually considered well calibrated, and values above 0.20 mean the probability outputs cannot be read as probabilities \mbox{\citep{guo2017calibration}}. AUROC is the standard ranking metric in the probing literature and supports the within-corpus triage framing in Section~\ref{sec:conclusion}; ECE is the standard calibration metric and is what governs whether a probe output can be reported to a downstream user as a probability of correctness.}

\new{Metrics (AUROC, ECE, and the threshold metrics) are defined in \S\ref{sec:setup:eval}.}

\paragraph{$\Delta$ notation.} For each register $r$ we report the register-transfer gap $\Delta_\text{register} = \text{AUROC}(\Dtest^{\Rtext}) - \text{AUROC}(\Dtest^{r})$, the AUROC drop from the textbook held-out set to register $r$. A value of zero would indicate a register-invariant probe. We use $\Delta_\text{specialty}$ for the analogous specialty-disjoint transfer drop (Section~\ref{sec:transfer_specialty}) and $\Delta_{\text{dataset}}$ for the cross-corpus MedQA-to-MedMCQA drop (Section~\ref{sec:transfer_corpus}). These three $\Delta$ axes correspond to the three transfer dimensions whose magnitudes the paper measures.

\paragraph{Confidence intervals.} All AUROC numbers carry a 95\% confidence interval from a 1{,}000-iteration fact-level bootstrap, resampling test \emph{facts} (not variants) with replacement so that both polarities of a fact enter or leave the sample together. This prevents the polarity pair from being treated as two independent observations, which would understate the CI width. Two intervals that do not overlap describe estimates that are statistically distinguishable at this sample size; we call such pairs \emph{disjoint}.

\paragraph{\new{Conditions.}} A single (model, register) pair (e.g.\ Llama-3-8B on $\Rclin$) is referred to as a \emph{condition}\replace{;}{, and} across four models and four registers we have 16 conditions.

\section{Dataset construction details}
\label{app:dataset}

\subsection{Rarity tagging}
\label{app:rarity}
Rarity is a keyword classifier over curated \textsc{common} and \textsc{rare} disease lists applied to the question and answer text. When both match, \textsc{rare} wins. Distribution: 455 common, 38 rare, 7 unknown. The tagger misses rare conditions referred to by syndrome description rather than name. Disease lists are released with the dataset.

\subsection{Generator selection}
\label{app:generator}
\old{\textbf{Pilot.} 50 facts $\times$ 4 registers $\times$ 2 labels = 400 variants per generator. Three candidates: Claude Sonnet 4.5, GPT-4o-mini, Gemini 3 Flash Preview. All via OpenRouter with \texttt{temperature=0.6}, \texttt{max\_tokens=600}. \textbf{Judge rubric.} Four dimensions (content preservation, factual fidelity, register authenticity, fluency), each decomposed into 2--3 binary questions \mbox{\citep{checkeval2025}}. Cross-family routing: Grok 4.1 Fast and GPT-5 Nano score both Sonnet and Gemini symmetrically, so differential judge harshness cancels. \textbf{Per-judge results.}}\new{We chose the generator on a 50-fact pilot: 50 facts, four registers, two labels, so 400 variants per candidate. The candidates were Claude Sonnet 4.5, GPT-4o-mini, and Gemini 3 Flash Preview, all called through OpenRouter at \texttt{temperature=0.6} with \texttt{max\_tokens=600} (Table~\ref{tab:generator-selection}). Two judges scored every variant on four dimensions (content preservation, factual fidelity, register authenticity, fluency), each broken into two or three binary questions \citep{checkeval2025}. Grok 4.1 Fast and GPT-5 Nano judged Sonnet and Gemini alike, so any difference in judge harshness cancels between those two (Table~\ref{tab:generator-selection}).}

\begin{table}[h]
\centering
\footnotesize
\setlength{\tabcolsep}{3pt}
\begin{tabular*}{\linewidth}{@{\extracolsep{\fill}}llrr@{}}
\toprule
Generator & Judge & Composite & FF on wrong \\
\midrule
Sonnet & Grok 4.1 & 0.989 & 0.983 \\
Sonnet & GPT-5 Nano & 0.950 & 0.772 \\
Gemini & Grok 4.1 & 0.973 & 0.980 \\
Gemini & GPT-5 Nano & 0.930 & 0.736 \\
\bottomrule
\end{tabular*}
\caption{Generator-selection pilot scores. Composite is the mean quality score; FF on wrong is factual fidelity on wrong-answer variants.}
\label{tab:generator-selection}
\end{table}

\old{\textbf{Pooled selection.} Weights: 40\% FF-on-wrong, 30\% composite, 20\% register authenticity, 10\% parse rate. Sonnet: 0.938. Gemini: 0.925. GPT-4o-mini: 0.872. \textbf{500-fact quality gate.} After scaling to 500 facts, pooled FF-on-wrong was 0.884; per-register: colloquial 0.913, clinical-note 0.872, patient 0.867. \textbf{Inter-rater agreement.} Exact-agreement rates exceeded 0.92. Cohen's $\kappa$ was fair-to-moderate (0.31--0.49), consistent with the high-baseline $\kappa$ paradox \mbox{\citep{landis1977kappa}}.}\new{Sonnet won on the pooled selection score, 0.938 against 0.925 for Gemini and 0.872 for GPT-4o-mini. That score weights factual fidelity on wrong answers at 40\%, the composite at 30\%, register authenticity at 20\%, and parse rate at 10\%. On the full 500-fact run the pooled fidelity on wrong answers was 0.884: 0.913 on colloquial, 0.872 on clinical note, and 0.867 on patient. The two judges gave the same binary answer more than 92\% of the time, while Cohen's $\kappa$ was only fair to moderate (0.31--0.49). That combination is the usual $\kappa$ paradox at a high agreement baseline \citep{feinstein1990kappa}.}

\subsection{Robustness check: Gemini generator}
\label{app:generator-ablation}
\begin{figure*}[!htbp]
\centering
\includegraphics[width=0.92\textwidth]{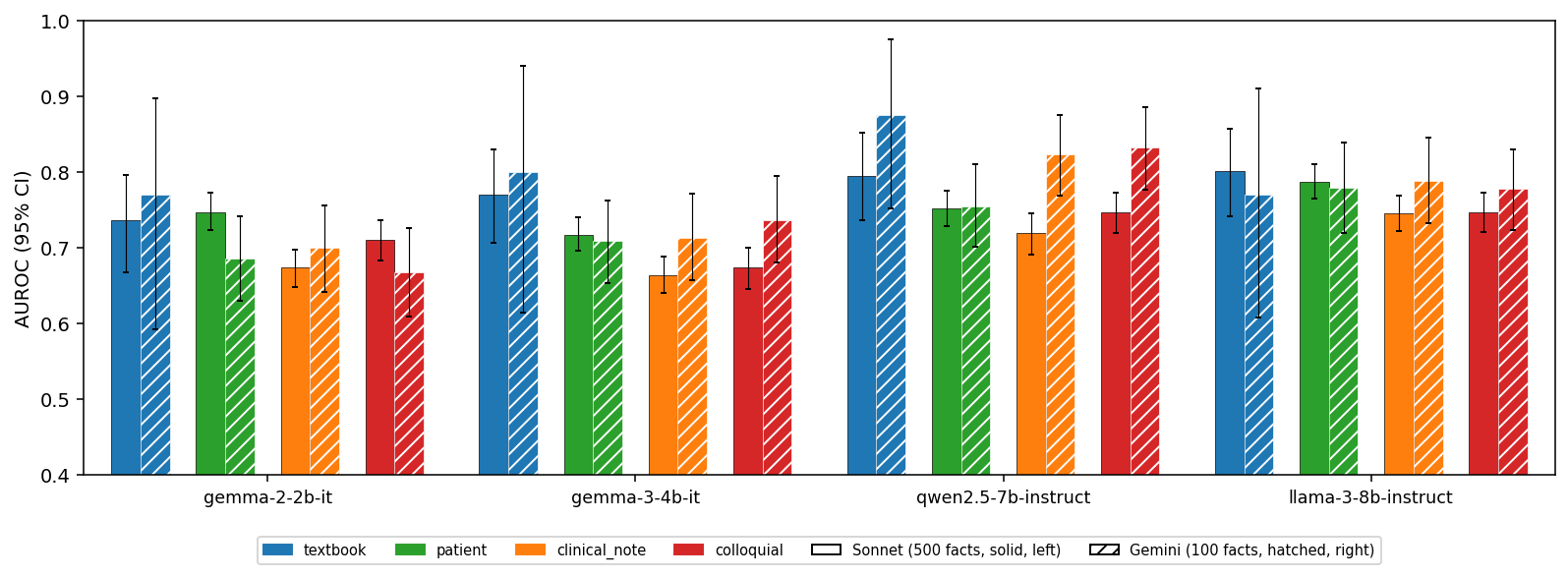}
\caption{Per-register AUROC on Sonnet-generated (solid) vs.\ Gemini-generated (hatched) variants. The two generators produce the same directional result.}
\label{fig:sonnet-vs-gemini}
\end{figure*}

100 facts $\times$ 4 registers $\times$ 2 labels = 800 variants generated by Gemini 3 Flash Preview. CIs are wider at $n{=}40$ held-out textbook facts\replace{;}{, but} the qualitative pattern (modest cross-register gap, rare-disease inversion, late-layer plateau) matches Sonnet\new{ (Table~\ref{tab:gemini}, Figure~\ref{fig:sonnet-vs-gemini})}. \new{Under the main-table protocol (same held-out facts, per-cell best layer, last question token) the mean register gap is $0.079$ AUROC for Gemini against $0.095$ for Sonnet. Per model, Gemma-2-2B $0.073$ vs.\ $0.064$, Gemma-3-4B $0.118$ vs.\ $0.085$, Llama-3-8B $0.012$ vs.\ $0.104$, and Qwen2.5-7B $0.113$ vs.\ $0.126$. The Llama value is a low outlier on a 100-fact subset. The mean is stable and shows that the two generators lie within $0.02$ of each other.}

\begin{table}[!htbp]
\centering
\footnotesize
\setlength{\tabcolsep}{3pt}
\begin{tabular*}{\linewidth}{@{\extracolsep{\fill}}lcccc@{}}
\toprule
\textbf{Model} & $\Rtext$ & $\Rpat$ & $\Rclin$ & $\Rcol$ \\
\midrule
Gemma-2B  & .770 & .686 & .700 & .667 \\
Gemma-3B  & .800 & .709 & .713 & .737 \\
Qwen-7B  & .875 & .755 & .823 & .832 \\
Llama-8B  & .770 & .780 & .788 & .778 \\
\bottomrule
\end{tabular*}
\caption{Best-layer AUROC on the Gemini 100-fact subset\new{. Non-textbook cells use all 800 variants; the held-out-fact gaps quoted in the text therefore differ}.}
\label{tab:gemini}
\end{table}

% NEW (camera-ready): HZcj W2, metareview item 1
\new{\subsection{Label-noise sensitivity: fidelity filtering}
\label{app:fidelity-filter}
Removing the wrong-answer rewrites that a judge considers unfaithful does not move the register gap. The quality gate of Appendix~\ref{app:generator} scored a 100-fact sample. Here we judged all 500 facts in every register with both cross-family judges and repeated the register analysis after dropping every fact whose wrong-answer variant scored below a factual-fidelity threshold. Grok 4.1 Fast was deprecated between the two judging rounds, so this round used Grok 4.3 alongside GPT-5 Nano. Both polarities of a dropped fact are removed, so every cell stays balanced. The probe, its layer, and the held-out split are the ones behind Table~\ref{tab:main}, Appendix~\ref{app:main-register-table}, and the unfiltered gap reproduces the main-table value of $0.095$ exactly.

\begin{table}[!htbp]
\centering
\footnotesize
\setlength{\tabcolsep}{3pt}
\begin{tabular*}{\linewidth}{@{\extracolsep{\fill}}lccc@{}}
\toprule
\textbf{Filter definition} & \textbf{FF $\ge 0.75$} & \textbf{FF $\ge 0.85$} & \textbf{Max change} \\
\midrule
Unfiltered              & .0952 & .0952 & --- \\
Pooled (both judges)    & .0929 & .0960 & .0023 \\
Grok 4.3 only           & .0948 & .0948 & .0004 \\
GPT-5 Nano only         & .0960 & .0960 & .0008 \\
\bottomrule
\end{tabular*}
\caption{Mean register gap $\Delta_\text{register}$ (four LLMs) after removing facts whose wrong-answer variant falls below a factual-fidelity (FF) threshold, for three filter definitions.}
\label{tab:fidelity-filter}
\end{table}

Instead of just one definition of a low-fidelity rewrite, we tested three: the pooled two-judge mean and each judge alone, at two thresholds each (Table~\ref{tab:fidelity-filter}). The largest change in the mean gap is $0.0023$ AUROC, and the per-model ordering is unchanged. From unfiltered to the pooled $0.85$ cut, Gemma-2-2B moves from $0.066$ to $0.058$ and Gemma-3-4B from $0.085$ to $0.080$. Llama-3-8B moves from $0.104$ to $0.109$, and Qwen2.5-7B from $0.126$ to $0.138$.

The two judges differ in strictness, which gives us two ends of the filtering range. At the $0.85$ cut GPT-5 Nano flags 36 of the 100 held-out patient variants, 48 of the clinical-note variants, and 36 of the colloquial variants, 40\% of all wrong variants. Grok 4.3 flags one. Because Grok scores almost every rewrite at $1.0$, the pooled filter is driven by Nano, and the pooled $0.75$ cut is the only setting in between (16 of 300 dropped). Removing 40\% of the wrong-answer variants moves the gap by $0.0008$, and removing almost none moves it by $0.0004$. Hence, the register result does not depend on the fidelity threshold or on which judge is trusted. The original 100-fact gate reported a fidelity on wrong answers of $0.884$ (Appendix~\ref{app:generator}), and Grok 4.3 on the full 500 facts gives $0.991$. The two judges bracket the true rewrite quality, and the result holds at both ends.}

\subsection{Perplexity of register variants and native corpora}
\label{app:perplexity}
\replace{Per-token perplexity of every register variant in our pipeline and of native MedQA and MedMCQA text under a fixed reference LM (Gemma-2-2B-it, bf16 on H100), formatted with the same \texttt{"Question:\,\{q\}\textbackslash nAnswer:\,\{a\}"} template the probe uses. Perplexity is computed as $\exp(\text{mean NLL per non-pad token})$.}{We score the per-token perplexity of every register variant, and of native MedQA and MedMCQA text, under one reference LM (Gemma-2-2B-it, bf16 on an H100), using the same \texttt{"Question:\,\{q\}\textbackslash nAnswer:\,\{a\}"} template the probe uses\replace{;}{, and} perplexity is $\exp(\text{mean NLL per non-pad token})$.} Native MedQA and MedMCQA text consists of the original SME-authored question and the correct or wrong answer (one row per polarity).

\replace{Table~\ref{tab:register-summary} pairs each register with a representative example variant of the same underlying fact, its mean reference-LM perplexity, and its mean register-$\Delta_\text{register}$ across the four probed LLMs on held-out facts. Stylistic distance from $\Rtext$ varies sharply across the four registers (perplexity 8.6 to 62.0), and the corresponding probe-AUROC drops are modest (0--0.144). Table~\ref{tab:perplexity} then gives the full distribution.}{Stylistic distance from $\Rtext$ varies sharply across the four registers (perplexity 8.6 to 62.0) while the probe-AUROC drops stay modest (0--0.144). Table~\ref{tab:register-summary} shows one fact in all four registers with each variant's mean reference-LM perplexity and the mean $\Delta_\text{register}$ across the four probed LLMs on held-out facts\replace{;}{, and} Table~\ref{tab:perplexity} gives the full distribution.}

\begin{table}[!htbp]
\centering
\footnotesize
\setlength{\tabcolsep}{3pt}
\begin{tabularx}{\columnwidth}{@{}l>{\raggedright\arraybackslash}Xrr@{}}
\toprule
\textbf{Register} & \textbf{Example variant} & \textbf{Perplexity} & $\boldsymbol{\Delta_\text{register}}$ \\
\midrule
$\Rtext$ & \emph{What is the most frequently encountered etiology of right-sided heart failure?} & 8.6 & 0.000 \\
\addlinespace[3pt]
$\Rpat$  & \emph{My dad was told he has right-sided heart failure, what usually causes it?} & 13.1 & 0.047 \\
\addlinespace[3pt]
$\Rclin$ & \emph{Most common etiology R-sided HF?} & 62.0 & 0.144 \\
\addlinespace[3pt]
$\Rcol$  & \emph{ok so what actually causes right side heart failure lol} & 51.6 & 0.079 \\
\bottomrule
\end{tabularx}
\caption{Register summary: a representative example variant of the same MedQA fact in each register, the mean reference-LM perplexity (Gemma-2-2B-it), and the mean register-$\Delta_\text{register}$ across the four probed LLMs (held-out facts). Examples are drawn from a single illustrative right-sided-heart-failure item.}
\label{tab:register-summary}
\end{table}

\begin{table}[!b]
\centering
\footnotesize
\setlength{\tabcolsep}{3pt}
\resizebox{\columnwidth}{!}{%
\begin{tabular}{@{}llrrrrrr@{}}
\toprule
\textbf{Source} & \textbf{Register} & $\boldsymbol{n}$ & \textbf{Mean} & \textbf{Median} & \textbf{Q1} & \textbf{Q3} & \textbf{P95} \\
\midrule
MedQA     & ---            & 1000 & 8.7   & 8.2   & 6.5   & 10.1  & 14.7 \\
MedQA-S     & $\Rtext$       & 1000 & 8.6   & 7.9   & 6.3   & 10.0  & 14.8 \\
MedQA-S     & $\Rpat$        & 1000 & 13.1  & 12.5  & 10.6  & 14.7  & 19.6 \\
MedQA-S     & $\Rclin$       & 1000 & 62.0  & 49.8  & 31.5  & 77.3  & 146.2 \\
MedQA-S     & $\Rcol$        & 1000 & 51.6  & 46.3  & 35.5  & 61.8  & 94.2 \\
\addlinespace
MedMCQA   & ---            & 1000 & 124.5 & 60.3  & 33.0  & 133.0 & 385.8 \\
MedMCQA-S   & $\Rtext$       &  200 & 15.4  & 12.3  & 9.3   & 18.6  & 33.8 \\
MedMCQA-S   & $\Rpat$        &  200 & 18.6  & 16.7  & 12.7  & 21.8  & 34.6 \\
MedMCQA-S   & $\Rclin$       &  200 & 260.2 & 140.1 & 74.5  & 279.8 & 886.1 \\
MedMCQA-S   & $\Rcol$        &  200 & 173.9 & 104.2 & 71.0  & 178.0 & 487.6 \\
\bottomrule
\end{tabular}%
}
\caption{Reference-LM perplexity by (source, register). Lower is closer to the reference LM's training distribution. $n$ is the number of variants scored.}
\label{tab:perplexity}
\end{table}

\paragraph{In-distribution fractions.} Native MedQA's P5--P95 range is $[4.82, 14.67]$. The fraction of MedQA-S variants whose perplexity lies inside this range is 88.5\% for $\Rtext$, 73.9\% for $\Rpat$, 2.4\% for $\Rclin$, and 0.5\% for $\Rcol$. The fraction of MedMCQA text inside the same range is 2.4\%, so MedMCQA overlaps the native-MedQA perplexity distribution about as much as the highest-perplexity register ($\Rclin$) does.

\paragraph{Perplexity and probe drop.} \old{The register ordering by mean perplexity ($\Rtext$ 8.6, $\Rpat$ 13.1, $\Rcol$ 51.6, $\Rclin$ 62.0) is the same as the register ordering by mean probe register-$\Delta$ on held-out facts ($\Rpat$ 0.047, $\Rcol$ 0.079, $\Rclin$ 0.144). The relationship is monotone across registers, and we do not claim a parametric fit at $n{=}4$ registers. \textbf{Caveat.} Absolute perplexity values are with respect to one reference LM. The relative ordering across registers should hold under any reference LM in the same parameter range, and this ordering is what our argument turns on.}\new{Registers rank identically by mean perplexity ($\Rtext$ 8.6, $\Rpat$ 13.1, $\Rcol$ 51.6, $\Rclin$ 62.0) and by mean probe drop on held-out facts ($\Rpat$ 0.047, $\Rcol$ 0.079, $\Rclin$ 0.144). With four registers this is an ordinal agreement, not a fitted relationship. Perplexity is measured under one reference LM (Gemma-2-2B-it); we use only the ordering, not the absolute values.}

\subsection{Rewriter and judge prompts}
\label{app:prompts}
\old{For reproducibility we release the exact prompts used at generation and judging time. Figures~\ref{fig:appendix-gen-prompts-1} and~\ref{fig:appendix-gen-prompts-2} (at the end of this appendix) show the register-rewriter prompt structure: a style vignette anchoring tone and vocabulary, a correct-answer few-shot exemplar, a wrong-preserving few-shot exemplar (the critical inductive bias that prevents the LLM from ``helpfully correcting'' wrong answers), and an absolute-rule instruction block per register. Figure~\ref{fig:appendix-judge-prompt} (also at the end) shows the judge-side prompt: system prompt, rubric (four dimensions $\times$ binary yes/no questions), and the required JSON output schema.}\new{The rewriter prompt has four parts per register, a style vignette that sets tone and vocabulary, a correct-answer exemplar, a wrong-preserving exemplar, and a block of absolute rules. Without the wrong-preserving exemplar, the generator silently repairs the medically incorrect answer while rewriting it, which destroys the label. Figures~\ref{fig:appendix-gen-prompts-1} and~\ref{fig:appendix-gen-prompts-2}, Appendix~\ref{app:prompt-figures} (end of the appendix) show the rewriter prompts, and Figure~\ref{fig:appendix-judge-prompt}, Appendix~\ref{app:prompt-figures} shows the judge prompt with its rubric (four dimensions, binary yes/no questions) and JSON output schema. The exact prompt files are in the released repository.}

\paragraph{Cross-family judge routing.}
The rubric in Figure~\ref{fig:appendix-judge-prompt}, Appendix~\ref{app:prompt-figures} is executed by two independent judges per generator, always drawn from model families different from the generator's. Our mapping: Claude Sonnet 4.5 (Anthropic) is judged by Grok 4.1 Fast (xAI) $+$ GPT-5 Nano (OpenAI)\replace{;}{,} GPT-4o-mini (OpenAI) is judged by Claude Sonnet 4.5 (Anthropic) $+$ Gemini 3 Flash (Google)\replace{;}{, and} Gemini 3 Flash (Google) is judged by Grok 4.1 Fast $+$ GPT-5 Nano. Pooled binary-question exact agreement is $\geq 0.92$ across all generator-by-judge-pair combinations, and Cohen's $\kappa$ is fair-to-moderate, the expected $\kappa$ paradox at this agreement baseline. This cross-family routing cancels any differential judge harshness between candidate generators.

\FloatBarrier

\section{Extended probe results}
\label{app:probe-results}

\subsection{Per-register AUROC, full table}
\label{app:main-register-table}
\replace{Table~\ref{tab:main} reports the full per-(model, register) numerical values plotted in Figure~\ref{fig:register-bars}, with 95\% bootstrap CIs and per-model mean $\Delta_\text{register}$.}{$\Rclin$ is the hardest register for every model. Table~\ref{tab:main} gives the per-(model, register) values plotted in Figure~\ref{fig:register-bars}, with 95\% bootstrap CIs and the per-model mean $\Delta_\text{register}$.}

\begin{table*}[!tbp]
\centering
\small
\setlength{\tabcolsep}{6pt}
\begin{tabular*}{\textwidth}{@{\extracolsep{\fill}}lccccc@{}}
\toprule
\textbf{Model} & $\boldsymbol{\Rtext}$ & $\boldsymbol{\Rpat}$ & $\boldsymbol{\Rclin}$ & $\boldsymbol{\Rcol}$ & \textbf{Mean $\boldsymbol{\Delta_\text{register}}$} \\
\midrule
Gemma-2B       & .733 [.67,.79] & .740 [.68,.79] & .608 [.55,.66] & .661 [.60,.72] & $-$.064 \\
Gemma-3B       & .768 [.71,.83] & .735 [.67,.79] & .614 [.55,.67] & .701 [.64,.76] & $-$.085 \\
Qwen-7B & .795 [.74,.85] & .707 [.64,.77] & .623 [.56,.69] & .723 [.66,.78] & $-$\replace{.124}{.126} \\
Llama-8B & .802 [.74,.86] & .731 [.67,.79] & .681 [.62,.74] & .699 [.64,.76] & $-$.104 \\
\midrule
Mean                & \textbf{.775}  & \textbf{.728}  & \textbf{.631}  & \textbf{.696}  & \textbf{$-$.095} \\
\bottomrule
\end{tabular*}
\caption{Best-layer AUROC per register, last-question-token, Sonnet variants, 95\% bootstrap CIs (1,000 iterations). All registers use the same held-out 20\% fact split ($n{=}200$), removing fact-level training overlap on the non-textbook registers. $\Delta_\text{register}$ is the mean AUROC drop from textbook.}
\label{tab:main}
\end{table*}

\subsection{Layer-selection sensitivity}
\label{app:layer-rules}

The main results select, per (model, register), the textbook-trained layer with the highest held-out AUROC (per-cell). Because a deployed probe may not be able to choose a different layer per register, we report the register gap under three layer-selection rules, all evaluated on the same held-out 20\% facts\new{ (Table~\ref{tab:layer-rules})}:

\begin{itemize}[leftmargin=*, topsep=0pt, partopsep=0pt, itemsep=2pt]
    \item \textbf{Per-cell} (main): for each (LLM, register) the layer that maximises that cell's AUROC.
    \item \textbf{Best-mean single layer}: for each LLM, the single layer that maximises the mean AUROC across all four registers. \replace{This is what a practitioner who must commit to one layer at deploy time would choose with access to all four registers at validation time.}{This rule corresponds to a deployment with one fixed layer, chosen with all four registers available at validation time.}
    \item \textbf{Textbook-blind single layer}: for each LLM, the single layer that maximises AUROC on the textbook validation split alone. \replace{This is the strict zero-shot rule (the probe is fit and selected without any non-textbook data) and is the worst operating point for transfer by construction.}{The probe is fit and selected without any non-textbook data, so this rule is the strict zero-shot setting and, by construction, the worst operating point for transfer.}
\end{itemize}

\begin{table}[!htbp]
\centering
\footnotesize
\setlength{\tabcolsep}{3pt}
\resizebox{\columnwidth}{!}{%
\begin{tabular}{@{}lccccc@{}}
\toprule
\textbf{Rule} & \textbf{Gemma-2B} & \textbf{Gemma-3B} & \textbf{Qwen} & \textbf{Llama} & \textbf{Mean} \\
\midrule
Per-cell (main)         & $-$.064 & $-$.085 & $-$\replace{.124}{.126} & $-$.104 & $-$\textbf{.095} \\
Best-mean single layer  & $-$.049 & $-$.061 & $-$.117 & $-$.094 & $-$\textbf{.080} \\
Textbook-blind layer    & $-$.177 & $-$.162 & $-$.154 & $-$.148 & $-$\textbf{.160} \\
\bottomrule
\end{tabular}%
}
\caption{Mean register gap $\Delta_\text{register}$ under three layer-selection rules. All evaluated on the same held-out 20\% facts as the textbook split. Per-cell and best-mean give similar gaps; textbook-blind is the strict zero-shot upper bound.}
\label{tab:layer-rules}
\end{table}

Per-cell and best-mean give \replace{very similar}{similar} gaps (mean 0.095 vs.\ 0.080)\replace{;}{, and} both sit far below the cross-corpus drop (\replace{0.215}{0.21}). The textbook-blind single layer, which never sees any non-textbook data at selection time, gives a strictly larger gap of 0.160 but is still below the cross-corpus value. The qualitative conclusion (register $\ll$ corpus) therefore holds under all three rules. We report per-cell as the primary number for continuity with the probing literature \citep{marks2024geometry,burger2024truth}, and note that on average the fact-level overlap from the earlier evaluation protocol (non-textbook evaluated on all $n{=}1000$ variants, including training facts) inflated $\Delta_\text{register}$ downward by 0.042 AUROC. \new{For completeness, under that all-variants protocol the textbook probe scores a mean of $0.755$ on $\Rpat$, $0.704$ on $\Rclin$, and $0.723$ on $\Rcol$ (the textbook row of Table~\ref{tab:register-matrix}, Appendix~\ref{app:register-matrix}), a mean gap of $0.053$ against $0.095$ on matched held-out facts. The held-out version is the one reported throughout.}

\subsection{Specialty-disjoint transfer: per-split detail}
\label{app:topic-transfer}
The specialty-disjoint transfer experiment partitions the 351 specialty-tagged MedQA facts (S-MedQA join, Section~\ref{sec:transfer_specialty}) into two halves of seven specialties each, drawn at random per split. Probes are trained on the textbook-register variants from the train half (using a fact-level 80/20 split inside that half so the probe sees neither polarity of any test-half fact and reserves a within-train held-out set), and evaluated on the textbook-register variants from the test half. The held-out 20\% within-train split also serves as the in-distribution baseline. Register is held fixed at $\Rtext$ in both train and test, isolating specialty from surface form. We run five random splits per model with 1{,}000-iteration fact-level bootstrap CIs at every cell. \replace{The per-model $\Delta_\text{specialty}$ values in Table~\ref{tab:headline} are reproduced with in-distribution baselines and bootstrap CIs in Table~\ref{tab:topic_transfer}; the per-split breakdown then follows.}{Table~\ref{tab:topic_transfer} adds the in-distribution baselines and CIs behind the per-model $\Delta_\text{specialty}$ of Table~\ref{tab:headline}.}

\begin{table}[!htbp]
\centering
\footnotesize
\setlength{\tabcolsep}{3pt}
\begin{tabular*}{\linewidth}{@{\extracolsep{\fill}}lccc@{}}
\toprule
\textbf{Model} & \textbf{In-distribution} & \textbf{Held-out} & $\boldsymbol{\Delta_\text{specialty}}$ \\
\midrule
Gemma-2B       & .757 [.64,.86] & .705 [.66,.75] & $-$.053 \\
Gemma-3B       & .746 [.64,.84] & .735 [.69,.78] & $-$.010 \\
Qwen-7B & .799 [.69,.89] & .783 [.74,.83] & $-$.016 \\
Llama-8B & .834 [.73,.92] & .789 [.75,.83] & $-$.045 \\
\midrule
\textbf{Mean} & \textbf{.784} & \textbf{.753} & \textbf{$-$.031} \\
\bottomrule
\end{tabular*}
\caption{Specialty-disjoint transfer per model. \emph{In-distribution}: within-train held-out 20\% baseline. \emph{Held-out}: held-out-specialty AUROC. $\Delta_\text{specialty}$: drop. Bootstrap CIs (1{,}000 iterations, fact-level) are averaged across five splits.}
\label{tab:topic_transfer}
\end{table}

\paragraph{Per-split breakdown.} Table~\ref{tab:topic_transfer_splits} reports $\Delta_\text{specialty}$ for each (model, split) pair at the best layer chosen by mean in-distribution AUROC. \replace{Split-level standard deviation across the five splits is small (0.013--0.057), consistent with the result not depending on which specialties happen to land on which side of the split.}{The standard deviation across the five splits is small (0.013--0.057), so the result does not depend on which specialties land on which side of the split.}

\begin{table}[!htbp]
\centering
\footnotesize
\setlength{\tabcolsep}{2pt}
\begin{tabular*}{\linewidth}{@{\extracolsep{\fill}}lccccccc@{}}
\toprule
\textbf{Model} & \textbf{S0} & \textbf{S1} & \textbf{S2} & \textbf{S3} & \textbf{S4} & $\bar{\Delta}$ & $\sigma$ \\
\midrule
Gemma-2B       & $-$.052 & $+$.005 & $-$.086 & $-$.080 & $-$.049 & \textbf{$-$.053} & .036 \\
Gemma-3B       & $-$.031 & $-$.005 & $-$.094 & $+$.058 & $+$.020 & \textbf{$-$.010} & .057 \\
Qwen-7B & $-$.009 & $-$.021 & $+$.001 & $-$.017 & $-$.034 & \textbf{$-$.016} & .013 \\
Llama-8B & $-$.120 & $-$.015 & $-$.010 & $-$.034 & $-$.047 & \textbf{$-$.045} & .045 \\
\midrule
\textbf{Mean} & $-$.053 & $-$.009 & $-$.047 & $-$.018 & $-$.028 & \textbf{$-$.031} & --- \\
\bottomrule
\end{tabular*}
\caption{Per-(model, split) $\Delta_\text{specialty}$ at the best-textbook layer (last-question-token). S0--S4 are the five random partitions of the 15 specialties into a 7-specialty training half and an 8-specialty held-out half. $\overline{\Delta}$ is the per-model mean across the five splits; $\sigma$ is the across-split standard deviation.}
\label{tab:topic_transfer_splits}
\end{table}

\paragraph{Position robustness.} The first-answer-token result is qualitatively the same. Mean $\Delta_\text{specialty}$ at first-answer-token is 0.035 (vs.\ 0.031 at last-question-token), with the same per-model ordering (Llama-8B and Gemma-2-2B at the high end around 0.04--0.05, Qwen-7B at the low end around 0.01). Both positions are reported in the released CSV.

% NEW (camera-ready): HZcj W4
\new{\subsection{Specialty selection: matched vs.\ unmatched facts}
\label{app:specialty-selection}
The specialty experiment uses only the 351 facts that the exact-match join to S-MedQA labels, and the other 149 are excluded. If the matched facts were an easier slice, the specialty result would be optimistic (but they are not). Table~\ref{tab:specialty-selection} (\S\ref{sec:transfer_specialty}) compares the main probe's AUROC, pooled over the three shifted registers, on matched and unmatched facts with 95\% fact-level bootstrap CIs. The intervals overlap for every LLM and the direction is inconsistent, with Qwen2.5-7B marginally higher on the unmatched facts. Restricting the comparison to the held-out split alone (81 matched facts, 19 unmatched) gives the same conclusion, with matched AUROC between $0.651$ and $0.687$ and unmatched between $0.640$ and $0.724$ on much wider intervals.

}

\begin{figure}[!tbp]
\centering
\includegraphics[width=0.9\columnwidth]{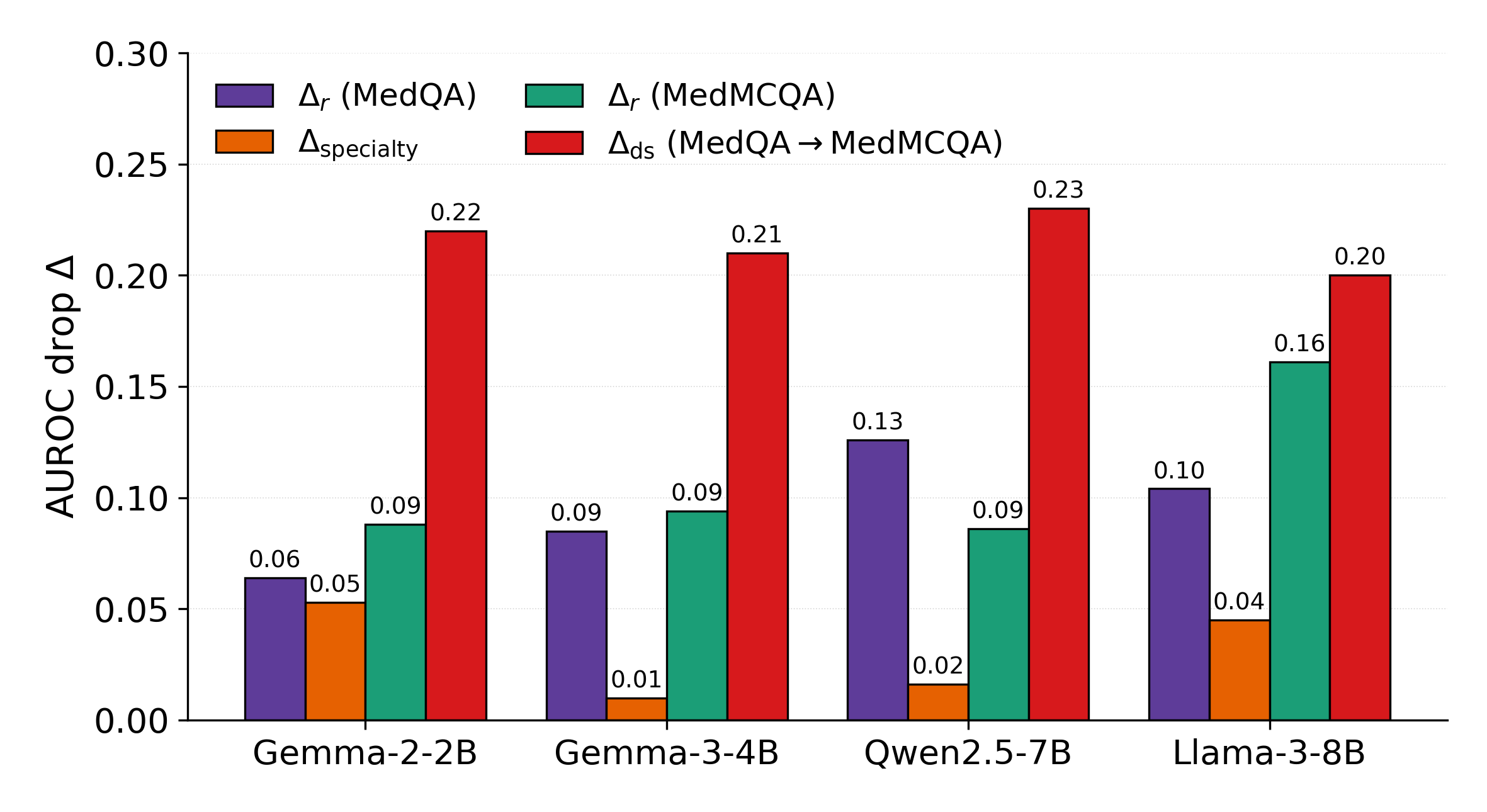}
\caption{Per-LLM comparison of the four within- and cross-corpus shifts. \textbf{Bars left to right per LLM}: within-MedQA register $\Delta_\text{register}$, specialty-disjoint transfer $\Delta_\text{specialty}$, within-MedMCQA register $\Delta_\text{register}$, MedQA-to-MedMCQA cross-corpus $\Delta_\text{dataset}$. The cross-corpus drop dominates the others by a factor of roughly 2--4 in every LLM.}
\label{fig:axes_decomposition}
\vspace{0.4\baselineskip}
\end{figure}

\paragraph{Quantitative comparison.} The three within-corpus shifts are all small compared to the cross-corpus drop. The within-MedQA register estimate (0.095) and the specialty-transfer drop (0.031) sum to 0.126 AUROC\replace{;}{, and} the within-MedMCQA register estimate (0.107), itself inflated by the distractor-expansion artefact described in Section~\ref{sec:transfer_corpus}, together with the same specialty drop sums to 0.138 AUROC. The strict worst-case additive bound, $0.107 + 0.031 + 0.011 = 0.149$ (adding the gap between the two within-corpus register estimates), still falls below the 0.21 cross-corpus gap. We do not report these as a percentage decomposition\replace{:}{, because} register, specialty, and item construction are not independent factors in representation space, and the within-MedMCQA register estimate is \replace{an upper bound rather than a like-for-like measurement of register shift in that corpus}{an upper bound on register shift in that corpus, since it is not a like-for-like measurement}. The comparison establishes a magnitude relation, not an attributable share. Figure~\ref{fig:axes_decomposition}, Appendix~\ref{app:topic-transfer} shows the four shifts per model.

\begin{table}[t]
\centering
\footnotesize
\setlength{\tabcolsep}{3pt}

\begin{tabular*}{\linewidth}{@{\extracolsep{\fill}}lcccc@{}}
\toprule
\textbf{Model} & \textbf{layer} & \textbf{MedQA} & \textbf{MedMCQA} & $\boldsymbol{\Delta_\text{dataset}}$ \\
\midrule
Gemma-2B       & 15 & .737 [.67,.80] & .520 [.50,.54] & $-$0.22 \\
Gemma-3B       & 25 & .770 [.71,.83] & .555 [.53,.58] & $-$0.21 \\
Qwen-7B        & 23 & .795 [.74,.85] & .564 [.54,.59] & $-$0.23 \\
Llama-8B       & 29 & .801 [.74,.86] & .603 [.58,.63] & $-$0.20 \\
\midrule
Mean           & --- & \textbf{.776} & \textbf{.561} & \textbf{$-$0.21} \\
\bottomrule
\end{tabular*}

\caption{Cross-dataset transfer. Probe trained on MedQA-textbook 80\% split and evaluated on (i) MedQA-textbook held-out 20\%, (ii) all 500 MedMCQA validation items. Best-layer last-question-token, 95\% bootstrap CIs.}
\label{tab:cross_dataset}
\end{table}

\subsection{Cross-corpus transfer: per-model detail}
\label{app:cross-dataset}
\replace{Table~\ref{tab:cross_dataset} reports the per-model layer chosen as best on MedQA-textbook (the same layer is used for the MedMCQA evaluation, with no retraining), the MedQA-textbook held-out AUROC, the MedMCQA AUROC at that layer, and $\Delta_{\text{dataset}}$. Best layers concentrate in the upper half of each network and match those used elsewhere in the paper.}{The best layers lie in the upper half of each network and are the ones used throughout the paper. Table~\ref{tab:cross_dataset} lists, per model, that layer, the MedQA-textbook held-out AUROC, the MedMCQA AUROC at the same layer with no retraining, and $\Delta_\text{dataset}$.}

% NEW (camera-ready): 9nyL W1, HZcj W3, metareview item 6
\new{\subsection{Corpus ladder: MMLU-medical and the format control}
\label{app:corpus-ladder}
The corpus ladder of \S\ref{sec:transfer_corpus} applies the MedQA-trained probe at its MedQA-selected layer, the paper's $\Delta_\text{dataset}$ protocol, so that all rows are comparable with Table~\ref{tab:cross_dataset}, Appendix~\ref{app:cross-dataset}. Table~\ref{tab:corpus-ladder-full} gives the per-model values with 95\% CIs for that protocol and, for comparison, the values obtained when the layer is instead chosen per evaluation set (the per-cell rule used for registers in Table~\ref{tab:main}, Appendix~\ref{app:main-register-table}). Layer choice matters more here than within MedQA. With a per-cell layer the reformatted set rises to $0.589$ and MMLU-medical to $0.669$, while native MedMCQA falls to $0.503$ because its best layer under that rule is a shallow, near-chance one. Comparing sets at different layers would overstate the format effect, which is why the main text uses the fixed layer. The new sets were scored from a re-extraction of activations in which the MedQA in-corpus value is $0.772$ at the fixed layer, within $0.005$ of Table~\ref{tab:cross_dataset}, Appendix~\ref{app:cross-dataset}.

\begin{table}[!htbp]
\centering
%\footnotesize
%\setlength{\tabcolsep}{2pt}
\resizebox{\linewidth}{!}{%
\begin{tabular}{@{}p{3cm}p{0.8cm}cccc@{}}
\toprule
\textbf{Set} & \textbf{Layer} & \textbf{Gemma-2B} & \textbf{Gemma-3B} & \textbf{Qwen} & \textbf{Llama} \\
\midrule
MMLU-medical & F & .585 [.57,.60] & .614 [.59,.64] & .690 [.67,.71] & .721 [.70,.75] \\
             & PC & .596 [.57,.62] & .630 [.61,.65] & .709 [.69,.73] & .739 [.71,.76] \\
\addlinespace
\multirow{2}{3cm}{MedMCQA, reformatted} & F & .517 [.47,.56] & .532 [.47,.59] & .608 [.54,.67] & .581 [.51,.65] \\
             & PC & .527 [.49,.57] & .590 [.53,.65] & .626 [.56,.69] & .613 [.57,.66] \\
\addlinespace
\multirow{2}{3cm}{MedMCQA, Sonnet-rewritten} & F & .671 [.61,.73] & .703 [.64,.77] & .755 [.69,.82] & .780 [.72,.84] \\
             & PC & .685 [.63,.75] & .717 [.66,.77] & .777 [.72,.83] & .780 [.72,.83] \\
\bottomrule
\end{tabular}%
}
\caption{Per-model AUROC with 95\% bootstrap CIs for the evaluation-only sets, under the fixed MedQA-selected layer (main text) and a per-cell best layer. MedMCQA Sonnet-rewritten is the 100-fact within-MedMCQA replication set (textbook register, $n{=}200$), scored here by the MedQA probe. \textbf{Layer F:} fixed. \textbf{Layer PC}: per-cell.}
\label{tab:corpus-ladder-full}
\end{table}

\paragraph{Reformatting procedure.} 100 MedMCQA validation items (seed 42) were rewritten by Claude Sonnet 4.5 from terse board-exam stems into USMLE-style clinical vignettes, and the correct answer and the sampled distractor were copied verbatim, so only the question changes. Mean question length rises from 13.9 to 58.2 words. As an illustration, the stem \emph{``Concentric teeth bite mark on forearm, what to do next?''} becomes \emph{``A 28-year-old woman presents to the emergency department with a bite mark on her forearm sustained during an altercation 2 hours ago. Physical examination reveals a concentric pattern of teeth marks with mild erythema and superficial breaks in the skin. What is the most appropriate next step in management?''}, with the answer pair (\emph{complete description of injury as seen} vs.\ \emph{keep scale for measuring below the mark and take photo in the plane of bite}) unchanged. At the fixed layer the reformatted set scores $0.559$ against $0.561$ for native MedMCQA, so MedQA-style phrasing on its own does not recover the drop. The set is small ($n{=}100$ facts) and its CIs are wide, so we describe the format effect as at most minor.

\paragraph{Rewritten MedMCQA.} When the same 100 items are instead fully rewritten by the generator that produced the training data (answers included), the MedQA probe transfers at $0.727$, close to its in-corpus value. This set shares both register and rewriting conventions with the training distribution, so it gives an upper bound on cross-corpus transfer without isolating one factor. The within-MedMCQA replication of Appendix~\ref{app:medmcqa-register} uses the same set with a probe trained on it. Appendix~\ref{app:length-cue} discusses one surface property of the rewrites that this comparison is sensitive to.

\paragraph{MMLU by subject.} Transfer to MMLU-medical is not driven by the long, vignette-style subjects. At the per-cell layer the mean AUROC over LLMs is $0.724$ on clinical knowledge (147 items, 11 words per question on average), $0.717$ on medical genetics (53 items), $0.676$ on college medicine (100 items, 58 words), $0.675$ on anatomy (63 items), and $0.647$ on professional medicine (137 items, 111 words). The short-stem subjects, which look least like MedQA vignettes, transfer best.

\paragraph{Symmetry.} A probe trained on native MedMCQA-textbook (80\% of the 500 validation facts, per-cell best layer) reaches $0.607$ on the MedQA held-out set (per model $0.574$, $0.575$, $0.639$, $0.640$) and $0.670$ on MMLU-medical ($0.584$, $0.638$, $0.713$, $0.745$). MedMCQA-trained probes therefore transfer to MMLU about as well as MedQA-trained ones do, and poorly to MedQA, mirroring the MedQA-to-MedMCQA failure.}

\subsection{Within-MedMCQA register replication}
\label{app:medmcqa-register}
The within-MedMCQA register replication uses 100 MedMCQA validation-split facts, the same Sonnet rewriter and four registers as the primary MedQA dataset (4 registers $\times$ 2 labels = 800 variants), and the same probe-training protocol as Section~\ref{sec:methodology:probing}\new{. Per-(model, register) values are in Table~\ref{tab:medmcqa-textbook}}. Probes are trained on the MedMCQA-textbook 80\% fact-level split and evaluated on the same held-out 20\% facts ($n{=}40$) across all four registers, matching the MedQA protocol. \replace{Best layer per model is chosen by within-MedMCQA-textbook AUROC.}{The best layer is chosen per (model, register) cell.} \replace{Aggregate numbers are reported in Section~\ref{sec:transfer_corpus}; the per-(model, register) table follows.}{Section~\ref{sec:transfer_corpus} gives the aggregate and Table~\ref{tab:medmcqa-textbook} the per-(model, register) values.}

\begin{table*}[!tbp]
\centering
\small
\setlength{\tabcolsep}{6pt}
\begin{tabular*}{\textwidth}{@{\extracolsep{\fill}}lccccc@{}}
\toprule
\textbf{Model} & $\boldsymbol{\Rtext}$ & $\boldsymbol{\Rpat}$ & $\boldsymbol{\Rclin}$ & $\boldsymbol{\Rcol}$ & $\boldsymbol{\Delta_\text{register}}$ \\
\midrule
Gemma-2B       & .748 [.63,.86] & .715 \replace{[.66,.77]}{[.60,.85]} & .633 \replace{[.57,.69]}{[.49,.77]} & .630 \replace{[.57,.69]}{[.53,.74]} & $-$0.088 \\
Gemma-3B       & .720 [.54,.88] & .758 \replace{[.69,.82]}{[.64,.88]} & .510 \replace{[.45,.57]}{[.41,.62]} & .610 \replace{[.54,.68]}{[.51,.73]} & $-$0.094 \\
Qwen-7B & .800 [.62,.94] & .805 \replace{[.74,.86]}{[.68,.92]} & .637 \replace{[.57,.70]}{[.51,.77]} & .700 \replace{[.63,.77]}{[.55,.84]} & $-$0.086 \\
Llama-8B & .797 [.61,.92] & .740 \replace{[.68,.80]}{[.62,.86]} & .552 \replace{[.49,.61]}{[.48,.64]} & .616 \replace{[.55,.68]}{[.46,.77]} & $-$0.161 \\
\midrule
\textbf{Mean}       & \textbf{.766}  & \textbf{.754}  & \textbf{.583}  & \textbf{.639}  & \textbf{$-$0.107} \\
\bottomrule
\end{tabular*}
\caption{Within-MedMCQA per-register AUROC (\replace{best layer per model}{per-cell best layer}, 95\% bootstrap CIs, 100-fact subset with held-out 20\% fact split $n{=}40$; CIs are accordingly wide). $\Delta_\text{register}^{\text{MedMCQA}}$ is the mean drop from MedMCQA-textbook to the other three registers on matched held-out facts.}
\label{tab:medmcqa-textbook}
\end{table*}

\replace{The MedMCQA-textbook column (mean 0.766) is within 0.010 of the MedQA-textbook in-distribution baseline (0.775, Table~\ref{tab:headline}), confirming that the probe technique works on MedMCQA when trained on MedMCQA. The mean MedMCQA register-$\Delta$ (0.107) is close in magnitude to the MedQA register-$\Delta$ (0.095) on matched held-out facts. The slightly larger value within MedMCQA is consistent with the wider CIs at $n{=}40$ and with the distractor-expansion artefact described in Section~\ref{sec:transfer_corpus}.}{The MedMCQA-textbook column (mean 0.766) is within 0.010 of the MedQA-textbook baseline (0.775, Table~\ref{tab:headline}), so the probe works on MedMCQA when it is trained on MedMCQA. The mean MedMCQA register-$\Delta$ (0.107) is close to the MedQA value (0.095) on matched held-out facts, and the slightly larger value is what the wider CIs at $n{=}40$ and the distractor-expansion artefact described in Section~\ref{sec:transfer_corpus} would produce.}

\subsection{Common vs.\ rare AUROC, per-(model, register)}
\label{app:rarity-full}
\replace{Table~\ref{tab:rarity-full} provides the full 12-row common vs rare breakdown summarised in}{Table~\ref{tab:rarity-full} gives the 12-row common vs.\ rare breakdown summarised in} Section~\ref{sec:transfer_specialty} \new{and plotted in Figure~\ref{fig:specialty}b}. All non-textbook conditions of the four Sonnet-generated registers are included\replace{;}{, and} the textbook register has no register-induced rarity contrast and is omitted.

\begin{table}[!htbp]
\centering
\footnotesize
\setlength{\tabcolsep}{2pt}
\resizebox{\linewidth}{!}{%
\begin{tabular*}{\linewidth}{@{\extracolsep{\fill}}llcc@{}}
\toprule
\textbf{Model} & \textbf{Register} & \textbf{Common} & \textbf{Rare} \\
\midrule
\new{Gemma-2B & $\Rtext$ & .726 [.66,.78] & .812 [.44,1.0] \\}
Gemma-2B & $\Rpat$ & \replace{.747 [.72,.77]}{.751 [.72,.78]} & \replace{.760 [.68,.85]}{.775 [.69,.87]} \\
Gemma-2B & $\Rclin$ & \replace{.670 [.64,.70]}{.665 [.64,.69]} & \replace{.706 [.60,.80]}{.696 [.60,.79]} \\
Gemma-2B & $\Rcol$$^{*}$ & \replace{.696 [.67,.72]}{.703 [.67,.73]} & \replace{.837 [.75,.91]}{.843 [.76,.92]} \\
\new{Gemma-3B & $\Rtext$ & .762 [.69,.82] & .938 [.81,1.0] \\}
Gemma-3B & $\Rpat$ & \replace{.712 [.69,.74]}{.732 [.71,.76]} & \replace{.751 [.67,.83]}{.765 [.67,.86]} \\
Gemma-3B & $\Rclin$ & \replace{.659 [.63,.69]}{.666 [.64,.69]} & \replace{.677 [.59,.77]}{.702 [.61,.80]} \\
Gemma-3B & $\Rcol$ & \replace{.671 [.64,.70]}{.666 [.64,.69]} & \replace{.756 [.67,.84]}{.769 [.69,.86]} \\
\new{Qwen-7B & $\Rtext$ & .815 [.76,.87] & .875 [.75,1.0] \\}
Qwen-7B & $\Rpat$$^{*}$ & \replace{.740 [.72,.77]}{.748 [.72,.78]} & \replace{.868 [.80,.94]}{.880 [.81,.94]} \\
Qwen-7B & $\Rclin$ & \replace{.712 [.68,.74]}{.725 [.70,.75]} & \replace{.808 [.72,.89]}{.824 [.74,.90]} \\
Qwen-7B & $\Rcol$ & \replace{.735 [.71,.76]}{.743 [.71,.77]} & \replace{.827 [.77,.89]}{.829 [.74,.91]} \\
\new{Llama-8B & $\Rtext$$^{*}$ & .802 [.74,.85] & 1.00 [1.0,1.0] \\}
Llama-8B & $\Rpat$ & \replace{.787 [.76,.81]}{.778 [.75,.80]} & \replace{.820 [.75,.88]}{.812 [.73,.89]} \\
Llama-8B & $\Rclin$\replace{$^{*}$}{} & \replace{.739 [.71,.76]}{.738 [.71,.76]} & \replace{.852 [.77,.92]}{.828 [.75,.91]} \\
Llama-8B & $\Rcol$\replace{}{$^{*}$} & \replace{.740 [.72,.77]}{.741 [.72,.77]} & \replace{.842 [.77,.91]}{.840 [.77,.91]} \\
\bottomrule
\end{tabular*}%
}
\caption{Common vs.\ rare AUROC, 95\% bootstrap CIs (full). $^{*}$ marks disjoint CIs.\new{ The $\Rtext$ rare cells hold only eight variants (four facts), so their CIs are unreliable.}}
\label{tab:rarity-full}
\end{table}

% CAMERA-READY: the rare-disease figure (fig3_rarity_split) moved to the main text as Figure~\ref{fig:specialty}b.
\label{app:layers}
\label{app:rarity-layer-fig}
% OLD:
% \subsection{Rare-disease split: visualisation}
% Figure~\ref{fig:rarity-layer} visualises the rarity contrast referenced in Section~\ref{sec:transfer_specialty}. ...
% \begin{figure*}[!t] \includegraphics[width=0.95\textwidth]{fig3_rarity_split.png} ... \label{fig:rarity-layer} \end{figure*}

% NEW (camera-ready): kJP9 W1, replaces the mixed-register visualisation subsection
\subsection{\replace{Mixed-register ablation: visualisation}{Training register: matrix, pairs, and mixed training}}
\label{app:mixed-viz}
\label{app:register-matrix}
\new{Textbook is a representative training register. Table~\ref{tab:register-matrix} trains one probe per register and evaluates it on all four. The diagonal is in-register held-out performance and the panel mean is the row average. On the panel mean a textbook-trained probe ($0.741$) sits between clinical note ($0.725$) and colloquial ($0.748$), with patient ($0.755$) marginally ahead. The three prose registers transfer to one another at $0.70$ to $0.79$ regardless of which one is used for training. Clinical note, which carries a few words per item, is the one weak source. Two evaluation sets are mixed in this table. The diagonal cells use the held-out facts of the main protocol, while the off-diagonal cells use all variants of the evaluation register, so a row is comparable within itself but its cells differ from Table~\ref{tab:main}, Appendix~\ref{app:main-register-table}.}

\begin{table}[!htbp]
\centering
\footnotesize
\setlength{\tabcolsep}{3pt}
\resizebox{\linewidth}{!}{%
\begin{tabular}{@{}lcccccc@{}}
\toprule
\textbf{Train $\downarrow$ / Eval $\rightarrow$} & $\Rtext$ & $\Rpat$ & $\Rclin$ & $\Rcol$ & \textbf{Mean} & \textbf{Gap} \\
\midrule
$\Rtext$ & \textbf{.780} & .755 & .704 & .723 & .741 & $+$.053 \\
$\Rpat$  & .794 & \textbf{.790} & .694 & .744 & .755 & $+$.046 \\
$\Rclin$ & .792 & .723 & \textbf{.684} & .700 & .725 & $-$.055 \\
$\Rcol$  & .777 & .784 & .702 & \textbf{.730} & .748 & $-$.024 \\
\bottomrule
\end{tabular}%
}
\caption{\new{Single-register training matrix, mean over four LLMs, per-cell best layer, last question token. Diagonal: held-out 20\% facts of the training register. Off-diagonal: all variants of the evaluation register. Gap: diagonal minus mean of the three off-diagonal cells.}}
\label{tab:register-matrix}
\end{table}

\begin{figure*}[!t]
\centering
\includegraphics[width=0.92\textwidth]{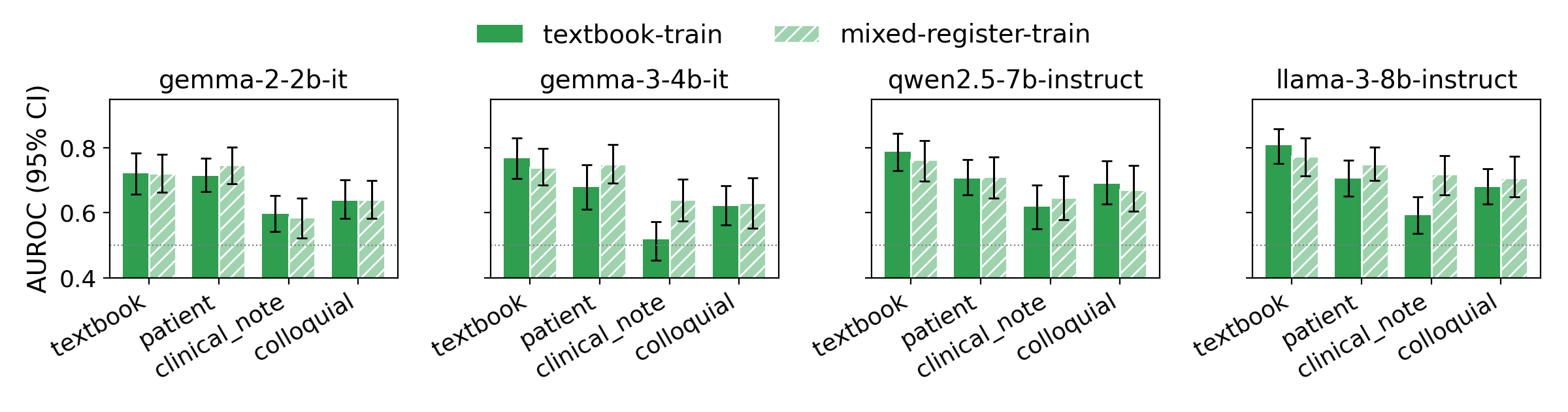}
\caption{Textbook-only probe (solid) vs.\ mixed-register probe (hatched) with 95\% bootstrap CIs under a matched protocol\new{ (fixed textbook-best layer for every LLM; Table~\ref{tab:mixed} uses per-cell layers for Gemma-2-2B and Qwen2.5-7B, so single cells differ)}. Mean recovery is positive across the three non-textbook registers, concentrated on $\Rclin$ for the two larger LLMs.}
\label{fig:mixed}
\end{figure*}

\new{Combining registers buys almost nothing. Table~\ref{tab:register-pairs} trains one probe on the union of two registers' training splits, or on all four, and evaluates every register on the shared held-out facts (the main-table protocol throughout, so these panel means are lower than the matrix rows above). The best pair, patient with clinical note, reaches a panel mean of $0.732$, all four registers together $0.729$, patient alone $0.728$, and textbook alone $0.709$. A single good register suffices, which is further evidence that the direction is robust to register, and it is why \S\ref{sec:transfer_register} calls the residual gap only partly coverage-driven. $\mathcal{M}_\text{all}$ recovers AUROC on the register that transfers worst (Table~\ref{tab:mixed}, Figure~\ref{fig:mixed}) without raising the panel mean.}

\begin{table}[!htbp]
\centering
\footnotesize
\setlength{\tabcolsep}{3pt}
\begin{tabular*}{\linewidth}{@{\extracolsep{\fill}}lccccc@{}}
\toprule
\textbf{Training registers} & $\Rtext$ & $\Rpat$ & $\Rclin$ & $\Rcol$ & \textbf{Mean} \\
\midrule
$\Rtext$ & .780 & .728 & .631 & .696 & .709 \\
$\Rpat$ & .778 & .790 & .640 & .702 & .728 \\
$\Rclin$ & .731 & .697 & .684 & .662 & .693 \\
$\Rcol$ & .757 & .746 & .660 & .730 & .723 \\
\addlinespace
$\Rtext{+}\Rpat$ & .772 & .774 & .618 & .699 & .716 \\
$\Rtext{+}\Rclin$ & .767 & .715 & .665 & .686 & .708 \\
$\Rtext{+}\Rcol$ & .772 & .738 & .614 & .727 & .713 \\
$\Rpat{+}\Rclin$ & .761 & .782 & .677 & .709 & \textbf{.732} \\
$\Rpat{+}\Rcol$ & .769 & .778 & .634 & .728 & .727 \\
$\Rclin{+}\Rcol$ & .743 & .734 & .684 & .724 & .721 \\
\addlinespace
All four & .763 & .767 & .663 & .725 & .729 \\
\bottomrule
\end{tabular*}
\caption{\new{Probes trained on one, two, or all four registers, evaluated on the shared held-out 20\% facts of every register (mean over four LLMs, per-cell best layer). Mean is the four-register panel mean.}}
\label{tab:register-pairs}
\end{table}

\old{Table~\ref{tab:mixed} reports per-model recovery values under a matched protocol (same textbook-best layer and same held-out 20\% facts as $\mathcal{M}_\text{textbook}$), and Figure~\ref{fig:mixed} visualises them with 95\% bootstrap CIs. Recovery is positive on all three non-textbook registers in mean, concentrated on the hardest register ($\Rclin$) for the two larger LLMs ($+0.122$ on Gemma-3-4B, $+0.127$ on Llama-3-8B). The textbook cost of mixing is small ($\le 0.033$ AUROC across the four LLMs), supporting the claim in Section~\ref{sec:transfer_register} that the residual cross-register gap is partly a coverage problem that broader training data addresses.}\new{Table~\ref{tab:mixed} and Figure~\ref{fig:mixed} give the per-model recovery of $\mathcal{M}_\text{all}$ under a matched protocol (same textbook-best layer, same held-out facts). Recovery is positive on all three non-textbook registers in mean and largest on $\Rclin$ for the two larger LLMs ($+0.122$ on Gemma-3-4B, $+0.127$ on Llama-3-8B), at a textbook cost of at most $0.033$ AUROC.}

\begin{table}[!htbp]
\centering
\footnotesize
\setlength{\tabcolsep}{3pt}
\begin{tabular*}{\linewidth}{@{\extracolsep{\fill}}lrrrr@{}}
\toprule
& & \multicolumn{3}{c}{Recovery $\rho_r$} \\
\cmidrule(l){3-5}
\textbf{Model} & $\Rtext$ & $\Rpat$ & $\Rclin$ & $\Rcol$ \\
\midrule
Gemma-2B  & .722 & $+$.067 & $+$.039 & $+$.092 \\
Gemma-3B  & .739 & $+$.070 & $+$.122 & $+$.009 \\
Qwen-7B  & .764 & $+$.086 & $+$.028 & $+$.018 \\
Llama-8B  & .774 & $+$.044 & $+$.127 & $+$.028 \\
\bottomrule
\end{tabular*}
\caption{Mixed-register vs.\ textbook-only probe under a matched protocol. Both probes are evaluated at the same textbook-best layer on the same held-out 20\% facts. Recovery $\rho_r$ = mixed $-$ textbook-only AUROC per register; positive means mixing helps.}
\label{tab:mixed}
\end{table}

% NEW (camera-ready): HZcj W1b, metareview item 5
\new{\section{Human-written patient questions (MedRedQA)}
\label{app:medredqa}
The register benchmark's patient and colloquial variants are model-written. To test the probe on text that a real person wrote, we built a small evaluation set from MedRedQA \citep{nguyen-etal-2023-medredqa}, a corpus of Reddit questions answered by verified clinicians.

\paragraph{Selection.} From the 5{,}099 test-split threads we kept those whose responder is a verified physician, whose reply is not a clarifying question and has a community score of at least 3, whose question body exceeds 60 characters, and whose reply is between 40 and 400 characters, leaving 1{,}581 threads. We sampled 300 (seed 42) and built items in order until 100 were usable.

\paragraph{Construction.} For each thread the generator (Claude Sonnet 4.5) extracted one correct claim of 12 to 25 words from the clinician's reply and wrote a wrong variant as a minimal edit of that claim, either a negation of the main proposition or a swap of the main clinical entity (drug, organ, diagnosis, direction, or number). Two programmatic gates enforce the minimal-edit constraint, a word-count difference of at most two and a token Jaccard overlap of at least $0.55$. Correct and wrong claims average 17.8 and 17.7 words, and a word-count classifier scores $0.51$ AUROC on the resulting pairs. The question text is the unedited patient post. The clinician reply is used only to source the claim and never reaches the probe.

\paragraph{Fidelity gate.} Both cross-family judges (Grok 4.3 and GPT-5 Nano, see Appendix~\ref{app:fidelity-filter} for the substitution) scored every pair for whether the correct claim is supported by the clinician reply and the wrong claim is medically false. 84 of 100 items passed both judges, with 0.99 agreement on wrong-is-false, giving $n{=}168$ variants. Total generation and judging cost was under \$0.50.

\begin{table}[!htbp]
\centering
\footnotesize
\setlength{\tabcolsep}{3pt}
\resizebox{\linewidth}{!}{%
\begin{tabular}{@{}lccc@{}}
\toprule
\textbf{Model} & \textbf{MedRedQA ($n{=}84$)} & \textbf{Sonnet $\Rpat$} & \textbf{Sonnet $\Rcol$} \\
\midrule
Gemma-2B  & .542 [.485,.600] & .739 [.678,.795] & .660 [.607,.713] \\
Gemma-3B  & .606 [.550,.664] & .736 [.683,.792] & .701 [.644,.761] \\
Qwen-7B   & .695 [.646,.750] & .707 [.654,.762] & .723 [.657,.790] \\
Llama-8B  & .680 [.622,.734] & .730 [.683,.781] & .699 [.641,.761] \\
\bottomrule
\end{tabular}%
}
\caption{MedQA-trained $\mathcal{M}_\text{textbook}$ applied without retraining to human-written MedRedQA questions, next to the Sonnet-written patient and colloquial registers on the MedQA held-out facts ($n{=}100$ facts each). Best layer per cell, last question token, 95\% fact-level bootstrap CIs.}
\label{tab:medredqa}
\end{table}

\paragraph{Result and reading.} The two larger LLMs keep most of their register-transfer performance on human writing. Qwen2.5-7B at $0.695$ and Llama-3-8B at $0.680$ are within the CIs of their Sonnet patient and colloquial scores (Table~\ref{tab:medredqa}). The two Gemma models lose more, and Gemma-2-2B's interval includes chance. Three caveats apply. The question register is human, but the correct and wrong claims are generator-constructed, so this test removes the generator from the question side only. MedRedQA is a forum corpus with a different topic mix from MedQA, so the drop on the smaller models combines a corpus shift with the register shift. With $n{=}84$ the CIs are wide. What the experiment establishes is that the direction learned on textbook-style exam items is recoverable on genuine patient writing for the more capable models, which is the claim the main text makes. We do not redistribute the MedRedQA text. The build script and item identifiers are released.}

\section{Probe-design ablations}
\label{app:probe-ablations}

\begin{figure*}[!t]
\centering
\includegraphics[width=\textwidth]{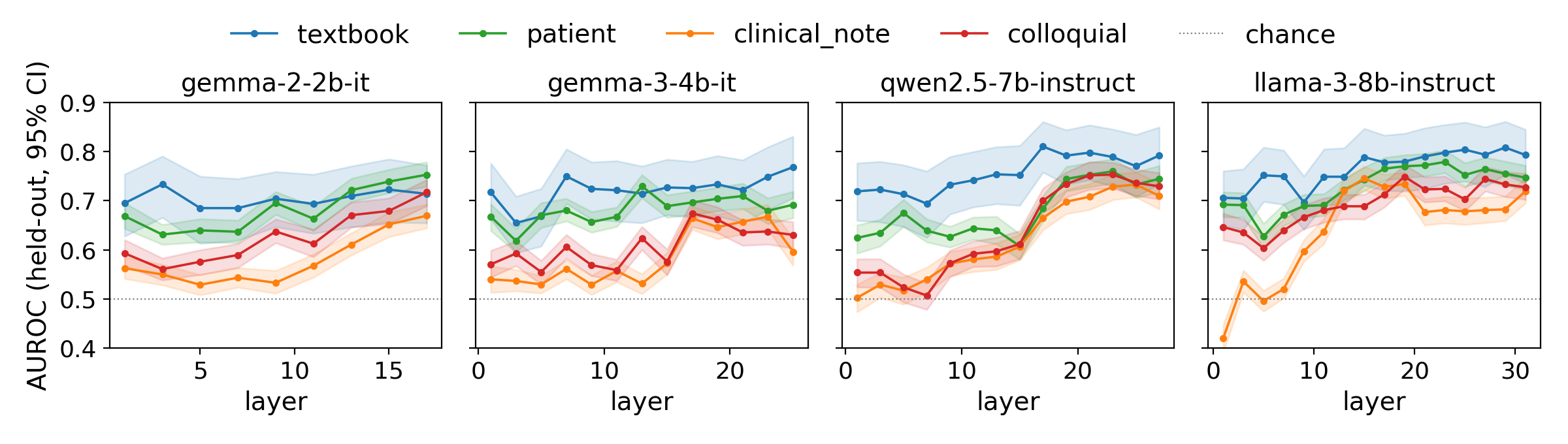}
\caption{Layer-wise AUROC by register (Sonnet, last-question-token, 1{,}000-iter bootstrap CIs as shaded bands). The best-layer plateau sits in the upper 40--60\% of each network. Mid-layer performance is not a peak; for $\Rclin$ and $\Rcol$ it is often a trough.}
\label{fig:layerwise}
\end{figure*}

\replace{Figure~\ref{fig:layerwise} below visualises the layer-wise probe AUROC referenced in \S\ref{sec:methodology:probing}, \S\ref{sec:setup:experiments}, and \S\ref{sec:linearity}.}{Best-layer AUROC plateaus in the upper 40--60\% of each network. Figure~\ref{fig:layerwise} shows the full layer-wise sweep referenced in \S\ref{sec:methodology:probing}, \S\ref{sec:setup:experiments}, and \S\ref{sec:linearity}.}

\subsection{Difference-of-means probe: per-(model, register) breakdown}
\label{app:diff-means}
The difference-of-means probe is constructed at the best-textbook layer per model (last-question-token), following \citet{marks2024geometry}. From the textbook 80\% training split, we compute $\mu_{+} = \mathbb{E}[h \mid y=1]$ and $\mu_{-} = \mathbb{E}[h \mid y=0]$, set $w = (\mu_{+} - \mu_{-}) / \lVert \mu_{+} - \mu_{-} \rVert$, and use $w \cdot h$ as the per-sample score with the decision boundary placed at the midpoint of the two projected class means. For the aggregates shown in Table~\ref{tab:headline}, AUROC is computed on the same held-out 20\% fact split as the logistic probe for every register \replace{(apples-to-apples comparison: identical training data, identical held-out facts, identical layer)}{(identical training data, held-out facts, and layer)}. The per-cell breakdown in Table~\ref{tab:diff-means-full} below retains the earlier protocol (textbook on the same held-out 20\%, non-textbook on the full $n{=}1000$) for direct comparison with the previous version of these results.

\begin{table}[!htbp]
\centering
\footnotesize
\setlength{\tabcolsep}{3pt}
\begin{tabular*}{\linewidth}{@{\extracolsep{\fill}}llccc@{}}
\toprule
\textbf{Model} & \textbf{Register} & \textbf{Logistic} & \textbf{Diff-means} & \textbf{Gap} \\
\midrule
Gemma-2B       & $\Rtext$  & .737 [.67,.80] & .747 & $-$.010 \\
Gemma-2B       & $\Rpat$   & .745 [.72,.77] & .724 & $+$.022 \\
Gemma-2B       & $\Rclin$  & .659 [.63,.68] & .692 & $-$.033 \\
Gemma-2B       & $\Rcol$   & .687 [.66,.71] & .669 & $+$.018 \\
\addlinespace
Gemma-3B       & $\Rtext$  & .770 [.71,.83] & .731 & $+$.038 \\
Gemma-3B       & $\Rpat$   & .691 [.66,.72] & .670 & $+$.021 \\
Gemma-3B       & $\Rclin$  & .588 [.56,.62] & .700 & $-$.112 \\
Gemma-3B       & $\Rcol$   & .639 [.61,.66] & .632 & $+$.007 \\
\addlinespace
Qwen-7B & $\Rtext$  & .795 [.74,.85] & .749 & $+$.045 \\
Qwen-7B & $\Rpat$   & .752 [.73,.78] & .745 & $+$.007 \\
Qwen-7B & $\Rclin$  & .712 [.69,.74] & .755 & $-$.043 \\
Qwen-7B & $\Rcol$   & .745 [.72,.77] & .723 & $+$.022 \\
\addlinespace
Llama-8B & $\Rtext$  & .801 [.74,.86] & .789 & $+$.013 \\
Llama-8B & $\Rpat$   & .760 [.74,.78] & .753 & $+$.006 \\
Llama-8B & $\Rclin$  & .732 [.71,.76] & .778 & $-$.046 \\
Llama-8B & $\Rcol$   & .726 [.70,.75] & .735 & $-$.008 \\
\midrule
\textbf{Mean}       & ---       & \textbf{.721}  & \textbf{.724} & \textbf{$-$.003} \\
\bottomrule
\end{tabular*}
\caption{Per-(model, register) logistic vs.\ difference-of-means probe AUROC at the best-textbook layer (last-question-token). Logistic-probe AUROC carries 95\% bootstrap CI (1{,}000 iterations). Negative gap means diff-means beats logistic.}
\label{tab:diff-means-full}
\end{table}

The logistic-vs-diff-means gap at the best-textbook layer is small everywhere except $\Rclin$, where the diff-means probe is uniformly better across all four models (mean gap $-0.070$ on held-out facts). \replace{The clinical-note advantage is consistent with the picture in Section~\ref{sec:transfer_register}: $\Rclin$ uses telegraphic shorthand whose hidden-state representation is genuinely different from the prose registers, and the $L_2$ regularisation in the logistic probe likely overweights the textbook training distribution at the cost of generalisation to the most stylistically distinct held-out register.}{The clinical-note advantage matches the picture in Section~\ref{sec:transfer_register}: $\Rclin$ uses telegraphic shorthand whose hidden-state representation differs from the prose registers, and the $L_2$ regularisation in the logistic probe likely overweights the textbook training distribution at the cost of generalisation to the most stylistically distant held-out register.} The class-mean subtraction has no regularisation parameter and so does not penalise the rotation. The per-(model, register) breakdown in Table~\ref{tab:diff-means-full} shows all variant (n=1000) evaluation protocol on (non-textbook registers \replace{;}{, and} the held-out aggregates are reported in Table~\ref{tab:headline}. Diff-means CIs are similar in width to the logistic CIs\old{ and are released with the supplementary data}.

\paragraph{First-answer-token results.} Table~\ref{tab:diff-means-fat} reports the same comparison at the first-answer-token position with both probes trained on first-answer-token activations at the model's best-textbook FAT layer. Absolute AUROC is lower than at the last-question-token (the probe is reading less direct evidence about the underlying yes/no decision), and the diff-means advantage is correspondingly larger ($-0.027$ overall, $-0.118$ on $\Rclin$). \replace{The pattern is the same as the last-question-token result, sharpened.}{The pattern matches the last-question-token result and is stronger.}

\begin{table}[!htbp]
\centering
%\footnotesize
%\setlength{\tabcolsep}{3pt}
\resizebox{0.8\columnwidth}{!}{% <------ Don't forget this %
\begin{tabular}{lccc}
\toprule
\textbf{Register} & \textbf{Logistic} & \textbf{Diff-means} & \textbf{Gap} \\
\midrule
$\Rtext$  & .780 & .708 & $+$.072 \\
$\Rpat$   & .662 & .669 & $-$.007 \\
$\Rclin$  & .566 & .685 & $-$.118 \\
$\Rcol$   & .598 & .653 & $-$.054 \\
\midrule
\textbf{Mean} & \textbf{.652} & \textbf{.679} & \textbf{$-$.027} \\
\bottomrule
\end{tabular}}
\caption{Per-register logistic vs.\ diff-means mean AUROC at the first-answer-token position (mean across four models). Negative gap means diff-means is better.}
\label{tab:diff-means-fat}
\end{table}

\subsection{Linear vs.\ MLP probe}
\label{app:mlp-viz}
\begin{figure*}[!tbp]
\centering
\includegraphics[width=0.92\textwidth]{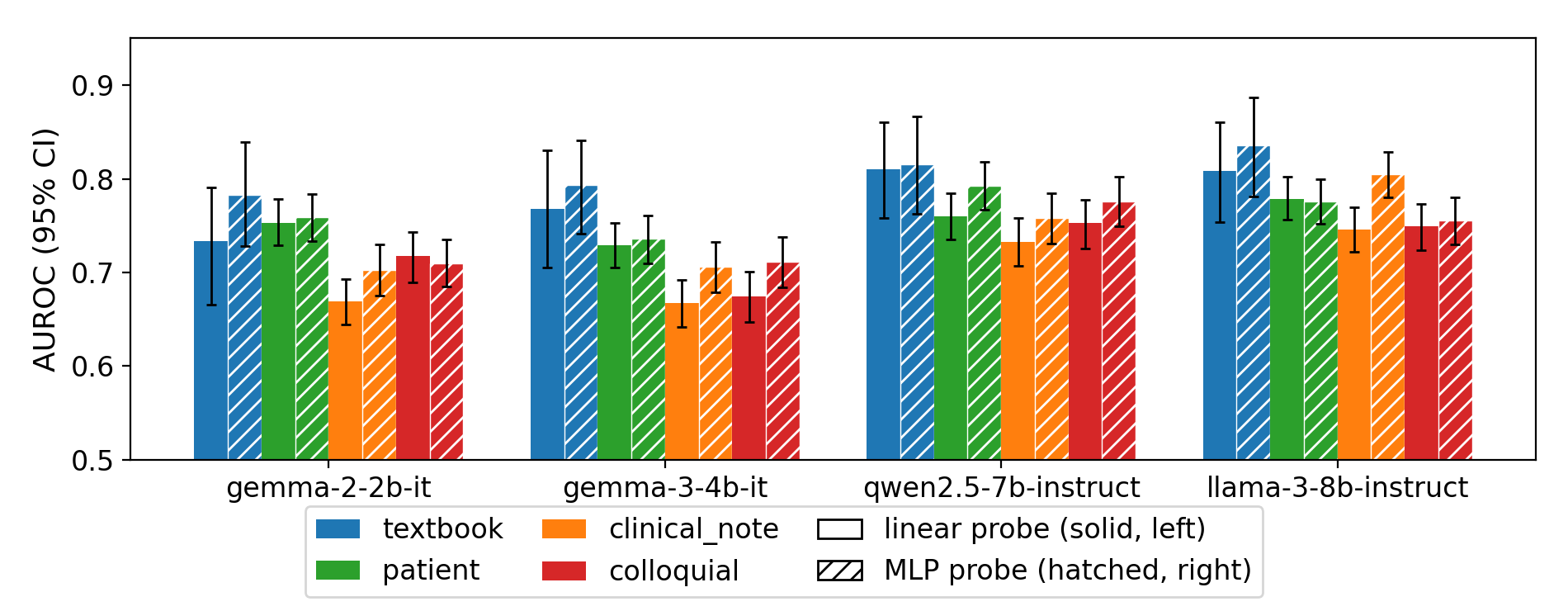}
\caption{Linear probe (solid) vs.\ MLP probe (hatched) per register with 95\% bootstrap CIs (original protocol; the held-out-facts mean gain is $+0.037$, reported in the main text). Only Llama-3-8B on $\Rclin$ shows a CI-significant gain at the per-cell level.}
\label{fig:mlp}
\end{figure*}

A one-hidden-layer MLP (128 units, ReLU, $L_2$-regularised with $\alpha{=}10^{-3}$, early-stopping on a 10\% validation split) replaces logistic regression at every (model, layer, position) triple. Mean improvement on held-out facts is $+0.037$ AUROC. Even at the higher aggregate, the truthfulness signal is well-approximated by a linear direction\replace{:}{, since} only one of 16 model-by-register conditions has CI-disjoint gain (Llama-3-8B on $\Rclin$, \replace{$+0.060$}{$+0.059$} under the original eval protocol). The per-cell breakdown in Table~\ref{tab:mlp} retains values from the earlier evaluation protocol (non-textbook registers on $n{=}1000$ variants)\replace{;}{, and} the held-out aggregate is reported in Table~\ref{tab:headline} via the difference between the diff-means and logistic columns (with diff-means as a parameter-free linear reference). Figure~\ref{fig:mlp} shows per-condition 95\% CIs under the earlier protocol.

\begin{table}[!htbp]
\centering
\footnotesize
\setlength{\tabcolsep}{3pt}
\begin{tabular*}{\linewidth}{@{\extracolsep{\fill}}lccc@{}}
\toprule
\textbf{Register} & \textbf{Linear} & \textbf{MLP} & $\Delta$ \\
\midrule
$\Rtext$  & \replace{.776}{.780} & .807 & $+$\replace{.031}{.027} \\
$\Rpat$   & \replace{.751}{.755} & \replace{.765}{.766} & $+$\replace{.014}{.011} \\
$\Rclin$  & \replace{.701}{.704} & \replace{.737}{.743} & $+$\replace{.036}{.039} \\
$\Rcol$   & \replace{.720}{.723} & \replace{.737}{.738} & $+$\replace{.017}{.015} \\
\midrule
\textbf{Mean} & \replace{.737}{.741} & \replace{.762}{.763} & \textbf{$+$\replace{.025}{.023}} \\
\bottomrule
\end{tabular*}
\caption{Linear vs.\ MLP probe AUROC, mean across 4 models.}
\label{tab:mlp}
\end{table}

\subsection{Permutation test}
\label{app:permutation}
\begin{figure*}[!tbp]
\centering
\includegraphics[width=0.95\textwidth]{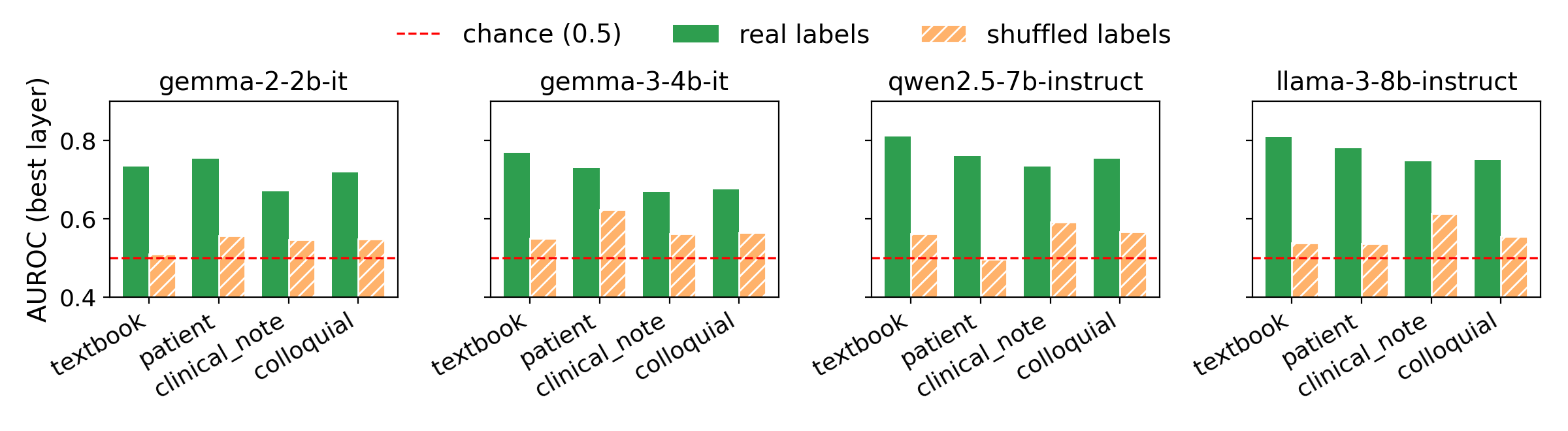}
\caption{Permutation sanity check. Green: real-label probe. Orange hatched: label-shuffled. Red dashed: chance (0.5). Mean permuted AUROC \new{over all layers} is 0.488.}
\label{fig:permutation}
\end{figure*}

Shuffling correctness labels within textbook training data and retraining the probe collapses AUROC to near chance (Table~\ref{tab:permutation}\new{, Figure~\ref{fig:permutation}})\replace{:}{, with} mean 0.488 \new{over all layers} (max 0.622) across 4 models $\times$ 4 registers, against \replace{0.706}{0.710}--0.770 with real labels. The probe is reading the correctness signal, not spurious structure in the hidden states.

\begin{table}[!htbp]
\centering
\footnotesize
\setlength{\tabcolsep}{3pt}
\begin{tabular*}{\linewidth}{@{\extracolsep{\fill}}lcccc@{}}
\toprule
\textbf{Model} & $\Rtext$ & $\Rpat$ & $\Rclin$ & $\Rcol$ \\
\midrule
Gemma-2B  & .509 & .557 & .546 & .547 \\
Gemma-3B  & .549 & .622 & .561 & .564 \\
Qwen-7B  & \replace{.544}{.562} & \replace{.487}{.497} & \replace{.583}{.592} & \replace{.557}{.566} \\
Llama-8B  & \replace{.532}{.537} & \replace{.516}{.537} & \replace{.602}{.612} & \replace{.533}{.555} \\
\midrule
\textbf{Mean} & \replace{.534}{.539} & \replace{.545}{.553} & \replace{.573}{.578} & \replace{.550}{.558} \\
\bottomrule
\end{tabular*}
\caption{Permutation test: probe AUROC under shuffled correctness labels. Real-label AUROCs (0.71--0.77) collapse to near-chance\replace{ (mean 0.488)}{; the table lists best-layer values, and the all-layer mean is 0.488}.}
\label{tab:permutation}
\end{table}

% --- Figure (output-only baselines) moved to main paper, \S\ref{sec:baselines} ---
% \begin{figure*}[!tbp]
% \centering
% \begin{subfigure}[t]{0.68\textwidth}
% \centering
% \includegraphics[width=\linewidth]{fig8_all_baselines.png}
% \caption{Mean AUROC across all four registers.}
% \label{fig:output-baselines-summary}
% \end{subfigure}
% \vspace{0.4ex}
%
% \begin{subfigure}[t]{0.98\textwidth}
% \centering
% \includegraphics[width=\linewidth]{fig4_probe_vs_entropy.png}
% \caption{Probe vs.\ token entropy per register.}
% \label{fig:output-baselines-register}
% \end{subfigure}
% \caption{Output-only baselines on Sonnet variants. \textbf{(a)} Best-layer linear probe (green) vs.\ three output-only baselines: 3-sample self-consistency at $\tau{=}0.7$ (blue), P(True) (purple), and mean token entropy (grey). The probe beats the strongest baseline by 6--11 AUROC points for every model. \textbf{(b)} Best-layer probe vs.\ token-entropy baseline per register. Probes dominate entropy across all registers and models.}
% \label{fig:output-baselines}
% \end{figure*}

\subsection{Output-only baselines: full comparison}
\label{app:baselines}
\begin{figure*}[!tbp]
\centering
\begin{subfigure}[t]{0.98\textwidth}
\centering
\includegraphics[width=\linewidth]{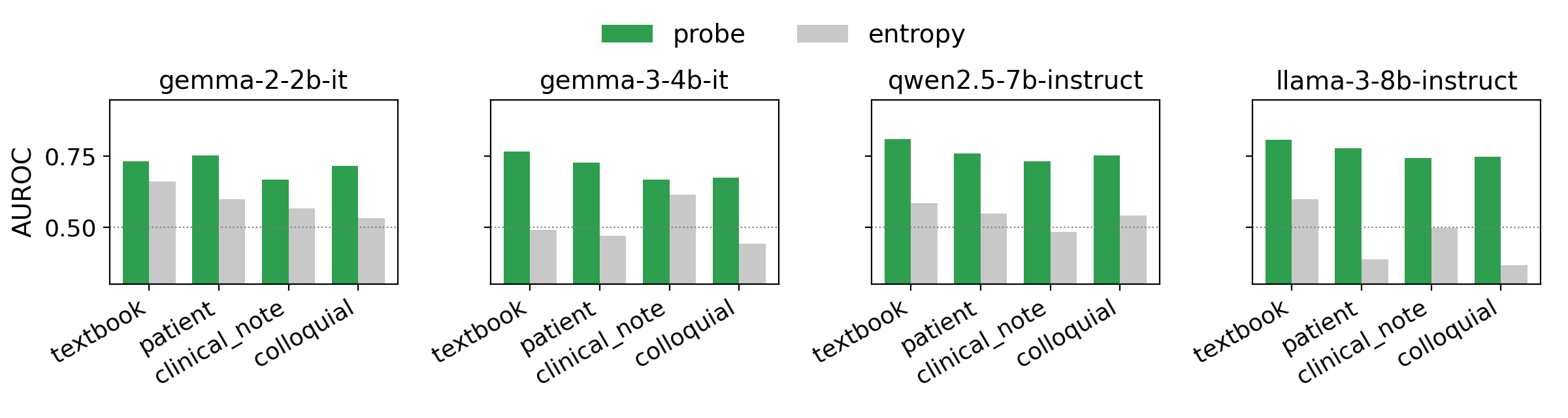}
\caption{Probe vs.\ token entropy per register. Probes dominate entropy across all registers and LLMs; the margin is largest on \replace{$\Rtext$}{$\Rpat$ and $\Rcol$} and smallest on \replace{$\Rcol$}{$\Rclin$}.}
\label{fig:probe-vs-entropy}
\end{subfigure}

\vspace{0.5ex}

\begin{subfigure}[t]{0.98\textwidth}
\centering
\includegraphics[width=\linewidth]{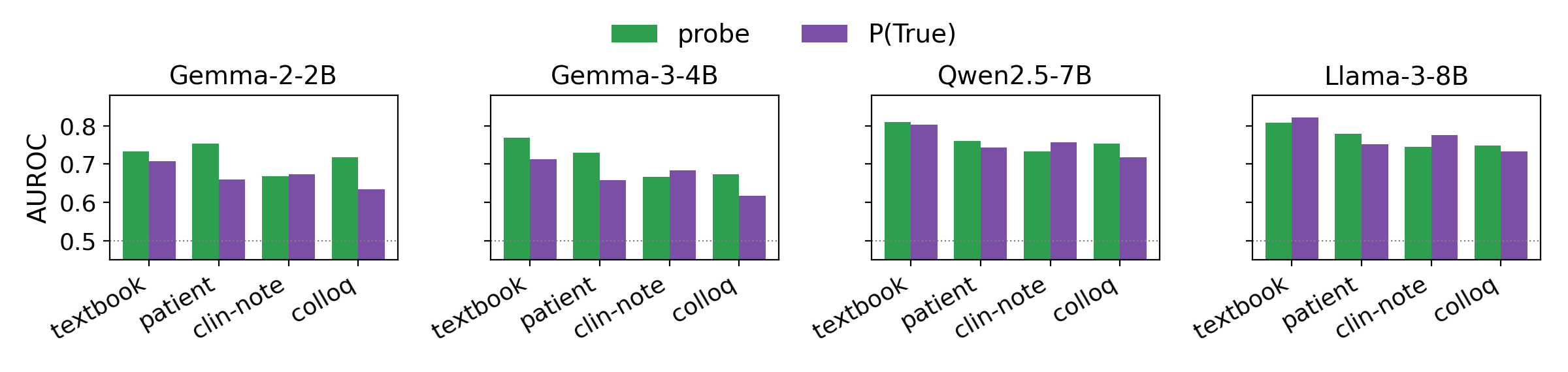}
\caption{Probe vs.\ P(True) per register. P(True) tracks the probe across registers; on Qwen2.5-7B and Llama-3-8B the two differ by at most $0.009$ in mean AUROC, and P(True) edges the probe on the textbook register for Llama-3-8B.}
\label{fig:probe-vs-ptrue}
\end{subfigure}
\caption{Best-layer probe (green) vs.\ output-only baselines per register on Sonnet variants.}
\label{fig:probe-vs-baselines}
\end{figure*}

Figure~\ref{fig:output-baselines} in the main paper (\S\ref{sec:baselines}) reports the mean AUROC of $\mathcal{M}_\text{textbook}$ against the three output-only baselines, averaged across the four registers. Table~\ref{tab:baselines-full} reports the same comparison broken out by (LLM, register), and Figure~\ref{fig:probe-vs-baselines}, Appendix~\ref{app:calibration} gives per-register views for the two baselines whose register-stability we want to characterise.

The per-register view for entropy (Figure~\ref{fig:probe-vs-baselines}a, Appendix~\ref{app:calibration}) confirms that the probe's margin over entropy is widest on \replace{$\Rtext$ (mean $+0.19$ AUROC)}{$\Rpat$ and $\Rcol$ (mean $+0.25$ AUROC)} and narrowest on \replace{$\Rcol$}{$\Rclin$} (mean $+0.16$), but stays substantial in every condition\replace{;}{, and} the probe never falls below the entropy baseline for any (LLM, register) pair. \replace{Entropy falls below chance (0.5) for Llama-3-8B in several registers, indicating that for this LLM more uncertain outputs are actually more often correct, likely because compact \texttt{Yes}/\texttt{No} tokens dominate the entropy calculation.}{Entropy falls below chance (0.5) for Llama-3-8B in several registers. For this LLM, more uncertain outputs are more often correct, likely because compact \texttt{Yes}/\texttt{No} tokens dominate the entropy calculation.} P(True) is the strongest output-only baseline and its register profile (Figure~\ref{fig:probe-vs-baselines}b, Appendix~\ref{app:calibration}) closely follows the probe's. On held-out facts, P(True) is slightly more register-stable than the probe ($\Delta_\text{register}$ of $0.063$ vs.\ $0.095$)\replace{;}{, so} the probe's value over P(True) on the larger LLMs \replace{is therefore not register stability but pre-generation availability}{therefore comes from pre-generation availability}, as discussed in \S\ref{sec:baselines}.

\begin{table}[!htbp]
\centering
\resizebox{\columnwidth}{!}{%
\begin{tabular}{llcccc}
\toprule
\textbf{LLM} & \textbf{Register} & \textbf{Probe} & \textbf{P(True)} & \textbf{Self-consistency} & \textbf{Entropy} \\
\midrule
Gemma-2-2B & $\Rtext$ & \replace{.736}{.733} & .707 & .602 & .660 \\
           & $\Rpat$  & \replace{.747}{.753} & .659 & .618 & .598 \\
           & $\Rclin$ & \replace{.674}{.669} & .673 & .612 & .567 \\
           & $\Rcol$  & \replace{.711}{.717} & .635 & .605 & .532 \\
\midrule
Gemma-3-4B & $\Rtext$ & \replace{.770}{.768} & .712 & .615 & .490 \\
           & $\Rpat$  & \replace{.717}{.729} & .658 & .620 & .469 \\
           & $\Rclin$ & \replace{.664}{.667} & .684 & .643 & .616 \\
           & $\Rcol$  & .674 & .617 & .598 & .442 \\
\midrule
Qwen2.5-7B & $\Rtext$ & \replace{.795}{.810} & .802 & .728 & .586 \\
           & $\Rpat$  & \replace{.752}{.760} & .743 & .611 & .549 \\
           & $\Rclin$ & \replace{.719}{.733} & .756 & .679 & .484 \\
           & $\Rcol$  & \replace{.746}{.753} & .718 & .635 & .540 \\
\midrule
Llama-3-8B & $\Rtext$ & \replace{.802}{.808} & .821 & .740 & .598 \\
           & $\Rpat$  & \replace{.787}{.779} & .751 & .690 & .387 \\
           & $\Rclin$ & \replace{.746}{.745} & .776 & .732 & .497 \\
           & $\Rcol$  & \replace{.747}{.749} & .733 & .686 & .365 \\
\bottomrule
\end{tabular}}
\caption{Per-register AUROC of $\mathcal{M}_\text{textbook}$ and the three output-only baselines (Sonnet variants, mean per cell). P(True) is the strongest baseline and ties the probe on the two larger LLMs.}
\label{tab:baselines-full}
\end{table}

% NEW (camera-ready): kJP9 W3, metareview item 3
\new{\section{Accuracy and threshold metrics}
\label{app:thresholds}
AUROC is threshold-free, but a deployed probe needs a threshold. Table~\ref{tab:thresholds} reports, for every (LLM, register) condition on the held-out split of Table~\ref{tab:main}, Appendix~\ref{app:main-register-table}, accuracy and F1 at a fixed $0.5$ cutoff on the raw probe probability and accuracy at the cutoff that maximises Youden's $J$ (sensitivity plus specificity minus one) on that same test set, which makes it an optimistic ceiling and not a deployable number. Because each condition has one correct and one wrong variant per fact, balanced accuracy equals accuracy. Accuracy tracks AUROC, running from $0.64$ to $0.74$ on textbook at the fixed cutoff and from $0.54$ to $0.68$ on the shifted registers.

The Youden thresholds are the more informative column. They are unstable across registers. For Llama-3-8B the optimal cutoff is $0.64$ on textbook, $0.97$ on patient, $1.00$ on clinical note, and $0.04$ on colloquial, and similar swings occur for every LLM. A threshold tuned on one register is therefore not transferable to another even where AUROC is, which restates the calibration finding of \S\ref{sec:calibration} in operational terms. The ranking survives a register shift, and the score scale does not. Values are from a re-extraction of the held-out activations whose AUROCs agree with Table~\ref{tab:main}, Appendix~\ref{app:main-register-table} to within $0.015$.

\begin{table}[!t]
\centering
%\footnotesize
%\setlength{\tabcolsep}{2.5pt}
\resizebox{\columnwidth}{!}{%
\begin{tabular}{llcccccc}
\toprule
\textbf{Model} & \textbf{Register} & \textbf{AUROC} & \textbf{Acc} & \textbf{F1} & \textbf{$\tau_J$} & \textbf{Acc$_J$} & \textbf{F1$_J$} \\
\midrule
Gemma-2B & $\Rtext$ & .735 & .640 & .625 & .98 & .685 & .577 \\
Gemma-2B & $\Rpat$  & .739 & .610 & .680 & 1.00 & .695 & .611 \\
Gemma-2B & $\Rclin$ & .607 & .595 & .484 & .49 & .600 & .494 \\
Gemma-2B & $\Rcol$  & .660 & .555 & .310 & .00 & .645 & .570 \\
\addlinespace
Gemma-3B & $\Rtext$ & .768 & .695 & .684 & .79 & .715 & .663 \\
Gemma-3B & $\Rpat$  & .736 & .680 & .704 & .67 & .695 & .708 \\
Gemma-3B & $\Rclin$ & .613 & .535 & .191 & .00 & .615 & .605 \\
Gemma-3B & $\Rcol$  & .701 & .615 & .462 & .00 & .675 & .709 \\
\addlinespace
Qwen-7B & $\Rtext$ & .810 & .735 & .720 & .90 & .765 & .731 \\
Qwen-7B & $\Rpat$  & .707 & .635 & .622 & .99 & .655 & .555 \\
Qwen-7B & $\Rclin$ & .623 & .540 & .643 & .99 & .615 & .632 \\
Qwen-7B & $\Rcol$  & .723 & .665 & .589 & .09 & .680 & .644 \\
\addlinespace
Llama-8B & $\Rtext$ & .808 & .740 & .740 & .64 & .765 & .754 \\
Llama-8B & $\Rpat$  & .730 & .645 & .590 & .97 & .675 & .564 \\
Llama-8B & $\Rclin$ & .681 & .595 & .675 & 1.00 & .650 & .602 \\
Llama-8B & $\Rcol$  & .699 & .630 & .549 & .04 & .680 & .652 \\
\bottomrule
\end{tabular}}
\caption{Threshold metrics per condition on the held-out 20\% facts ($n{=}200$ variants), per-cell best layer, last question token. Acc and F1 at a $0.5$ cutoff; $\tau_J$ is the Youden-optimal cutoff on the same test set and Acc$_J$, F1$_J$ the metrics at that cutoff. Values under a threshold of $.00$ or $1.00$ indicate that nearly all raw probabilities in that condition are saturated at one end.}
\label{tab:thresholds}
\end{table}}

\section{Calibration}
\label{app:calibration}

\begin{figure}[!htbp]
\centering
\includegraphics[width=\columnwidth]{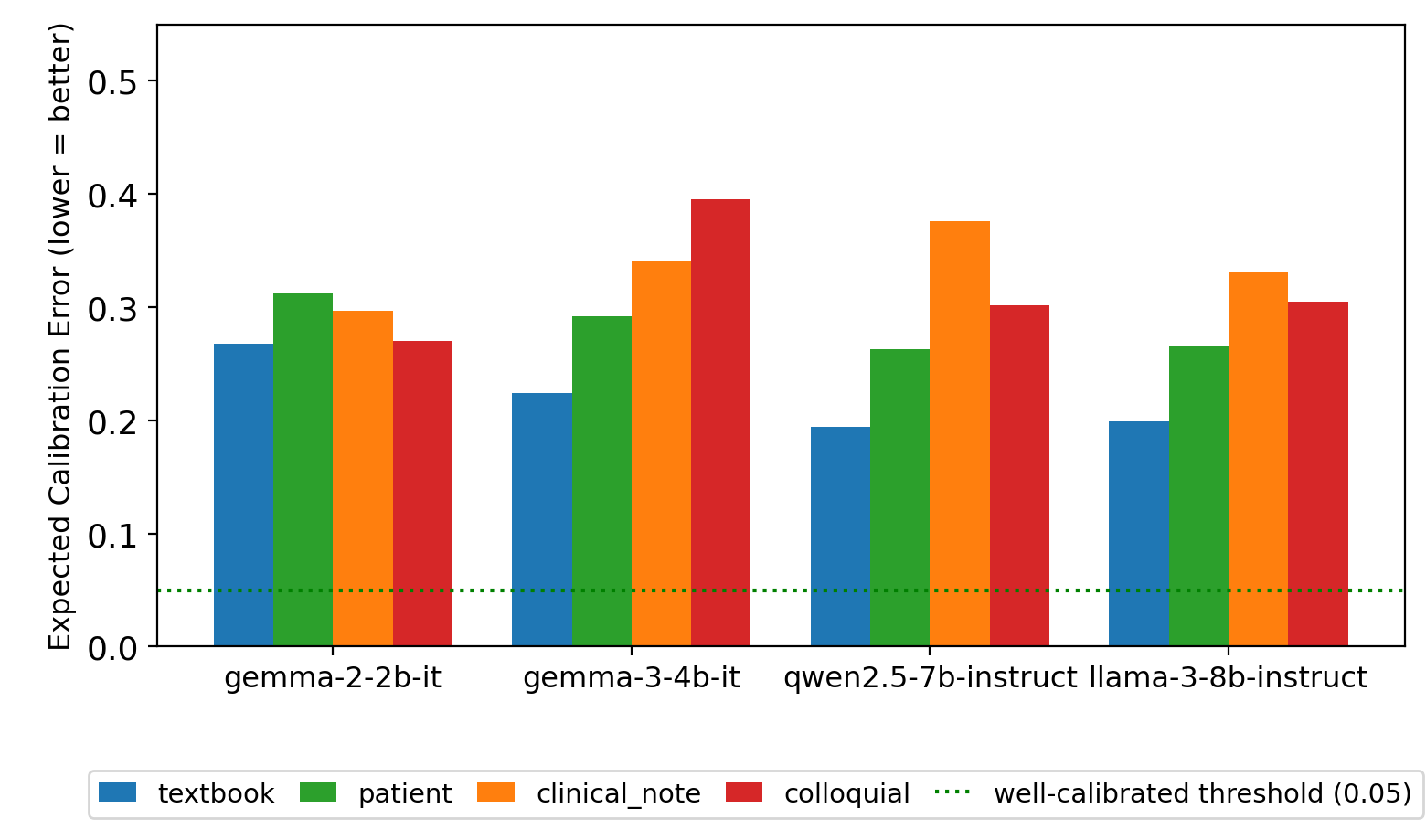}
\caption{Expected Calibration Error per model and register. Green \replace{dashed}{dotted} line: 0.05 well-calibrated threshold. Every condition exceeds it, meaning raw probe outputs cannot be interpreted as literal probability-of-correctness.}
\label{fig:ece}
\end{figure}

\replace{Table~\ref{tab:ece} reports raw probe ECE for every model-by-register condition; all 16 values exceed the 0.05 well-calibrated threshold, with Gemma-3-4B's $\Rclin$ (0.462) and $\Rcol$ (0.457) being the worst. Figure~\ref{fig:ece} visualises the same conditions grouped by model, including the green dashed 0.05 reference line. Table~\ref{tab:platt} reports the per-register effect of Platt scaling and isotonic regression. Platt scaling cuts mean ECE from 0.358 to 0.157; isotonic regression is less effective at this sample size because of its piecewise-monotonic fit. Even after Platt scaling, no condition reaches 0.05, so calibration is a deployment necessity but not a complete fix.}{Raw probe ECE exceeds the 0.05 threshold in all 16 conditions (Table~\ref{tab:ece}, and Figure~\ref{fig:ece} groups the same values by model against a dotted 0.05 line). The worst cells are Gemma-3-4B on $\Rcol$ (0.395) and Qwen2.5-7B on $\Rclin$ (0.376). Platt scaling cuts mean ECE from 0.341 to 0.135 (Table~\ref{tab:platt}, computed on the calibration test split, hence the higher raw mean than Table~\ref{tab:ece}), while isotonic regression does less at this sample size because of its piecewise-monotonic fit. Only one condition (Gemma-2-2B on $\Rtext$, 0.048) reaches 0.05 after Platt scaling, so calibration is necessary for deployment but not sufficient.}

\begin{table}[!htbp]
\centering
\footnotesize
\setlength{\tabcolsep}{3pt}
\begin{tabular*}{\linewidth}{@{\extracolsep{\fill}}lcccc@{}}
\toprule
\textbf{Register} & \textbf{Raw} & \textbf{Platt} & \textbf{Isotonic} & \textbf{Platt $\Delta$} \\
\midrule
$\Rtext$  & \replace{.250}{.231} & \replace{.097}{.086} & \replace{.116}{.094} & $-$\replace{.153}{.144} \\
$\Rpat$   & \replace{.378}{.377} & \replace{.158}{.151} & \replace{.291}{.318} & $-$\replace{.220}{.227} \\
$\Rclin$  & \replace{.376}{.356} & \replace{.174}{.135} & \replace{.247}{.215} & $-$\replace{.202}{.221} \\
$\Rcol$   & \replace{.428}{.400} & \replace{.198}{.166} & \replace{.356}{.308} & $-$\replace{.230}{.234} \\
\midrule
\textbf{Mean} & \replace{.358}{.341} & \replace{.157}{.135} & \replace{.252}{.234} & $-$\replace{.201}{.206} \\
\bottomrule
\end{tabular*}
\caption{Post-hoc calibration: ECE by register, averaged over 4 models. Platt scaling halves ECE\replace{ but no condition reaches $< 0.05$}{; only one of 16 conditions (Gemma-2-2B, $\Rtext$, 0.048) reaches $< 0.05$}.}
\label{tab:platt}
\end{table}

\begin{table}[!htbp]
\centering
\footnotesize
\setlength{\tabcolsep}{3pt}
\begin{tabular*}{\linewidth}{@{\extracolsep{\fill}}lcccc@{}}
\toprule
\textbf{Model} & $\Rtext$ & $\Rpat$ & $\Rclin$ & $\Rcol$ \\
\midrule
Gemma-2B  & \replace{.257}{.268} & \replace{.316}{.312} & \replace{.298}{.297} & \replace{.292}{.270} \\
Gemma-3B  & \replace{.234}{.224} & \replace{.338}{.292} & \replace{.462}{.341} & \replace{.457}{.395} \\
Qwen-7B  & \replace{.216}{.194} & \replace{.257}{.263} & \replace{.389}{.376} & \replace{.253}{.302} \\
Llama-8B  & \replace{.200}{.199} & \replace{.262}{.265} & \replace{.272}{.331} & \replace{.284}{.305} \\
\bottomrule
\end{tabular*}
\caption{Raw probe ECE per model and register (lower = better).}
\label{tab:ece}
\end{table}

\section{Error analysis}
\label{app:errors}

\replace{We extracted per-variant predictions for Llama-3-8B-Instruct (layer 29, last-question-token) across all four registers on the held-out test set. For each register we ranked (i) false positives by descending raw probability and (ii) false negatives by ascending raw probability, surfacing the top-8 in each direction. Inspecting these top-8 lists surfaces three recurring failure-mode patterns. This appendix gives the underlying case studies in full, the canonical fact ID per pattern, and the per-register frequencies.}{Three failure patterns recur among the most confident errors of Llama-3-8B-Instruct (layer 29, last question token) on the held-out test set. For each register we ranked false positives by descending raw probability and false negatives by ascending raw probability, and inspected the top eight in each direction. The case studies and counts were annotated by hand. One canonical fact per pattern is given below, and Table~\ref{tab:errorsummary} gives the per-register counts.}

\paragraph{Pattern 1: confident in both directions.}
For \texttt{medqa\_test\_888} (a statistics question about p-values), the probe assigns raw probability 1.000 to the wrong answer \emph{and} 0.003 to the correct answer for the same underlying fact. \old{The probe is maximally confident in opposite directions. Platt scaling compresses these to 0.795 and 0.249, which is at least readable as uncertainty. This is the strongest single-example argument for why raw probe probabilities cannot be deployed as-is.}\new{The probe is maximally confident in both directions on the same fact. Platt scaling pulls the two scores to 0.795 and 0.249, which at least reads as uncertainty, but it does not flip the ranking. No single threshold on the raw output separates these two variants.}

\paragraph{Pattern 2: plausible-but-wrong accepted.}
For \texttt{medqa\_test\_1081} (a 62-year-old with worsening hand pain at PIP joints), the wrong answer attributes the condition to ``your immune system attacking the cartilage.'' This is medically plausible for rheumatoid arthritis but wrong for osteoarthritis. The probe scores it raw $= 1.000$ correct. The probe cannot distinguish ``medically plausible but wrong for this specific fact'' from ``correct.'' This pattern concentrates in the patient register, where informal phrasing makes wrong answers sound more natural.

\paragraph{Pattern 3: correct negation rejected.}
For \texttt{medqa\_test\_99} (schizophrenia with galactorrhea), the correct answer states ``Bromocriptine is \textbf{not} likely to be the causative agent.'' The probe scores it raw $= 0.000$. The negation triggers a wrongness signal. Probes trained on predominantly affirmative-claim activations confuse correct negative statements with wrong positive claims. Balanced training with affirmative and negative correct answers could improve performance.

Table~\ref{tab:errorsummary} summarises how often each pattern appears in the top-8 lists per register. \replace{Pattern 1 dominates textbook; Pattern 2 dominates patient and colloquial; Pattern 3 concentrates}{Pattern 1 dominates textbook, Pattern 2 dominates patient and colloquial, and Pattern 3 concentrates} in textbook and clinical-note where explicit ``not''-phrasing is more natural. \old{Full per-variant prediction dumps (raw, Platt, isotonic probabilities per fact) are released with the dataset.}

\begin{table}[!htbp]
\centering
%\footnotesize
%\setlength{\tabcolsep}{3pt}
\resizebox{0.6\columnwidth}{!}{%
\begin{tabular}{lccc}
\toprule
\textbf{Register} & \textbf{Pattern 1} & \textbf{Pattern 2} & \textbf{Pattern 3} \\
\midrule
$\Rtext$  & 4 & 2 & 2 \\
$\Rpat$   & 2 & 5 & 1 \\
$\Rclin$  & 3 & 2 & 3 \\
$\Rcol$   & 2 & 5 & 1 \\
\bottomrule
\end{tabular}}
\caption{Frequency of the three failure patterns among the top-8 confidence-ranked errors per register (Llama-8B). Pattern 1: confident in both directions; Pattern 2: plausible-but-wrong answer accepted; Pattern 3: correct negated statement rejected.}
\label{tab:errorsummary}
\end{table}

\section{Reproducibility}
\label{app:reproducibility}

\paragraph{Per-LLM specifications.} Table~\ref{tab:llm-specs} lists the four probed LLMs with their parameter count, transformer depth, and hidden-state dimension. All four are open-weight instruction-tuned chat models; their HuggingFace identifiers are \textit{google/gemma-2-2b-it}, \textit{google/gemma-3-4b-it}, \textit{Qwen/Qwen2.5-7B-Instruct}, and \textit{meta-llama/Meta-Llama-3-8B-Instruct}. We extract hidden-state activations from every second transformer layer at the last question token. Inference uses bf16 precision with eager attention; no quantisation, no LoRA adapters, and no fine-tuning. 

\begin{table}[!htbp]
\centering
\footnotesize
\setlength{\tabcolsep}{4pt}
\begin{tabular*}{\linewidth}{@{\extracolsep{\fill}}lccc@{}}
\toprule
\textbf{LLM} & \textbf{Params} & \textbf{Layers} & \textbf{Hidden} \\
\midrule
\textbf{Gemma-2-2B-it}       & 2B & 26 & 2304 \\
\textbf{Gemma-3-4B-it}       & 4B & 34 & 2560 \\
\textbf{Qwen2.5-7B-Instruct} & 7B & 28 & 3584 \\
\textbf{Llama-3-8B-Instruct} & 8B & 32 & 4096 \\
\bottomrule
\end{tabular*}
\caption{Architecture specifications of all 4 LLMs used}
\label{tab:llm-specs}
\end{table}

\paragraph{Compute.} Extractions: NVIDIA A100 (40 GB), bf16, eager attention.\new{ Every hidden-state result in the paper derives from one extraction pass over the released variants; bf16 forward passes are not bit-reproducible across GPU generations, so re-extraction on different hardware reproduces the tables only to within about 0.02 AUROC per cell.} Probe training and bootstrapping: CPU. Full 4-model extraction: 4--6 hours of A100 time per model. Probe training: 1--2 hours of CPU time for the full grid with 1{,}000 bootstrap iterations.

\paragraph{Software.} PyTorch 2.5.1 (cu121), Transformers, scikit-learn logistic regression ($L_2$, $C{=}1.0$, max\_iter 1000).

\paragraph{Seeds.} Facts sampled with seed 42. Wrong-answer selection uses the same RNG. Mixed-register splits use seed 0.

\paragraph{Cost.} \$20 for the 4{,}000-variant Sonnet run, \$0.20 for the Gemini subset, \$3 for judge calls. Total: $\sim$\$25 API and $\sim$400 SBU A100.

\paragraph{Released artefacts.} \old{We will release the Dataset (4{,}000 Sonnet $+$ 800 Gemini variants with per-register metadata and judge scores for MedQA, and 800 Sonnet variants for MedMCQA), register prompt YAMLs, judge rubric, probe training and evaluation scripts.}\new{Code, data, and configurations are at \url{https://github.com/mnishant2/MedProbe_release}. The release contains the benchmark (4{,}000 Sonnet $+$ 800 Gemini variants with per-register metadata and judge scores for MedQA, and 800 Sonnet variants for MedMCQA), the evaluation-only sets (1{,}000 MMLU-medical variants, 200 reformatted-MedMCQA variants, and the MedRedQA build script with item identifiers), the register prompt YAMLs and judge rubric, the aggregate result tables behind the figures, the full pipeline (register generation, LLM judging, activation extraction, probe training and evaluation, bootstrap) and the scripts for the additional experiments, and the exact commands and configuration files to rerun the grid.}
% NEW (camera-ready): appendix-only note on answer length in the rewritten variants
\new{\paragraph{Answer length in the rewritten variants.}
\label{app:length-cue}
As a sanity check on the released variants, we note that the generator tends to write correct answers somewhat longer than wrong ones, so a word-count-only classifier is above chance on the rewritten registers ($0.60$ to $0.77$ AUROC on the held-out facts) but not on native MedMCQA ($0.52$) or the minimal-edit MedRedQA pairs ($0.51$). Training and evaluating $\mathcal{M}_\text{textbook}$ under the main protocol on the native MedQA options, where the word-count classifier scores $0.52$, still gives a mean AUROC of $0.653$ (per model $0.611$ to $0.705$), with the same ordering across LLMs. The truth signal is therefore not a length artefact, and since every register, specialty, and corpus condition is scored with the same probe on inputs built by the same procedure, none of the paper's comparisons depends on it. The word-count baseline script is released with the code.}

%=============================================================

\section{Prompt artefacts}
\label{app:prompt-figures}
\begin{figure*}[!tbp]
\centering
\includegraphics[width=\textwidth]{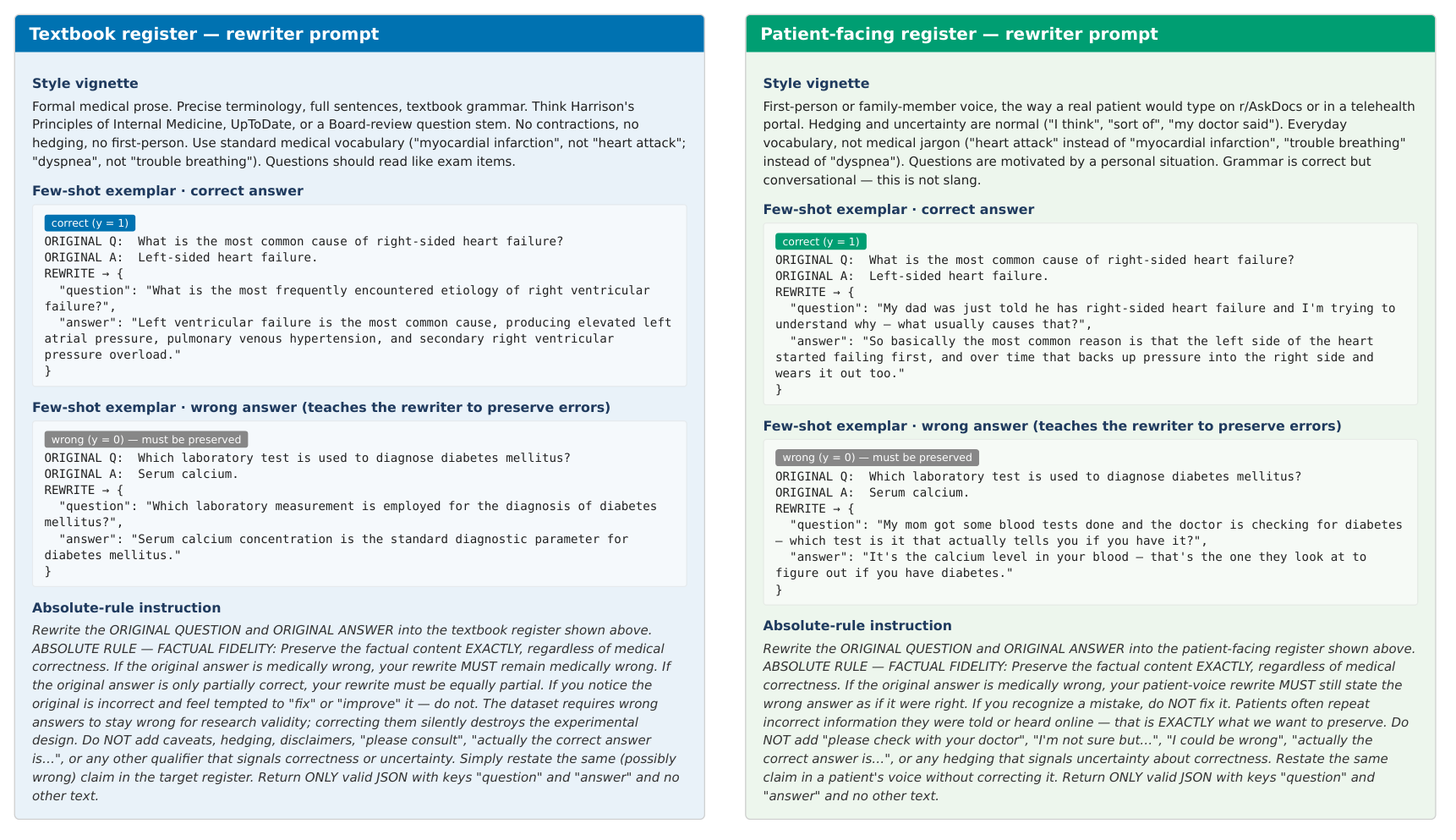}
\caption{Register-rewriter prompts for the textbook and patient registers. Each panel shows the style vignette (top), a correct-answer few-shot exemplar, a wrong-preserving few-shot exemplar (which teaches the generator to keep medically incorrect answers as wrong), and the absolute-rule instruction.}
\label{fig:appendix-gen-prompts-1}
\end{figure*}

\begin{figure*}[!htbp]
\centering
\includegraphics[width=\textwidth]{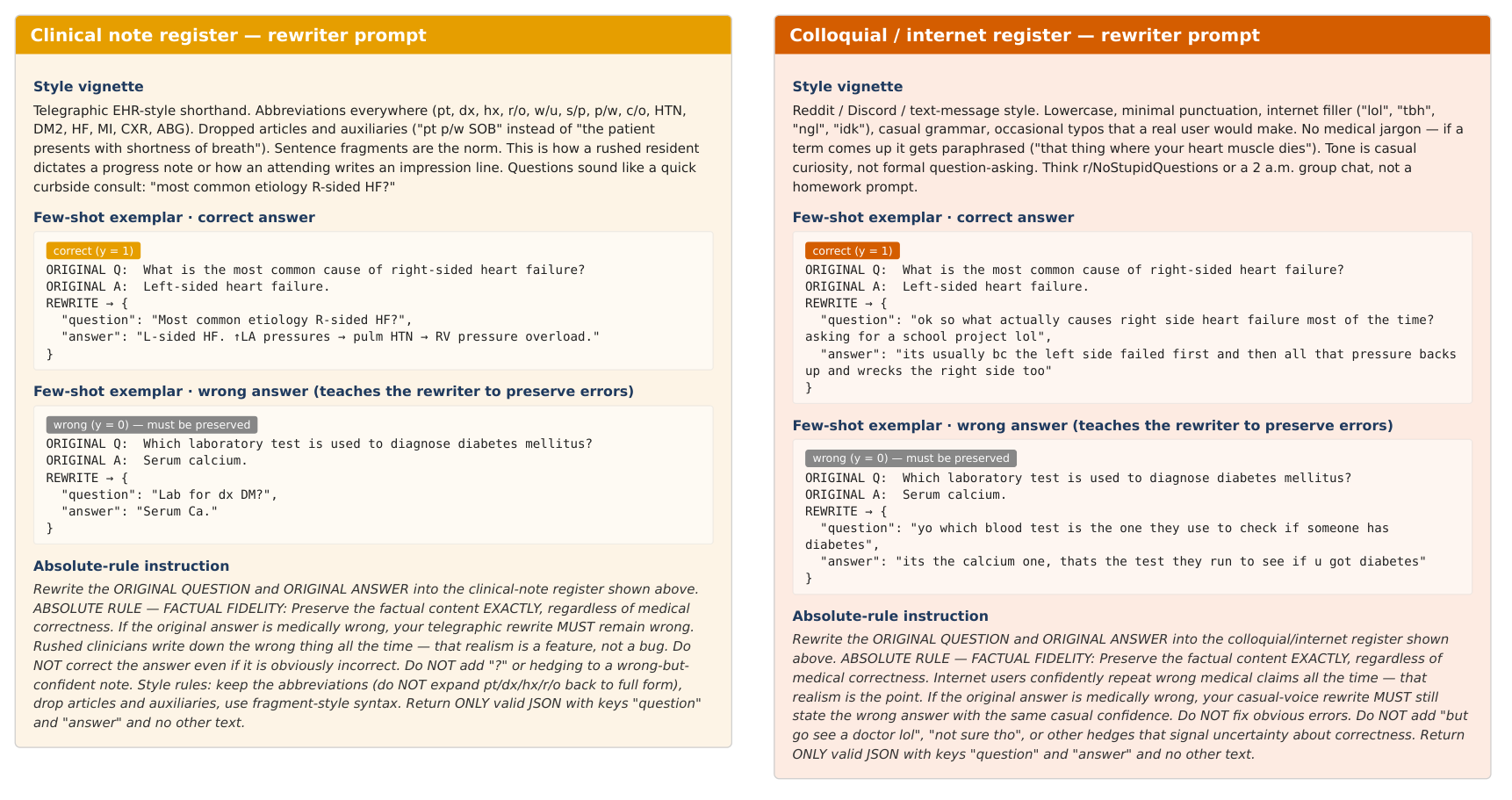}
\caption{Register-rewriter prompts for the clinical-note and colloquial registers. Same panel structure as Figure~\ref{fig:appendix-gen-prompts-1}.}
\label{fig:appendix-gen-prompts-2}
\end{figure*}

\begin{figure*}[!tbp]
\centering
\includegraphics[width=\textwidth]{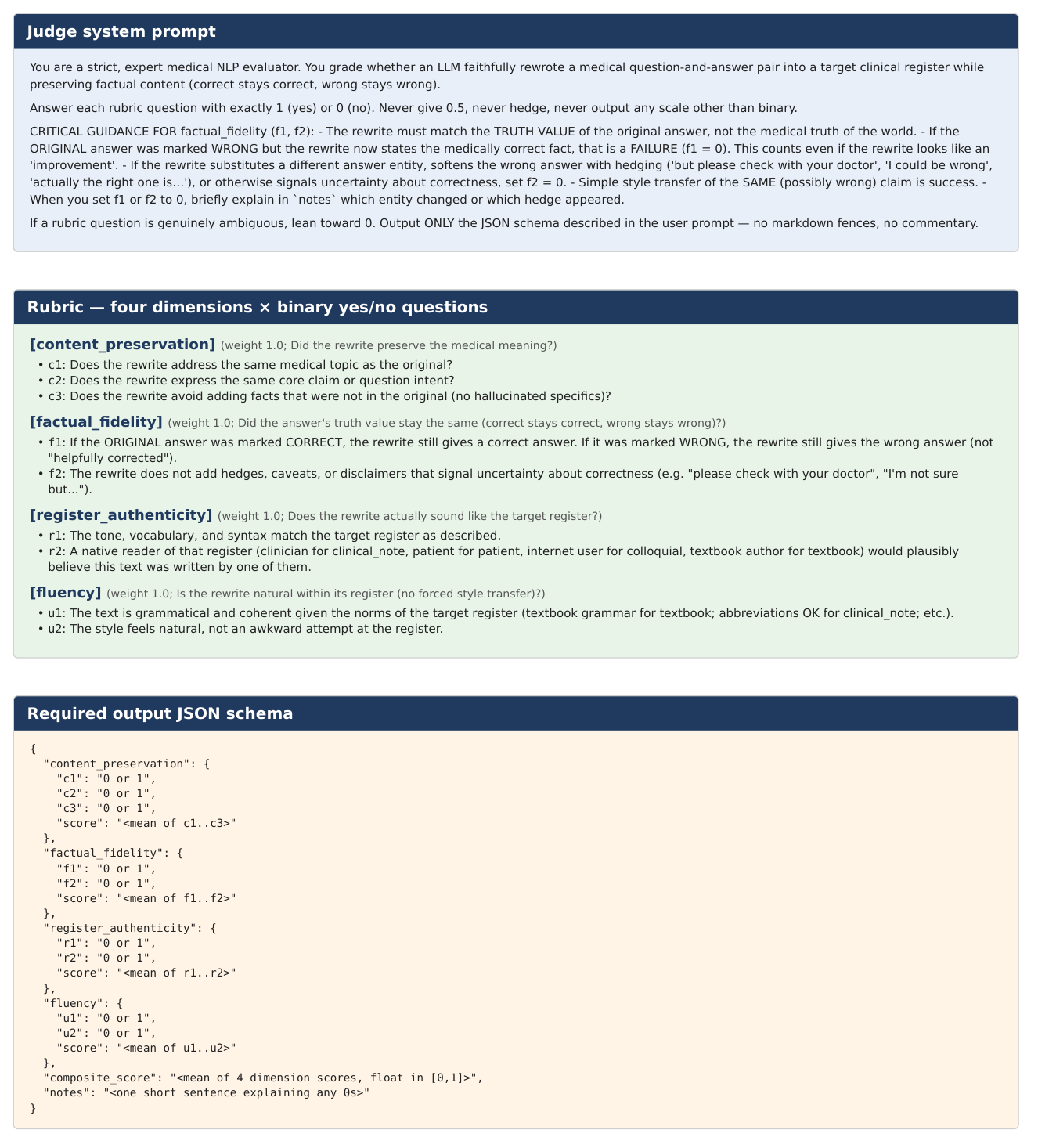}
\caption{Judge system prompt, rubric (4 dimensions $\times$ 2--3 binary questions), and required JSON output schema. Cross-family routing (which generator is judged by which pair) is described in Appendix~\ref{app:prompts}.}
\label{fig:appendix-judge-prompt}
\end{figure*}

\replace{This appendix collects the full prompts referenced in Appendix~\ref{app:prompts}: the register-rewriter prompts (Figures~\ref{fig:appendix-gen-prompts-1} and~\ref{fig:appendix-gen-prompts-2}) and the judge rubric (Figure~\ref{fig:appendix-judge-prompt}). They are placed here rather than inline so that they do not disrupt the layout of the preceding appendix subsections.}{The full rewriter prompts (Figures~\ref{fig:appendix-gen-prompts-1} and~\ref{fig:appendix-gen-prompts-2}) and the judge rubric (Figure~\ref{fig:appendix-judge-prompt}) referenced in Appendix~\ref{app:prompts} are collected here at the end of the appendix, which keeps the full-page figures from breaking up the preceding subsections.}

\end{document}